\documentclass{article}
\usepackage[final]{template}
\usepackage[utf8]{inputenc} 
\usepackage[T1]{fontenc}    
\usepackage{url}            
\usepackage{booktabs}       
\usepackage{amsfonts,amsmath,dsfont}       
\usepackage{nicefrac}       
\usepackage{microtype}      
\usepackage{xcolor}         
\usepackage{graphicx}

\usepackage{amsthm}
\usepackage{bbm}
\usepackage{amssymb}
\usepackage{xcolor}

\newtheorem{theorem}{Theorem}
\newtheorem{lemma}{Lemma}

\newtheorem{definition}{Definition}
\newtheorem{remark}{Remark}
\newtheorem{claim}{Claim}

\title{\vspace{-0.1in}On Non-asymptotic Theory of Recurrent Neural Networks in Temporal Point Processes\vspace{-0.1in}}
\author{%
 Zhiheng Chen${}^1$, ~ Guanhua Fang${}^{2*}$, ~ Wen Yu${}^2$ \\
${}^1$ Shanghai Center for Mathematical Sciences, Fudan University \\
${}^2$ Department of Statistics and Data Science, Fudan University \\
${}^*$ fanggh@fudan.edu.cn
}

\begin{document}

\maketitle

\begin{abstract}
Temporal point process (TPP) is an important tool for modeling and predicting irregularly timed events across various domains.
Recently, the recurrent neural network (RNN)-based TPPs have shown practical advantages over traditional parametric TPP models. 
However, in the current literature, it remains nascent in understanding neural TPPs from theoretical viewpoints. 
In this paper, we establish the excess risk bounds of RNN-TPPs under many well-known TPP settings. 
We especially show that an RNN-TPP with no more than four layers can achieve vanishing generalization errors.
Our technical contributions include the characterization of the complexity of the multi-layer RNN class, the construction of $\tanh$ neural networks for approximating dynamic event intensity functions, and the truncation technique for alleviating the issue of unbounded event sequences.
Our results bridge the gap between TPP's application and neural network theory.
\end{abstract}

\newpage

\section{Introduction}
Temporal point process (TPP) \citep{daley2003introduction, daley2008introduction} is an important mathematical framework that provides tools for analyzing and predicting the timing and patterns of events in continuous time. 
TPP particularly deals with event streaming data where the events occur at irregular time stamps, which is different from classical time series analysis that often assumes a regular time spacing between data points. 
In real world applications, the events could be anything from transactions in financial markets \citep{bauwens2009modelling, hawkes2018hawkes} to user activities in online social network platforms \citep{farajtabar2017coevolve, fang2023group}, earthquakes in seismology \citep{wang2012markov, laub2021elements}, neural spikes in biological experiments \citep{perkel1967neuronal, williams2020point}, or failure times in survival analysis \citep{aalen2008survival, fleming2013counting}. 

With advent of artificial intelligence in last decades, the neural network \citep{mcculloch1943logical} has been proved to be a powerful 
architecture that can be adapted to different applications with distinct purposes.
In modern machine learning, researchers have also incorporated deep neural networks into TPPs to handle complex patterns and dependencies in event data, leading to advancements in many areas such as recommendation systems \citep{du2015time, hosseini2017recurrent}, social network analysis \citep{du2016recurrent, zhang2021vigdet}, healthcare analytics \citep{li2018learning, enguehard2020neural}, etc.
Many new TPP models have been proposed in the recent literature, including but not limited to, recurrent temporal point process \cite{du2016recurrent},
fully neural network TPP model \cite{omi2019fully},
transformer Hawkes process \cite{zuo2020transformer}; 
see 
\cite{shchur2021neural,lin2022exploring} and the references therein
for a more comprehensive review.

Despite the recent process in TPP's applications as mentioned above, there is a lack of understanding in neural TPPs from the theoretical perspective. 
A fundamental question remains: whether the neural network-based TPP can provably have a small generalization error?
In this paper, we provide an affirmative answer to this question for \textit{recurrent neural network} (RNN, \cite{medsker1999recurrent})-based TPPs.
To be specific, we establish the non-asymptotic rates of generation error bounds under mild model assumptions and provide the construction of RNN architectures that could approximate many widely-used TPPs, including homogeneous Poisson process, non-homogeneous Poisson process, self-exciting process, etc.

There are a few challenges in developing the theory of RNN-based TPPs. 
(a) \textit{Characterization of functional space.}
In the machine learning theory, it is necessary to specify the model space to derive any generalization errors. 
In our setting, the thing becomes more complicated since the model should be data-dependent (i.e., adapts to the past events). Otherwise, the model could not capture the information in event history and fail to provide a good fitting.
(b) \textit{Expressive power of RNN architecture.}
RNN is the most widely adopted neural architecture in TPP modelling. 
However, it remains questionable whether the RNNs can approximate most well-known temporal point processes.
If the answer is yes, it would be of great interest to know how many hidden layers and how large hidden dimensions will be sufficient for the approximation.
(c) \textit{Expressive power of activation function.}
In modern neural networks, the activation function is chosen to be a simple non-linear function for the sake of computational feasibility.  
In RNNs, it is taken as the ``tanh" by default. 
Then it is important to understand the approximability of tanh activation functions.
(d) \textit{Variable length of event sequence.}
Unlike the standard RNN's modelling where each sample is assumed to have the same number of observations (events), the event sequences in our setting may vary from one to another.
In addition, their lengths are potentially unbounded. These add difficulties in computing the complexity of the model space.

To overcome the above challenges, we adopt the following approaches. 
(a) In TPPs, the intensity function is the core. 
We recursively construct the multi-layer hidden cells through RNNs to store the event information and adopt the suitable output layer to compute the intensity value. 
Equipped with suitable input embeddings, our construction can capture the information of event history and adapt to variable lengths of event sequence.
(b) For four main categories of TPPs, homogeneous Poisson process, non-homogeneous Poisson process, self-exciting process, and self-correcting process, we carefully study their intensity formula.
We can decompose the intensity function into different parts and approximate them component-wisely. 
Our construction explicitly gives the upper bounds on the model depth, the width of hidden layers, and parameter weights of the RNN architecture to achieve a certain level of approximation accuracy.
(c) We use the results in a recent work \citep{de2021approximation}, where they provide the approximation ability of one- and two-layer $\tanh$ neural networks. We adapt such results to our specific RNN structure and give the universal approximation results for each of the intensity components.  
(d) Thanks to the exponential decay property of the tail probability of the sequence length, we are able to use the truncation technique to decouple the randomness of independent and identically distributed (i.i.d.) samples and the lengths of event sequences.
For the space of truncated loss functions, the space complexity can be obtained through calculating the covering number. The classical chaining methods in empirical process theory can hence be applied as well.

Our main technical contributions can be summarized as follows.

(i)  
In the analysis of the stochastic error in the excess risk of RNN-based TPPs, we provide a truncation technique to decompose the randomness into a bounded component and a tail component. By carefully balancing between the two parts, we establish a nearly optimal stochastic error bound. Additionally, we also derive the complexity of the multi-layer RNN-based TPP class, where we precisely analyze and compute the Lipschitz constant of RNN architecture. This extends the existing result in \cite{chen2019generalization} where they only give the Lipschitz constant of a single-layer RNN. 
Therefore, our truncation technique and the Lipschitz result of multi-layer RNNs can be useful and of independent interest for many other related problems.

(ii) We establish the approximation error bounds for the intensity functions of TPPs of four main categories. 
To the best of our knowledge, there is very few work \citep{de2021approximation} on studying the approximation property of $\tanh$ activation function. Our work is the first one to provide approximation results for RNN-based statistical models. 
Our construction procedure largely depends on the Markov nature \citep{laub2021elements} of self-exciting processes so that we can design hidden cells to store sufficient information of past events. 
Moreover, we decompose the excitation function into different parts. Each of them is a simple smooth function (i.e. either exponential function or trigonometric function) that can be well approximated by a single-layer $\tanh$ network.
Our construction method can be viewed as a useful tool in analyzing other sequential-type neural networks.

(iii) 
We illustrate the differences between the architectures of classical RNNs and RNN-based TPPs. 
Note the fact that the observed events happen at the discrete time grids, while the TPP models should take into account the continuous time domain.
Therefore, the interpolation of values in hidden cells at each time point is important and necessary.
We show that improper interpolation mechanisms (e.g. constant, linear, exponential decay interpolation) may fail to provide RNN-based TPP with the universal approximation ability.
Our result indicates that the input embedding plays an important role in interpolating the hidden states. 

The rest of paper is organized as follows.
In Section \ref{sec:pre}, the background of TPPs, the formulation of RNN-based TPPs, and useful notations are introduced.
The main theories along with high-level explanations are given in Section \ref{main results section}.
The technical tools for analyzing stochastic errors are provided in Section \ref{stochastic error section(main)}.
The construction procedures for approximating different types of intensity functions are listed in Section \ref{approximation error section(main)}.
In Section \ref{counterexample section}, we provide explanations that the improper interpolation of hidden states in RNN-TPPs may lead to unsatisfactory approximation results.
The concluding remarks are given in Section \ref{sec:discuss}.

\section{Preliminaries} \label{sec:pre}

\subsection{Framework Specification}

We observe a set of $n$ irregular event time sequences, 
\begin{eqnarray}\label{eq:train:data}
\mathbf D_{train} := \{S_i; i = 1, ..., n\} = \{(t_{i,1}, ..., t_{i,N_{ei}}); i = 1, ..., n\},
\end{eqnarray}
where $0 < t_{i,1} < ... < t_{i,j} < ... < t_{i,N_{ei}} \leq T$ with $T$ being the end time point, and $N_{ei}$ is the number of events in the $i$-th sequence, $S_i$.
It is assumed that each of $S_i$'s is independently generated from a TPP model with an unknown intensity function $\lambda^{\ast}(t)$ defined on $[0,T]$. 
That is, 
\[\lambda^{\ast}(t) := \lim_{dt \rightarrow 0}
\frac{\mathbb E[N[t, t+dt) | \mathcal H_t]}{dt},\]
where $N[t, t+dt) := N(t+dt) - N(t)$ with $N(t) := \sharp\{i: t_i \leq t \}$ being the number of events observed up to time $t$, and 
$\mathcal H_t := \sigma(\{N(s); s < t\})$ is the history filtration before time $t$.

In the literature of TPP's learning \citep{shchur2021neural}, the primary goal is to estimate $\lambda^{\ast}(t)$ based on $\mathbf D_{train}$.
Throughout the current work, we adopt the negative log-likelihood function as our objective. To be specific,  for any event time sequence $S = (t_1, .., t_{N_e})$, we define 
\begin{eqnarray}\label{eq:ll:loss}
\text{loss}(\lambda, S) 
:= - \left\{ \sum_{j=1}^{N_e} \log \lambda(t_j) - \int_0^T \lambda(t)\mathrm{d}t \right\}.
\end{eqnarray}
Then the estimator can be defined as
\begin{eqnarray}\label{eq:main:est}
    \hat \lambda &:=& \arg \min_{\lambda \in \mathcal F} \text{loss}(\lambda) \nonumber \\ 
    &:= & \arg \min_{\lambda \in \mathcal F} \left\{\frac{1}{n} \sum_{i=1}^n \text{loss}(\lambda, S_i) \right\},
\end{eqnarray}
where $\mathcal F$ is a user-specified functional space. 
For example, in the existing works, $\mathcal F$ can be taken as any space of parametric models \citep{schoenberg2005consistent, laub2021elements}, nonparametric models \citep{cai2022latent, fang2023group}, or neural network models \citep{du2016recurrent, mei2017neural}.

In the language of deep learning, $\mathbf D_{train}$ is also called a training data set.
$\text{loss}(\lambda)$ is known as the loss function of predictor $\lambda$.
$\hat \lambda$ defined in \eqref{eq:main:est} is the empirical risk minimizer (ERM). 
To evaluate the performance of $\hat \lambda$, a common practice in machine (deep) learning is using the excess risk \citep{hastie2009elements, james2013introduction, vidyasagar2013learning, shalev2014understanding}.
To be mathematically formal, we define 
\begin{eqnarray}\label{eq:def:gen:err}
    \text{ER}(\hat \lambda) := 
    \mathbb E[\text{loss}(\hat \lambda, S_{test})] - 
    \mathbb E[\text{loss}(\lambda^{\ast}, S_{test})],
\end{eqnarray}
where $S_{test}$ is a testing sample, i.e., a new event time sequence, which is independent of $\mathbf D_{train}$ and also follows the intensity $\lambda^{\ast}(t)$.
The expectation here is taken with respect to the new testing data. 
We give a proof of $\text{ER}(\hat \lambda) \geq 0$ in the supplementary. As a result, \eqref{eq:def:gen:err} is a well-defined excess risk under our model setup.

\subsection{RNN Structure}
\label{RNN function section}

Throughout this paper, we consider $\mathcal F$ to be a space of RNN-based TPP models. An arbitrary intensity function $\lambda$ in $\mathcal F$, indexed by the parameter $\theta$, is defined through the following recursive formula,
\begin{eqnarray}\label{eq:form:lam}
    \lambda_{\theta}(t;S) &:=& f \left(W_x^{(L+1)} h^{(L)}(t;S) + b^{(L+1)}\right) \in \mathbb R^1, ~~ \text{for}~ t \in (t_j, t_{j+1}],
\end{eqnarray}
where the hidden vector function $h^{(L)}(t; S)$ has the following hierarchical form,
\begin{eqnarray}\label{eq:form:h}
    h^{(1)}(t; S) &=& \sigma\left(W_x^{(1)} x(t;S) + W_h^{(1)} h_{j}^{(1)} + b^{(1)}\right), \nonumber  \\
    h^{(2)}(t; S) &=& \sigma\left(W_x^{(2)} h^{(1)}(t;S) + W_h^{(2)} h_{j}^{(2)} + b^{(2)}\right), \nonumber \\
    &\vdots& \nonumber \\
    h^{(L)}(t; S) &=& \sigma\left(W_x^{(L)} h^{(L-1)}(t;S) + W_h^{(L)} h_{j}^{(L)} + b^{(L)}\right), ~~ \text{for}~ t \in (t_j, t_{j+1}],
\end{eqnarray}
with
\begin{eqnarray}\label{eq:form:h:j}
h_j^{(1)} &=& \sigma\left(W_x^{(1)} x(t_j;S) + W_h^{(1)} h_{j-1}^{(1)} + b^{(1)}\right), \nonumber \\
h_j^{(2)} &=& \sigma\left(W_x^{(2)} h_j^{(1)} + W_h^{(2)} h_{j-1}^{(2)} + b^{(2)}\right), \nonumber \\
&\vdots& \nonumber \\
h_j^{(L)} &=& \sigma\left(W_x^{(L)} h_j^{(L-1)} + W_h^{(L)} h_{j-1}^{(L)} + b^{(L)}\right), ~~ \text{for} ~ j \in \{1, ..., N_e\}. 
\end{eqnarray}
Here
$\sigma$, $f$ are two known activation functions of the hidden layers and the output layer, respectively. 
Both of them are pre-determined by the user.
We specifically take $\sigma (x) = \tanh(x) = (\exp(x) - \exp(-x))/(\exp(x) + \exp(-x))$ and $f(x) = \min\{\max\{x, l_f\}, u_f\}$, where $l_f$  and $u_f$ are two fixed positive constants. 
The input embedding vector function $x(t; S)$ is also known to the user before training. In the current work, we particularly take $x(t; S) = (t, t - F_S(t))^{\top}$ where $F_S(t) = t - t_j$ for $t \in (t_j, t_{j+1}]$, $\forall j \in N_e$. 
The model parameters consist of 
$W_x^{(l)}$, $W_h^{(l)}$, and $b^{(l)}$ ($1 \leq l \leq L$). For notational simplicity, we concatenate all parameter matrices and vectors and write as $\theta = \{W_x^{(l)}, W_h^{(l)}, b^{(l)}; 1\leq l \leq L + 1\}$, where $W_h^{(L+1)} \equiv \mathbf{0}$.  
By default, we take the initial values $t_0 \equiv 0$ and $h_0^{(l)} \equiv \mathbf 0$ for $1 \leq l \leq L$. The last time grid $t_{N_e + 1} \equiv T$.
We call the model defined through equations \eqref{eq:form:lam} - \eqref{eq:form:h:j} as the RNN-TPP.

\begin{figure}
    \centering
    \includegraphics[width = 0.48 \textwidth]{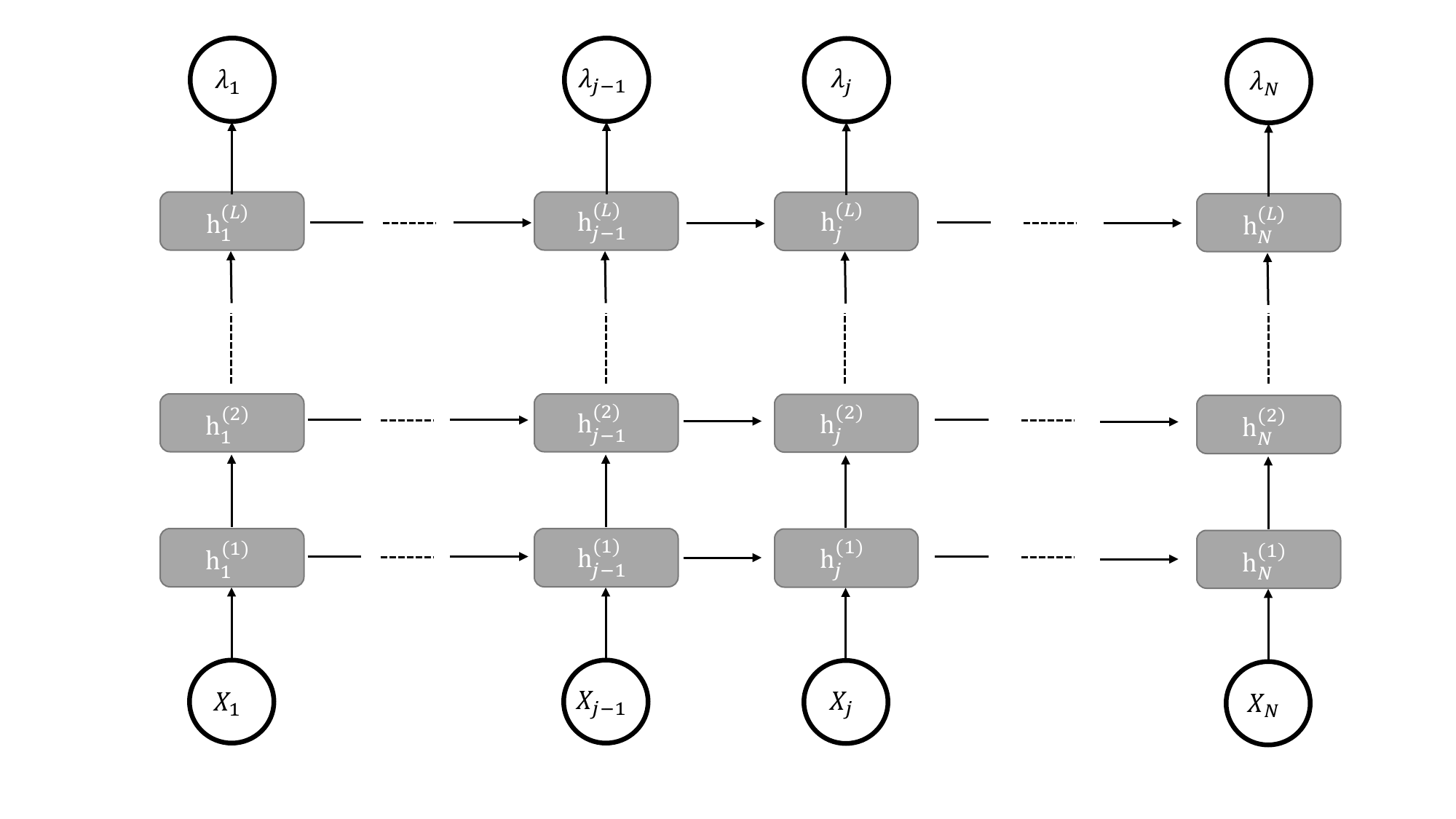}
    \includegraphics[width = 0.48 \textwidth]{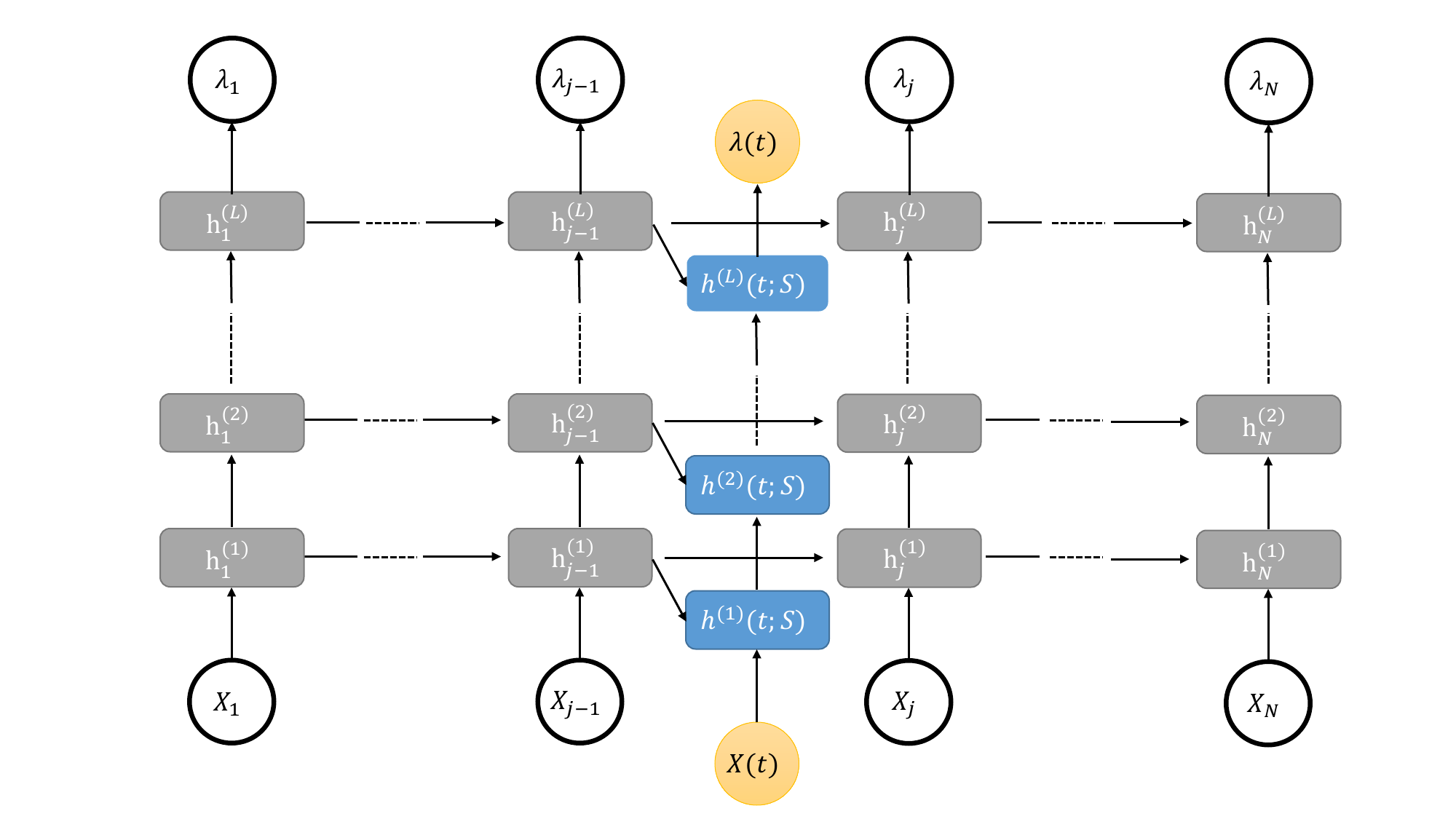}
    \caption{Left: the classical RNN architecture. Right: the RNN-TPP architecture given in \eqref{eq:form:lam} - \eqref{eq:form:h:j}. The blue box represents the interpolation of hidden states.}
    \label{fig::RNN:structure}
\end{figure}

Moreover, we define the maximum hidden size $D := \max\{ d_1, d_2 \cdots d_L\}$, where $d_l$ is the dimension of the $l$-th hidden layer, and the parameter norm 
$$
\|\theta\| := \max \left\{\|W_x^{(l)}\|_2, \|W_h^{(l)}\|_2, \|b^{(l)}\|_2; 1\leq l \leq L + 1\right\} .
$$
Then the RNN-TPP class $\mathcal{F}$ is described by
\begin{eqnarray}
    \mathcal F = \mathcal{F}_{L, D, B_m, l_f, u_f} := 
    \{\lambda_{\theta} ;~ \|\theta\| \leq B_m \},
    \label{rnn_func_class(main)}
\end{eqnarray}
where $B_m$ may depend on the hidden size $D$ and the sample size $n$.
To help readers gain more intuitions, a graphical illustration of the network structure is given in Figure \ref{fig::RNN:structure}.

\begin{remark}
    The default choice \citep{de2021approximation} of activation function $\sigma(x)$ in RNNs is $\tanh(x)$.
    In practice, the number of layers $L$ is usually no more than 4.
\end{remark}

\begin{remark}
    By the constructions \eqref{eq:form:lam} - \eqref{eq:form:h:j}, it is not hard to see that the intensity $\lambda_{\theta}(t; S)$ is a left-continuous function of $t$.
    In other words, it is a well-defined predictable function with respect to the information filtration generated by event sequence $S$.
\end{remark}

\begin{remark}
  In the standard application of RNN models, the training data usually consist of discrete-time sequences (e.g., sequences of tokens in natural language processing (NLP) \citep{yin2017comparative, tarwani2017survey}; time series in financial market forecasting \citep{cao2019financial, chimmula2020time}).
  Therefore, the classical (single-layer) RNN architecture is defined only through the discrete time grids. That is, the hidden vector at $j$-th grid is
  \begin{align}
    h_j &= \sigma\left(W_x x_j + W_h h_{j-1} + b_h \right), \nonumber 
\end{align}
   where $x_j$ is the corresponding embedding input. The prediction at time step $j$ is given by $y_j = f(W_y h_j + b_y) \in \mathbb R$.
   In contrast, the RNN-based TPP model should take into account any time point $t$ between grids $t_j$ and $t_{j+1}$. 
   Hence the interpolation of $h^{(l)}(t;S)$ between $h_j^{(l)}$ and $h_{j+1}^{(l)}$ is heuristically necessary to give reasonable model predictions over the entire time interval $(t_j, t_{j+1}]$.
\end{remark}

\begin{remark}
    In the literature, there exist a few methods to interpolate the hidden embedding between $h_j^{(L)}$ and $h_{j+1}^{(L)}$.
    In \cite{du2016recurrent}, a constant embedding mechanism is used, i.e. $h^{(l)}(t;S) \equiv h_j^{(l)}$ for $t \in (t_j, t_{j+1}]$ and any $j$ and $l$.
    In \cite{mei2017neural}, the author adopted an exponential decay method to encode the hidden representations under an extended RNN architecture, Long Short Term Memory (LSTM) network.
    More recently, \cite{rubanova2019latent} used the neural ordinary differential equation (ODE) method for solving the intermediate hidden state $h^{(l)}(t;S)$.

    It can be shown that the first two interpolation methods are unable to precisely capture the true intensity in the sense of excess risk. We will give the explanation in Section \ref{counterexample section}; see Theorem \ref{thm:counterexample}.
\end{remark}

\begin{remark}
    Our result still holds if \text{tanh} is replaced with other Sigmoidal-type activation functions \citep{cybenko1989approximation} (e.g., ReLU \citep{fukushima1969visual}).
    In the literature of TPP modelling, the most common choice of $f(x)$ is the Softplus function \citep{dugas2001intolerance, zhou2022neural}, $\log(1 + \exp(x))$, which ensures $\lambda_{\theta}(t;S)$ to be positive and differentiable. 
    Our result also holds if we take $f(x)$ to be  
    $\min\{\max\{\log(1 + \exp(x)), l_f\}, u_f\}$ with $0 < l_f < u_f$.
    Introducing $l_f$ and $u_f$ only serves the technical purpose, i.e., the predicted intensity value is bounded from above and below.
\end{remark}

\subsection{Classical TPPs}
\label{tpp model section}

In the statistical literature, TPPs can be categorized into several types based on the nature of the intensity functions.
Four main categories are summarized as follows.

\noindent 
\textbf{Homogeneous Poisson process} \citep{kingman1992poisson}. It is the simplest type where events occur completely independently of one another, and the intensity function is constant, i.e., $\lambda^{\ast}(t) \equiv \lambda$, where $\lambda$ is unknown and needs to be estimated.

\noindent 
\textbf{Non-homogeneous Poisson process} \citep{kingman1992poisson, daley2003introduction}.
In this model, the intensity function varies over time but is still independent of past events. That is, $\lambda^{\ast}(t)$ is a non-constant unknown function that is usually estimated via certain nonparametric methods.

\noindent
\textbf{Self-exciting process} \citep{hawkes1974cluster}. 
Future events are influenced by past events, which can lead to clustering of events in time. A well-known example is the Hawkes process \citep{hawkes1971spectra, hawkes1974cluster}, where the intensity function takes form,
\begin{eqnarray}\label{eq:label:hawkes}
    \lambda^{\ast}(t) = \lambda_0(t) + \sum_{j: t_j < t} \mu(t - t_j),
\end{eqnarray}
where $\lambda_0(t)$ and $\mu(t)$ are some positive functions which are called the background intensity
and excitation/impact function, respectively.
In many applications \citep{laub2021elements}, the excitation function takes the exponential form that $\mu(t) = \alpha \exp(-\beta t)$, which allows the efficient computation.
The model defined in \eqref{eq:label:hawkes} is also known as the \textit{linear} self-exciting process since the intensity is in an additive form of different components.
More generally, the non-linear self-exciting process \citep{bremaud1996stability}
\begin{eqnarray}\label{eq:label:non:linear}
    \lambda^{\ast}(t) = \Psi\left(\lambda_0(t) + \sum_{j: t_j < t} \mu(t - t_j)\right),
\end{eqnarray}
is also considered in the literature,
where $\Psi$ is a non-linear function. 

\noindent \textbf{Self-correcting process} \citep{isham1979self, ogata1984inference}. The occurrence of an event decreases the likelihood of future events for some time period. 
To be mathematically formal, the intensity postulates the formula,
\begin{eqnarray}\label{eq:self-correct}
\lambda^{\ast}(t) = \Psi\left(\mu t - \sum_{j: t_j < t} \alpha\right),
\end{eqnarray}
where both $\mu$ and $\alpha$ are positive and $\Psi$ may be a non-linear function.

\subsection{Notations}
Let $a\wedge b = \min\{a,b\}$ and $a\vee b = \max\{a,b\}$. We use $\mathbb N$ and $\mathbb Z$ to denote the set of nonnegative integers and all integers, respectively. Denote $[n] = \{1,2 \cdots, n\}$ for a positive integer $n$. Let $\lceil a \rceil = \min\{b\in \mathbb Z, b \geq a\}$. For a set $A$, denote $\#(A)$ to be its cardinality. For a vector $x=(x_1, \cdots, x_d)^{\top} \in \mathbb R^d$, denote its Euclidean norm as $\|x\|_2 =\sqrt{\sum_{i=1}^d x_i^2}$.  Write $a_N \lesssim b_N$ if there exists some constant $C> 0$ such that $a_N \leq C b_N$ for all index $N$, and the range of $N$ may be defined case by case.
For a function $f$ defined on some domain, denote $\|f\|_{L^\infty}$ as its essential upper bound. 
For $s \in \mathbb N$, the Sobolev norm $\|f\|_{W^{s, \infty}([0,T])}$ is defined as $\|f\|_{W^{s, \infty}([0,T])} = \max_{0\leq |\alpha| \leq m} \|D^{\alpha}f\|_{L^{\infty}([0,T])}$. For a constant $B_0 > 0$, the $B_0$-ball  of Sobolev space $ W^{s, \infty}([0,T])$ is defined as
\begin{align}
    W^{s, \infty}([0,T], B_0) := \left\{f \in W^{s, \infty}([0,T]), \|f\|_{W^{s, \infty}([0,T])} \leq B_0 \right\} . \nonumber
\end{align}
For constant $C_0 > 0$, the ball $C^{s, \infty}([0,T], C_0)$ is a subset of $W^{s, \infty}([0,T], C_0)$ which contains all $s$-order smooth functions. We use $O(\cdot)$ to hidden all constants and use $\tilde{O}(\cdot)$ to denote $O(\cdot)$ with hidden log factors. Throughout this paper, $\alpha$,  $\beta$, $\gamma$, $\mathcal{C}$, and $\mathcal{C}_1$ are positive real numbers and may be defined case by case.

\section{Main Results}
\label{main results section}
Recent applications in event stream analyses have witnessed the usefulness of TPPs with the incorporation of RNNs. 
However, there is no study in the existing literature to explain why the RNN structure in TPP modeling is so useful from the theoretical perspective.
We attempt to answer the question of whether the RNN-TPPs can \textit{provably} have small generalization error or excess risk. 
Our answer is \textbf{positive}! 
When the event data are generated according to the classical models described in Section \ref{tpp model section}, we show that the RNN-TPPs can perfectly generalize such data.

To make our presentation easier, we only need to focus on the self-exciting processes.
\footnote{Homogeneous Poisson, non-homogeneous Poisson, and self-correcting process can be treated similarly due to the following reasons.
If we take $\mu(t) \equiv 0$ in \eqref{eq:label:hawkes}, the linear self-exciting process reduces to the homogeneous Poisson or non-homogeneous Poisson process. In the RNN-TPP architecture, we can take the input embedding function $x(t; S) = (t, t - F_S(t), N(t-))$, i.e., using an additional input dimension to store the number of past events. Then establishing the excess risk of self-correcting process is technically equivalent to that of non-homogeneous Poisson process.}
To start with, we first consider the linear case \eqref{eq:label:hawkes}.

Some regularity assumptions should be stated before we present the main theorem. 

(A1) There exists a constant $B_0 > 0$ such that $\lambda_0 \in  W^{s, \infty}([0,T], B_0)$, where $s\geq 1$, $s \in \mathbb N$.

(A2) $\int_0^T \mu(t) \mathrm{d} t  := c_\mu < 1$.

(A3) There exists a positive constant $B_1$ such that $\inf_{t \in [0,T]} \lambda_0(t) \geq B_1$. 

Assumption (A1) assumes the boundedness of the background intensity, which is also common in neural network approximation studies. Assumption (A2) is standard in the literature of Hawkes process, which guarantees the existence of a stationary version of
the process when $\lambda_0(t)$ is constant. Assumption (A3) is an informative lower bound assumption, which ensures that sufficient intensity exists in any subdomain of $[0,T]$.

Now we can present the results on the
non-asymptotic bound of excess risk \eqref{eq:def:gen:err} under model \eqref{eq:label:hawkes}.

\begin{theorem}
\label{main theorem}
    Under model \eqref{eq:label:hawkes} and RNN-TPP class $\mathcal F = \mathcal{F}_{L, D, B_m, l_f, u_f}$ defined as \eqref{rnn_func_class(main)}, 
    suppose that assumptions (A1)-(A3) hold, 
    then for $n$ i.i.d. sample series $\{S_i, i \in [n]\}$, with probability at least $1-\delta$,
    the excess risk \eqref{eq:def:gen:err} of ERM \eqref{eq:main:est} satisfies:
    
    (i) (Poisson case) If $\mu \equiv 0$, for $L = 2$, $D = \tilde{O}(n^{\frac{1}{2(s+1)}})$, $B_m = \tilde{O}(n^{\frac{s+1}{4}})$, $l_f = B_1 \wedge 1$, and  $u_f = B_0$, 
    \begin{align}
        \mathbb E[\text{loss}(\hat \lambda, S_{test})] - \mathbb E[\text{loss}(\lambda^{\ast}, S_{test})]
        \leq \tilde{O}\left( n^{-\frac{s}{2(s+1)}} \right);
        \label{eq:result:poisson}
    \end{align}
    (ii) (Vanilla Hawkes case) If $\mu(t) = \alpha\exp(-\beta t)$, for $L = 2$, $D = \tilde{O}(n^{\frac{1}{2(s+1)}})$, $B_m = \tilde{O}((\log n)^{3s^2\log^2 n})$, $l_f = B_1 \wedge 1$, and $u_f = B_0 + O(\log n)$, 
    \begin{align}
        \mathbb E[\text{loss}(\hat \lambda, S_{test})] - \mathbb E[\text{loss}(\lambda^{\ast}, S_{test})]
        \leq \tilde{O}\left( n^{-\frac{s}{2(s+1)}} \right);
    \end{align} 
    (iii) (General case) If $\mu \in C^{k,\infty}([0,T], C_0)$, $k \geq 2$, $k \in \mathbb N$, for $L = 2$, $D = \tilde{O}(n^{\frac{1}{2}\left(\frac{1}{s+1} \vee \frac{5}{k+4}\right)})$, $B_m = \tilde{O}( (\log n)^{3s^2\log^2 n})$, $l_f = B_1 \wedge 1$, and $u_f = B_0 + O(\log n)$,
    \begin{align}
        \mathbb E[\text{loss}(\hat \lambda, S_{test})] - \mathbb E[\text{loss}(\lambda^{\ast}, S_{test})]
        \leq \tilde{O}\left( n^{-\frac{1}{2}\left(\frac{s}{s+1} \wedge \frac{k-1}{k+4}\right)} \right) .
     \end{align}
\end{theorem}

As suggested in Theorem \ref{main theorem}, there exists a two-layer RNN-TPP model whose excess risk becomes vanishing when the size of the training set goes to infinity.
The width of such network grows with the sample size, while the depth remains two.

\begin{remark}
    Here we require the depth of RNN-TPP $L=2$ due the fact that  $\lambda_0 \in  W^{s, \infty}([0,T], B_0)$.
    However, if we allow $\lambda_0$ to be sufficiently smooth (i.e., $\lambda_0 \in C^{\infty}([0,T])$), we only need one-layer $\tanh$ neural network to approximate $\lambda_0$. As a result, the number of layers of RNN-TPP can be reduced to one.
\end{remark}
Now we consider the true model to be a non-linear Hawkes process, which is given in \eqref{eq:label:non:linear}. For simplicity, we only consider the case $\mu(t) = \alpha \exp(- \beta t)$, which is
\begin{align}
    \lambda^{\ast}(t) = \Psi\left(\lambda_0(t) + \sum_{t_i < t} \alpha\exp(-\beta (t - t_i))\right).
    \label{non linear intensity(main)}
\end{align}
The regularity of $\Psi$ is presented as Assumption (A4).

(A4) Function $\Psi$ is $L$-Lipschitz, positive and bounded. In other words, there exist $ \tilde{B_1}, \tilde{B_0} > 0$ such that $\tilde{B_1} \leq \Psi \leq \tilde{B_0}$ and $|\Psi(x_1) - \Psi(x_2)| \leq L |x_1 - x_2|$ for any $x_1, x_2$.

We have a similar bound of excess risk \eqref{eq:def:gen:err} under model \eqref{non linear intensity(main)}. 

\begin{theorem}
\label{non linear case thm}
    (Nonlinear Hawkes Case) Under model \eqref{non linear intensity(main)} and RNN-TPP class $\mathcal F = \mathcal{F}_{L, D, B_m, l_f, u_f}$ defined as \eqref{rnn_func_class(main)}, 
    suppose that assumptions (A1) and (A4) hold, 
    then for $n$ i.i.d. sample series $\{S_i, i \in [n]\}$, 
    with probability at least $1-\delta$,
    for $L = 4$, $D = \tilde{O}(n^{\frac{1}{4}})$, $B_m = \tilde{O}((\log n)^{3s^2\log^2 n})$, $l_f = \tilde{B}_1 \wedge 1$, and $u_f = \tilde{B}_0$, the excess risk \eqref{eq:def:gen:err} of ERM \eqref{eq:main:est} satisfies:
    \begin{align}
         \mathbb E[\text{loss}(\hat \lambda, S_{test})] - 
    \mathbb E[\text{loss}(\lambda^{\ast}, S_{test})] \leq \tilde{O}\left( n^{-\frac{1}{4}} \right).
    \end{align}
\end{theorem}

For the non-linear case, as indicated by Theorem \ref{non linear case thm}, we require a deeper RNN-TPP with four layers to achieve the vanishing excess risk. Under the Lipschitz assumption of $\Psi$, the width of the hidden layers is of order $n^{1/4}$. When $\Psi$ is allowed to have higher-order smoothness, the width can reduce to that of the vanilla Hawkes case.

\begin{remark}
    (i) Two additional layers of RNN are required for the approximation of the arbitrary non-linear Lipschitz continuous function $\Psi$. (ii) For the model $\lambda^{\ast}(t) = \Psi\left(\lambda_0(t) + \sum_{t_i < t} \mu(t-t_i)\right)$ with general excitation function $\mu$, we can obtain the similar excess risk bound using the same technique in the proof of Theorem \ref{main theorem}. 
\end{remark}

To better explain the excess risks that obtained in Theorems \ref{main theorem}-\ref{non linear case thm}, we depend on the following decomposition lemma. 
\begin{lemma}
Let $\check \lambda^{\ast} = \arg\min_{\lambda \in \mathcal{F}} \mathbb E[\text{loss}(\lambda, S_{test})]$, for any random sample $\{S_i, i \in [n]\}$, the excess risk of ERM \eqref{eq:main:est} satisfies
\begin{align}
    \mathbb E[\text{loss}(\hat \lambda, S_{test})] - \mathbb E[\text{loss}(\lambda^{\ast}, S_{test})] &\leq \underbrace{2\sup_{\lambda \in \mathcal{F}} \Big|\mathbb E[\text{loss}(\lambda, S_{test})] - \frac{1}{n} \sum_{i\in [n]} \text{loss}(\lambda,S_i)\Big|}_{\text{stochastic error}} \nonumber \\
    & + \underbrace{\mathbb E[\text{loss}(\check \lambda^{\ast}, S_{test})] - \mathbb E[\text{loss}(\lambda^{\ast}, S_{test})]}_{\text{approximation error}}. 
    \label{bias-variance decomposition(main)}
\end{align}
 \label{variance-bias decomposition lemma(main)}  
\end{lemma}
By Lemma \ref{variance-bias decomposition lemma(main)}, the excess risk of ERM is bounded by the sum of two terms, the stochastic error $2\sup_{\lambda \in \mathcal{F}} |\mathbb E[\text{loss}(\lambda, S_{test})] - n^{-1} \sum_{i\in [n]} \text{loss}(\lambda,S_i)|$ and the approximation error $\mathbb E[\text{loss}(\check \lambda^{\ast}, S_{test})] - \mathbb E[\text{loss}(\lambda^{\ast}, S_{test})]$. The first term can be bounded by the complexity of the function class $\mathcal{F}$ using the empirical process theory, where the unboundedness of the loss function needs to be handled carefully; we present the details in section \ref{stochastic error section(main)}. The second term characterizes the approximation ability of the RNN function class 
$\mathcal{F}$ to the true intensity $\lambda^{\ast}$ under the measure of the expectation of the negative log-likelihood loss function. In order to bound this term, we need to carefully construct a suitable RNN which can approximate $\lambda^{\ast}$ well. This has not been studied yet in the literature; see section \ref{approximation error section(main)} for the details.

Based on Lemma \ref{variance-bias decomposition lemma(main)}, the results in Theorem \ref{main theorem} admit the following form, 
\begin{align}
    \mathbb E[\text{loss}(\hat \lambda, S_{test})] - 
    \mathbb E[\text{loss}(\lambda^{\ast}, S_{test})] \leq O\left(\frac{C(N)}{\sqrt{n}} + \frac{1}{R(N)}\right), \nonumber
\end{align}
where $ C(N)/\sqrt{n}$ is the stochastic error and $1/R(N)$ is the approximation error. $C(N)$ is the complexity of RNN function class $\mathcal{F}$ and $R(N)$ is the corresponding approximation rate, where $N$ is a tuning parameter. For the Poisson case, 
we can construct a two-layer RNN-TPP with $O(N)$ width to achieve $O(N^{-s})$ approximation error. 
Hence $C(N) = O(N)$, $R(N) = O\left(N^s\right)$, and the final excess risk bound is $\tilde{O}(n^{-\frac{s}{2(s+1)}})$ in \eqref{eq:result:poisson}.  For the vanilla Hawkes case, since the exponential function is $C^{\infty}$-smooth, 
we only need extra $O(\text{Poly}(\log N))$ hidden cells in each layer to obtain $\tilde{O}(N^{-s})$ approximation error, and then we have the same order excess risk bound. For the general case, motivated by the vanilla Hawkes case, we decompose $\mu \in C^{k,\infty}([0,T], C_0)$ into two parts. One part is a polynomial of exponential functions which can be well approximated by $O(\text{Poly}(\log N))$-width \text{tanh} neural network. 
The other part is a function $\tilde{\mu} \in C^{k,\infty}([0,T], \tilde{C_0})$ satisfying $\tilde{\mu}^{(j)}(0+) = \tilde{\mu}^{(j)}(T-)$, $j = 0,1,\cdots,k-1$. It is easy to check that the $r$-th Fourier coefficients of $\tilde \mu$, $\hat{\mu}_r$, decay at the rate of $r^{-k}$. 
Then it is sufficient to approximate the first $N$ functions in the Fourier expansion of $\tilde \mu$ to get $\tilde{O}(N^{-(k-1)})$ approximation error, which additionally costs $\tilde{O}\left(N^{5}\right)$ complexity (see section \ref{approximation for general case} for details). Combining this with the approximation result of $\lambda_0$, we get the final bound $\tilde{O}( n^{-\frac{1}{2}\left(\frac{s}{s+1} \wedge \frac{k-1}{k+4}\right)})$. Similarly, for the nonlinear Hawkes case, we need $\tilde{O}(N)$ complexity to obtain $\tilde{O}(N^{-1})$ approximation error, which leads to $\tilde{O}({n}^{-\frac{1}{4}})$ excess risk bound.

As we emphasize in the above remarks, the number of layers depends on the smoothness of $\lambda_0$. If $\lambda_0 \in C^{\infty}([0,T])$ and $\|\lambda_0\|_{W^{s,\infty}} \leq C^s$, we only need one-layer $\tanh$ neural network to approximate $\lambda_0$, hence the number of layers in RNN-TPP can be reduced to one.

\section{Stochastic Error}
\label{stochastic error section(main)}

In this section, we focus on the stochastic error in \eqref{bias-variance decomposition(main)}. 
This type of stochastic error for the RNN function class has been studied in the recent literature, such as \cite{chen2019generalization} and \cite{Tu2020UnderstandingGI}. 
However, they only consider the case where the lengths of the input sequences are bounded, which is not applicable under the TPP setting. Here we establish an upper bound of the stochastic error in \eqref{bias-variance decomposition(main)} by a novel decoupling technique to make the classical results applicable. This technique can be used in many other related problems.

\subsection{Main Variance Term}

We first give out some mild assumptions for the RNN-TPP function class $\mathcal{F}$ under a more general framework.

(B1) The embedding function $x(\cdot)$ is bounded by a constant $B_{in}(T)$ on the time domain $[0,T]$, i.e. $\|x(\cdot)\|_2 \leq B_{in}(T)$.

(B2) The parameter $\theta$ lies in a bounded domain $\Theta$. 
More precisely, we assume that the spectral norms of weight matrices (vectors) and other parameters are bounded respectively, i.e., $\|W_x^{(l)}\|_2 \leq B_x$, $\|W_h^{(l)}\|_2  \leq B_h$,  $\|b^{(l)}\|_2 \leq B_b$, $1 \leq l \leq L+1$, and $B_m = \max\{B_b, B_h, B_x\}$.

(B3) Activation functions $\sigma$ and $f$ are Lipschitz continuous with parameters $\rho_{\sigma}$ and $\rho_f$ respectively, $\sigma(0)=0$, and there exists $|b_0| \leq B_b$ such that $f(b_0) = 1$ . Additionally, $\sigma$ is entrywise bounded by $B_{\sigma}$, 
and $f$ satisfies $l_f \leq \|f\|_{L^{\infty}} \leq  u_f$.

Now we consider the first term of \eqref{bias-variance decomposition(main)}. For convenience, we denote $X_\theta = \mathbb E [\text{loss}(\lambda_\theta, S_{test})] - n^{-1} \sum_{i = 1}^{n} \text{loss}(\lambda_\theta, S_i)$. 

\begin{theorem} \label{main variance theorem}
     Under assumptions (B1)-(B3) and suppose the event number $N_e$ satisfies the tail condition
     \[\mathbb P(N_e \geq s) \leq a_N \exp(-c_N s), ~ s \in \mathbb N,\] 
     with probability at least $1-\delta$, we have
    \begin{align}
        \sup_{\theta \in \Theta} |X_\theta| \leq &
        \frac{192}{\sqrt{n}}\left(T + \frac{1}{l_f}\right)(s_0+1)u_f \Bigg(\sqrt{\log\left(\frac{4}{\delta}\right)} + D \sqrt{(3L+2)}\left(\sqrt{\log\left(1+ M(s_0)\right)}+1\right) \nonumber \\
        & + \frac{1}{(1-\exp(-c_N))^2}~\Bigg)~. \nonumber
    \end{align}
    Thus
    \begin{align}
        \sup_{\|\theta\| \leq B_m} |X_\theta| \leq \tilde{O}\left(\sqrt{\frac{D^2 L^2}{n}} \right)~,
        \label{sto err simplified version}
    \end{align}
    where $s_0 = \lceil{c_N}^{-1}\left(\log\left(2a_{N}n/\delta\right) - 1 \right)\rceil$, $M(s) = \rho_f B_m\sqrt{D}(B_\sigma \sqrt{D} \vee B_{in}(T) \vee 1) ( \gamma^L \vee 1) (s+1)^{L-1}(\beta^{s+1}-1)/(\beta-1) $, $\gamma = \rho_\sigma B_x$, $\beta = \rho_\sigma B_h$. 
\end{theorem}
\begin{remark}
    There exist constants $a_N, c_N$ so that the tail condition $\mathbb P(N_e \geq s) \leq a_N \exp(-c_N s), s \in \mathbb N$ always holds for (non) homogeneous Poisson processes, linear and nonlinear Hawkes processes, and self-correcting processes under weak assumptions. To be more concrete, Lemma \ref{lemma of N_e} in the following section gives a result for the linear case.
\end{remark}
\begin{remark}
    For one-layer RNN with width $D$ and bounded sequence length $T$, \cite{chen2019generalization} gives a $\tilde{O}\big(\sqrt{{D^3 T}/{n}}\big)$ type stochastic error bound. Our bound reduces the term $D^3$ to $D^2$, thanks to the bounded output layer, i.e., $f(x) = \min\{\max\{x, l_f\}, u_f\}$. The term $D^2$ is also order-optimal by noticing that the number of free parameters in a single-layer RNN is at least $D^2$. 
\end{remark}

The stochastic error in \eqref{main variance theorem} is  
mainly determined by the complexity of the RNN function class $\mathcal{F}$, which will be discussed in the following section. To obtain this bound, we need to handle the unboundedness of the event number. We use a truncation technique to decouple the randomness of the tail of $N_e$, which allows us to use classical empirical process theory to derive the upper bound. Our computation is motivated by \cite{chen2019generalization}, which gives the generalization error bound of a single-layer RNN function class.

\subsection{Key Techniques}

To be reader-friendly, the main techniques for proving Theorem \ref{main variance theorem} are summarized as follows.

\subsubsection{Probability Bound of Events Number}

Define $N_{e(n)} := \max \{N_{ei}, 1 \leq i \leq n\}$. The following lemma characterizes the tails of event number $N_e$ and $N_{e(n)}$ under model \eqref{eq:label:hawkes} and assumptions (A1) and (A2) (For assumption (A1), we only need $\lambda_0 \leq B_0$ in this section). The proof is similar to Proposition 2 in \cite{hansen2015lasso}; see supplementary for the details.

\begin{lemma}
    \label{lemma of N_e}
    For model \eqref{eq:label:hawkes}, under assumptions (A1) and (A2), with probability at least $1-\delta$, we have
    \begin{align}
        N_{e(n)} <  \frac{1}{1-c_\mu \eta} \left( \frac{2}{\log(\eta)} \log\left(\frac{2n\sqrt{B_0 T}}{\delta(1-c_\mu)} \right) + \eta(B_0 T) \right). \nonumber
    \end{align}
    Hence
    \begin{align}
        \mathbb P\left(N_e = s\right) \leq \mathbb P\left(N_e \geq s\right) \leq \frac{2\sqrt{B_0 T}}{1-c_\mu} \exp\left( \frac{\log(\eta)}{2} \left[\eta(B_0 T) - (1-c_\mu\eta)s\right] \right), \nonumber
    \end{align}
    where $\eta \in \left(1, c_\mu^{-1}\right)$.
    Let $a_N = 2\sqrt{B_0 T} \exp(\log(\eta_0)\eta_0(B_0 T)/2)/(1-c_\mu)$ and $c_N = \log(\eta_0)(1-c_\mu\eta_0)/2$ with $\eta_0 \in \left(1, c_\mu^{-1}\right)$ being fixed. Then
    \begin{align}
        \mathbb P\left(N_e = s\right) \leq \mathbb P\left(N_e \geq s\right) \leq a_N \exp(- c_N s) .
        \label{prob. bound of N_e}
    \end{align} 
    \label{prob. bound event number lemma}
\end{lemma}
Our result is more refined than Proposition 2 in \cite{hansen2015lasso}, with computing all the constants and giving a tuning parameter to control the probability bound.

For the nonlinear case \eqref{eq:label:non:linear}, under Assumption (A4), we can obtain results similar to the non-homogeneous Poisson case, which are included in the above Lemma. 

\subsubsection{From Unboundedness to Boundedness}
The following lemma is the key to handling  the unboundedness of $X_\theta$, i.e., the unboundedness of the loss function.
For any $s \in \mathbb N$, we let $X_\theta(s) = \mathbb E \left[\text{loss}(\lambda_\theta, S_{test}) \mathbbm{1}_{\{N_e \leq s\}}\right] - n^{-1} \sum_{i = 1}^{n} \text{loss}(\lambda_\theta, S_i) \mathbbm{1}_{\{N_{ei} \leq s\}}$ and $E_\theta(s) = \mathbb E \left[\text{loss}(\lambda_\theta, S_{test}) \mathbbm{1}_{\{N_{e} > s\}}\right]$. 
\begin{lemma}
    For any $ s \in \mathbb N$ and nonempty parameter set $\Theta$, we have
    \begin{align}
        \mathbb P\left(\sup_{\theta \in \Theta}|X_\theta| > t \right) \leq  \mathbb P\left(\sup_{\theta \in \Theta}|X_\theta(s)| + \sup_{\theta \in \Theta}|E_\theta(s)| > t \right) + \mathbb P(N_{e(n)} > s) . 
        \label{eq:prob_decomp}
    \end{align}
\label{prob. decomposion lemma(main)}
\end{lemma}
\begin{proof}[Proof of Lemma \ref{prob. decomposion lemma(main)}]
    For $\forall \omega \in \{N_{e(n)} \leq s\}$, we have
    \begin{align}
        X_\theta(\omega) &= \mathbb E [\text{loss}(\lambda_\theta, S_{test})] - \frac{1}{n} \sum_{i = 1}^{n} \text{loss}(\lambda_\theta, S_i)(\omega) \nonumber \\
        &= \mathbb E [\text{loss}(\lambda_\theta, S_{test})] - \frac{1}{n} \sum_{k = 1}^{n} \text{loss}(\lambda_\theta, S_i) \mathbbm{1}_{\{N_{ei} \leq s\}} (\omega) \nonumber \\
        &= X_\theta(s)(\omega) + E_\theta(s)(\omega). \nonumber
    \end{align}
    Hence, under the condition $N_{e(n)} \leq s$, we have $|X_\theta| \leq |X_\theta(s)| + |E_\theta(s)|$, thus $\sup_{\theta \in \Theta} |X_\theta| \leq \sup_{\theta \in \Theta} |X_\theta(s)| + \sup_{\theta \in \Theta} |E_\theta(s)|$. Then 
    \begin{align}
        \mathbb P\left(\sup_{\theta \in \Theta}|X_\theta| > t \right)  &= \mathbb P\left(\sup_{\theta \in \Theta}|X_\theta| > t,~ N_{e(n)} \leq s\right) + \mathbb P\left(\sup_{\theta \in \Theta}|X_\theta| > t,~ N_{e(n)} > s\right) \nonumber \\
        &\leq \mathbb P\left( \sup_{\theta \in \Theta} |X_\theta(s)| + \sup_{\theta \in \Theta} |E_\theta(s)| > t,~ N_{e(n)} \leq s\right) \nonumber \\
        & ~~ + \mathbb P\left(\sup_{\theta \in \Theta}|X_\theta| > t,~ N_{e(n)} > s\right) \nonumber \\
        &\leq \mathbb P\left(\sup_{\theta \in \Theta}|X_\theta(s)| + \sup_{\theta \in \Theta}|E_\theta(s)| > t \right) + \mathbb P\left(N_{e(n)} > s\right). \nonumber
    \end{align}
\end{proof}

The consequence of this lemma is to decompose $\mathbb P\left(\sup_{\theta \in \Theta}|X_\theta| > t \right)$ into two parts. The first part $\mathbb P\left(\sup_{\theta \in \Theta}|X_\theta(s)| + \sup_{\theta \in \Theta}|E_\theta(s)| > t \right)$ is the tail probability of the supremum of a set of bounded variables, and can therefore be handled by standard empirical process theory. The second part $\mathbb P(N_{e(n)} > s)$ is the tail probability of $N_{e(n)}$. Thanks to Lemma \ref{lemma of N_e},  this term can be controlled by the exponential decay property of the sub-critical point process. 
By choosing suitable $s$, we can make \eqref{eq:prob_decomp} sharper.
This result plays a key role in stochastic error calculations.

\subsubsection{Complexity of the RNN-TPP Class}
To get the result in Theorem \ref{main variance theorem}, we need to compute the complexity of the RNN function class which is specified in section \ref{RNN function section}. There are many possible complexity measures in deep learning theory \citep{suh2024survey}, and here we choose \textit{covering number} which can be well computed for the RNN function class. 
In our setup, the key to the computation of the covering number is finding the Lipschitz continuity constant of RNN-TPPs, which separates the spectral norms of weight matrices and the total number of parameters \citep{chen2019generalization}. 

Consider two different sets of parameters $\theta_1 = \{W_{x,1}^{(l)}, W_{h,1}^{(l)}, b_1^{(l)}; 1\leq l \leq L + 1\}$, $\theta_2 = \{W_{x,2}^{(l)}, W_{h,2}^{(l)}, b_2^{(l)}; 1\leq l \leq L + 1\}$. Denote $\Delta_b^l = \|b_1^{(l)}-b_2^{(l)}\|_2$, $\Delta_h^l =  \|W_{h,1}^{(l)} - W_{h,2}^{(l)}\|_2$, $\Delta_x^l =  \|W_{x,1}^{(l)} - W_{x,2}^{(l)} \|_2$, $1\leq l \leq L+1$ $(\Delta_h^{L+1} \equiv 0)$.
The following lemma characterizes the Lipschitz constant of $\lambda_{\theta}$.
\begin{lemma}
    Under Assumptions (B1)-(B3), given an input sequence of length $N_S$, $S = \{t_i\}_{i=1}^{N_S} \subset [0,T]$ (here we set $t_{N_S+1} = T$), for $t \in (t_i, t_{i+1}]$, $1\leq i\leq N_S$, and $\theta_1, \theta_2 \in \Theta$, we have
    \begin{align}
        \left| \lambda_{\theta_1}(t;S) - \lambda_{\theta_2}(t;S) \right| &\leq  \rho_f \gamma \left(\sum_{l = 0}^{L-1} \gamma^l S_i^l \Delta_b^{L-l} +  B_{\sigma} \sqrt{D}  \sum_{l = 0}^{L-2} \gamma^l S_i^l \Delta_x^{L-l} + B_{in}(T) \gamma^{L-1} S_i^{L-1} \Delta_x^{1} \right. \nonumber \\
        &\quad \quad \quad \left. + B_{\sigma} \sqrt{D} \sum_{l = 0}^{L-1} \gamma^l S_{i-1}^l \Delta_h^{L-l} \right)  
        + \rho_f \Delta_b^{L+1} + \rho_f B_{\sigma} \sqrt{D} \Delta_x^{L+1}, 
        \label{delta_lambda(main)}
    \end{align}
    where $\beta = \rho_\sigma B_h$, $\gamma = \rho_\sigma B_x$, $S_i^l = \sum_{j=0}^{i} \tbinom{j+l}{l} \beta^j$ ($S_{-1}^l = 0$), and $d = \max \{d_l | 1\leq l \leq L+1\} $. We set $\sum_{l=a}^{b} A_l = 0$ if $a>b$.
    \label{lemma of lip continuity}
\end{lemma}
The proof of Lemma \ref{lemma of lip continuity} is based on the induction. The full proof is given in the supplementary.
Our result is an extension of Lemma 2 in \cite{chen2019generalization}, where they only consider the family of \textbf{one-layer} RNN models. 
Lemma \ref{lemma of lip continuity} is of independent interest and can be useful in any other problems regarding RNN-based modeling. 
Using Lemma \ref{lemma of lip continuity}, we can establish a covering number bound for $\mathcal{F}$ under a ``truncated" distance.

Denote $\mathcal{N}\left(\mathcal F, \epsilon, d(\cdot, \cdot)\right)$ as the covering number of metric space $\mathcal F$, i.e., the minimal cardinality of a subset $\mathcal{C} \subset \mathcal F$ that covers $\mathcal F$ in scale $\epsilon$ with respect to the metric $d(\cdot, \cdot)$.
Given a fixed integer $N_0$, We define a truncated distance,
\begin{align}
    d_{N_0}(\lambda_{\theta_1}, \lambda_{\theta_2}) = \sup_{\#(S) \leq {N_0}} \left\|\lambda_{\theta_1}(t;S) - \lambda_{\theta_2}(t;S) \right\|_{L^{\infty}[0,T]} ~. \nonumber
\end{align}
The following lemma gives an upper bound of $\mathcal{N}\left(\mathcal{F}, \epsilon, d_{N_0}(\cdot, \cdot) \right)$.
\begin{lemma}
    Under assumptions (B1)-(B3), for any $\epsilon > 0$ and $\mathcal{F} = \mathcal{F}_{L, D, B_m, l_f, u_f}$ defined as \eqref{rnn_func_class(main)}, the covering number $\mathcal{N}\left(\mathcal{F}, \epsilon, d_{N_0}(\cdot, \cdot) \right)$ is bounded by
    \begin{align}
        \mathcal{N}\left(\mathcal{F}, \epsilon, d_{N_0}(\cdot, \cdot) \right) \leq \left( 1 + \frac{C({N_0})(3L+2)B_m\sqrt{D}}{\epsilon}\right)^{D^2(3L+2)}, \nonumber
    \end{align}
    where $C(N_0) = \rho_f (B_\sigma \sqrt{D} \vee B_{in}(T) \vee 1) ( \gamma^L \vee 1) ({N_0}+1)^{L-1} (\beta^{{N_0}+1}-1)/(\beta-1) $, $\gamma = \rho_\sigma B_x$, and $\beta = \rho_\sigma B_h$.
    \label{cov number bound of F}
\end{lemma}

By Lemma \ref{cov number bound of F}, taking $N_0=s$, we can get the non-asymptotic bound of $X_\theta(s)$, which is an important step to obtain the first part of \eqref{eq:prob_decomp}.

\section{Approximation Error}
\label{approximation error section(main)}
In this section, we focus on the approximation error,  i.e., the second part of \eqref{bias-variance decomposition(main)}. 
The approximation error of deep neural networks has been broadly studied in the literature
\citep{schmidt2020nonparametric,shen2019deep,jiao2023deep,lu2021deep}. However, most of them only consider the ReLU activation case, which is different from $\tanh$, the activation function usually chosen for RNNs. 
Recently, \cite{de2021approximation} studied the approximation properties of shallow $\tanh$ neural networks, which provides a technical tool for our analysis. 
To the best of our knowledge, the approximation ability of RNN-type networks has not been fully studied in the literature. 
Here we propose a family of approximation results for the intensities of various TPP models stated in section \ref{tpp model section}.

\subsection{Poisson Case}
\label{approximation for poisson case}

We start with the the approximation of (non-homogeneous) Poisson process, whose intensity is independent of the event history, i.e. $\lambda^{\ast}(t) = \lambda_0(t)$, where $\lambda_0(t)$ is an unknown function. In this case, we do not need to take into account the transfer of information in the time domain. 
To be precise, we can take $W_h^{l} = 0$ for $l \in [L]$. Then the problem degenerates to a standard neural network approximation problem. Using the approximation results for $\tanh$ neural networks in \cite{de2021approximation}, we can get the following approximation result.

\begin{theorem}
\label{poisson thm}
    (Approximation for Poisson process) 
Under model $\lambda^{\ast}(t) = \lambda_{0}(t)$ and assumptions (A1) and (A3), for 
    $N \geq 5$,
    $N \in \mathbb N$, there exists an RNN-TPP $\hat{\lambda}^N$ as stated in section \ref{RNN function section} with $L=2$, $l_f = B_1$, $u_f = B_0$, and input function $x(t;S) = t$ such that
    \begin{align}
        |\mathbb E[\text{loss}(\hat{\lambda}^{N}, S_{test})] - \mathbb E[\text{loss}(\lambda^{\ast}, S_{test})]| 
        \leq 15 \exp \left({2 B_0 T} \right) (T + 2 B_1^{-1}) \frac{\mathcal{C} T^{s}}{N^s}, 
    \end{align}
    where $\mathcal{C} = \sqrt{2s}5^s/(s-1)!$ . Moreover, the width of $\hat{\lambda}^N$  satisfies $D \leq 3\lceil s/2\rceil + 6N$ and the weights of $\hat{\lambda}^N$ are less than 
    \begin{align}
        \mathcal{C}_1 \left[\frac{\sqrt{2s}5^s}{(s-1)!} B_0 T ^s \right]^{-\frac{s}{2}} N^{\frac{1+s^2}{2}} (s(s+2))^{3s(s+2)}, \nonumber
    \end{align}
    where $\mathcal{C}_1$ is an universal constant.
\end{theorem}

\begin{figure}
    \centering
    \includegraphics[width =  \textwidth]{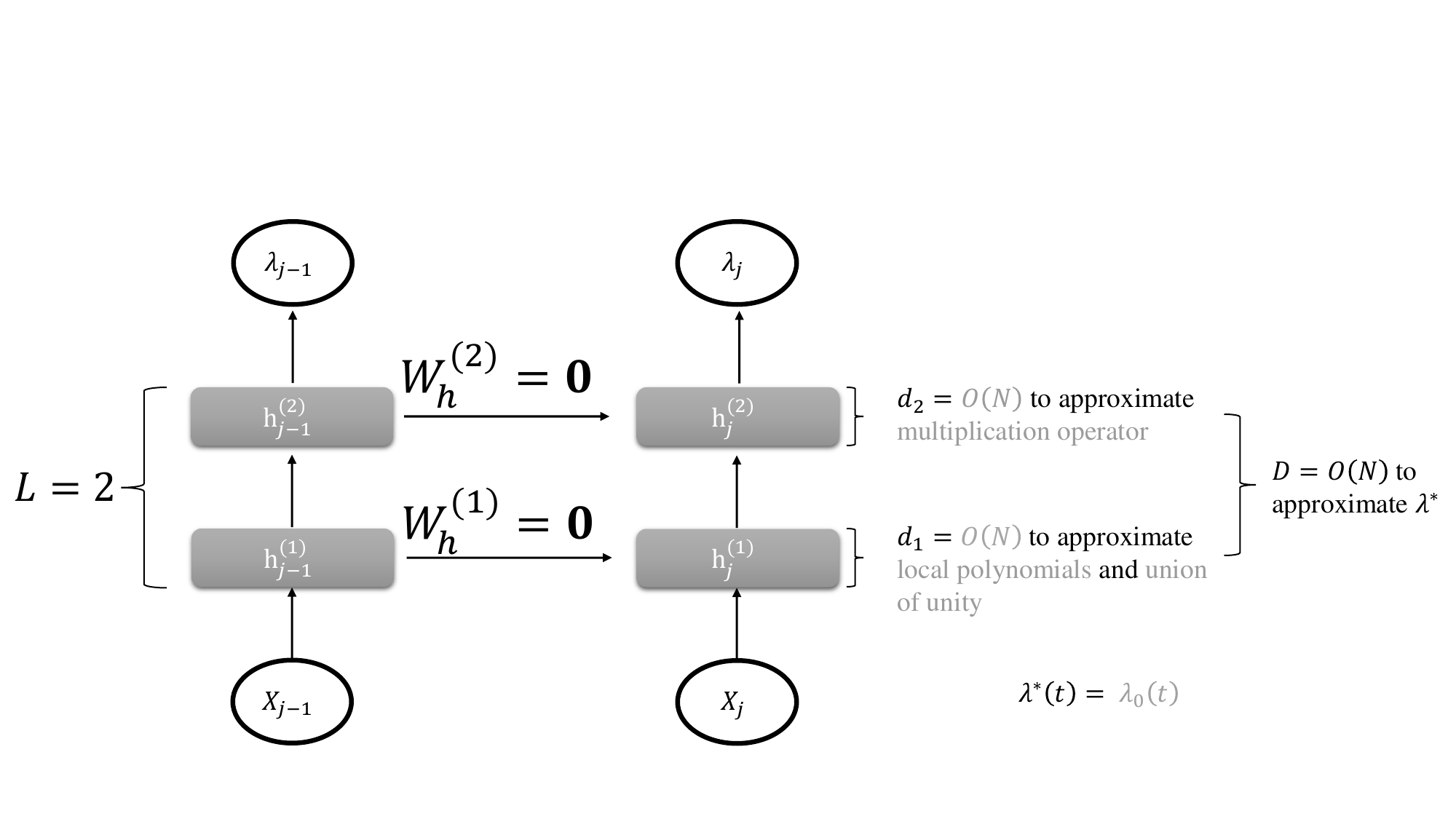}
    \caption{The construction of RNN-TPP for the case of Poisson processes.}
    \label{fig::RNN:Poisson}
\end{figure}

A graphical representation of RNN approximation is given in Figure \ref{fig::RNN:Poisson}.
For the non-homogeneous Poisson models, the RNN-TPP $\hat{\lambda}^{N}$ in Theorem \ref{poisson thm} is indeed a two-layer neural network. 
From Theorem \ref{poisson thm}, we need an RNN-TPP with $O(N)$ width and $B_m = O(N^{\frac{s^2 + 2}{2}})$ to obtain $O(N^{-s})$ approximation error. Combining with Theorem \ref{main variance theorem}, we can get the part (i) of Theorem \ref{main theorem}.

\subsection{Vanilla Hawkes Case}
\label{approximation for vanilla hawkes case}
\begin{figure}
    \centering
    \includegraphics[width =  \textwidth]{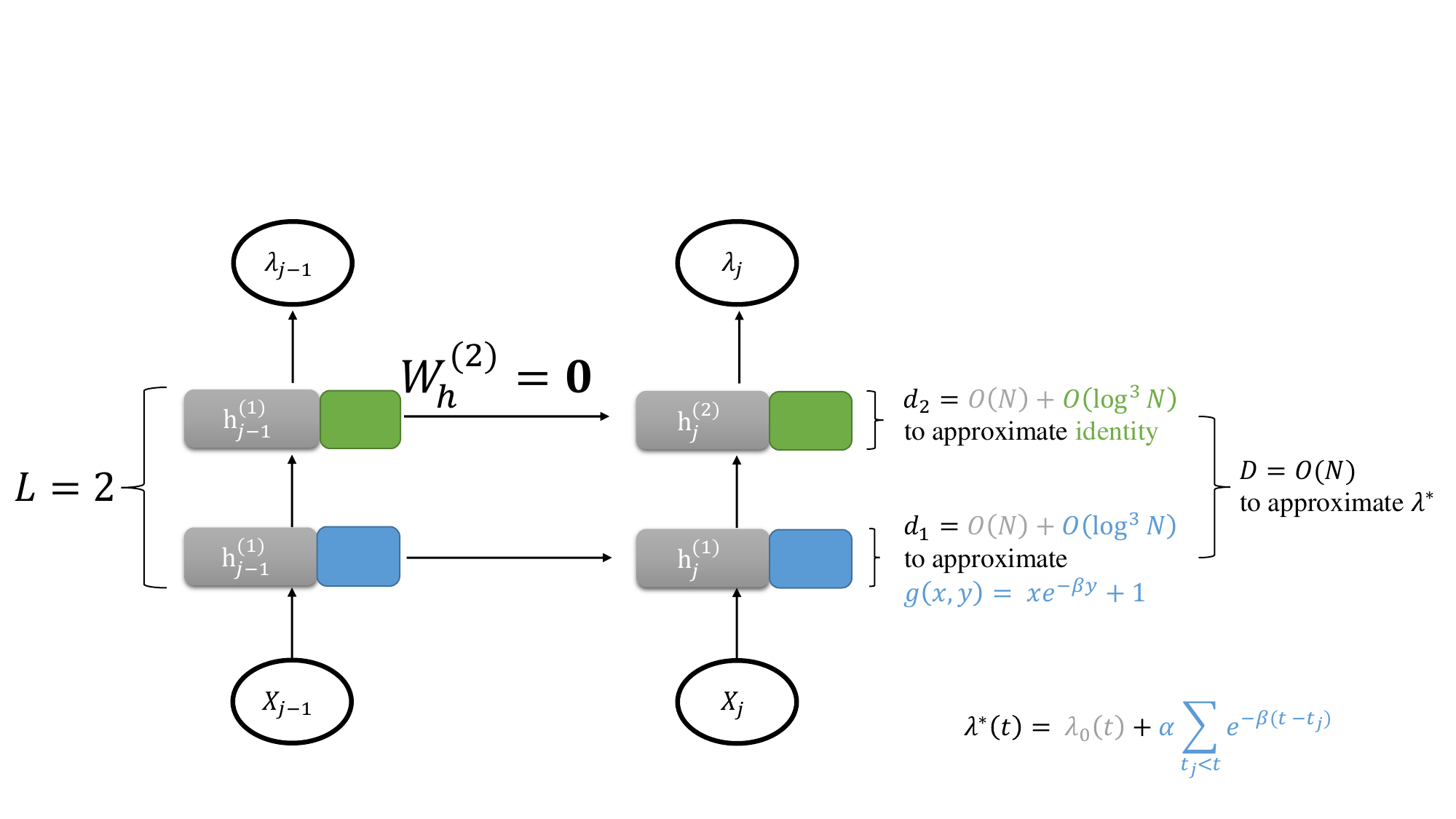}
    \caption{The construction of RNN-TPP for the case of the vanilla Hawkes process.}
    \label{fig::RNN:Vanilla Hawkes}
\end{figure}

Recall that the intensity of the vanilla Hawkes process has the form
\begin{align}
    \lambda^{\ast}(t) = \lambda_0(t) + \sum_{j:t_j < t} \alpha \exp\{-\beta (t - t_j)\} . 
    \label{vanilla hawkes intensity}
\end{align}
Different from Poisson process, the intensity of the vanilla Hawkes process depends on historical events. Hence it can not be approximated by a simple neural network and needs the recurrent structure. We construct an RNN-TPP to approximate the intensity using the Markov property of \eqref{vanilla hawkes intensity}. Specifically, note that if we have observed the first $k$ event times $\{t_1, \cdots, t_k\}$, then for any $t$ satisfying $t_k < t \leq t_{k+1}$, we have
\begin{align}
    \lambda^{\ast}(t) - \lambda_0(t)&= \sum_{j:t_j < t} \alpha \exp\{-\beta (t - t_j)\} \nonumber \\
    &=  \exp(-\beta(t-t_k)) \sum_{j:t_j \leq t_k} \alpha \exp\{-\beta (t_k - t_j)\} \nonumber\\
    &=  (\lambda^{\ast}(t_k)  - \lambda_0(t_k) +  \alpha)\exp(-\beta(t-t_k)). \nonumber
\end{align}
Therefore, we can use the hidden layers in RNN-TPP to store the information of $\lambda^{\ast}(t_k)  - \lambda_0(t_k)$ and then compute $\lambda^{\ast}(t) - \lambda_0(t)$ with the help of input $t-t_k$. 
Together with the approximation of $\lambda_0$, we can obtain the final approximation result.
A graphical illustration of the above construction procedures are given in Figure \ref{fig::RNN:Vanilla Hawkes}.

\begin{theorem}
\label{hawkes thm}
    (Approximation for Vanilla Hawkes process) Under model \eqref{vanilla hawkes intensity}, assumptions (A1), (A3), and $\alpha / \beta < 1 $, for $N \geq 5$, $N \in \mathbb N$, there exists an RNN-TPP $\hat{\lambda}^N$ as stated in section \ref{RNN function section} with $L = 2$, $l_f = B_1$, $u_f = B_0 + O(\log N)$, and input function $x(t;S) = (t, t-F_S(t))^\top$ such that
    \begin{align}
         |\mathbb E[\text{loss}(\hat{\lambda}^{N}, S_{test})] - \mathbb E[\text{loss}(\lambda^{\ast}, S_{test})]| \lesssim  \frac{(\log N)^2}{N^s}  .
    \end{align}
    Moreover, the width of $\hat{\lambda}^N$  satisfies $D = O(N) $ and the weights of $\hat{\lambda}^N$ are less than 
    \begin{align}
    \mathcal{C}_1 (\log(N))^{12s^2(\log(N))^2}~, \nonumber
    \end{align}
    where $\mathcal{C}_1$ is a constant related to $s, B_0, \beta$, and $T$.
\end{theorem}

Due to the smoothness of the exponential function, the approximation rate in Theorem \ref{hawkes thm} only adds the $\log(N)$ term compared with the results in Theorem \ref{poisson thm}. Similarly, combining with Theorem \ref{main variance theorem}, we can easily get the part (ii) of Theorem \ref{main theorem}.

\subsection{Linear Hawkes Case}
\label{approximation for general case}
\begin{figure}
    \centering
    \includegraphics[width =  \textwidth]{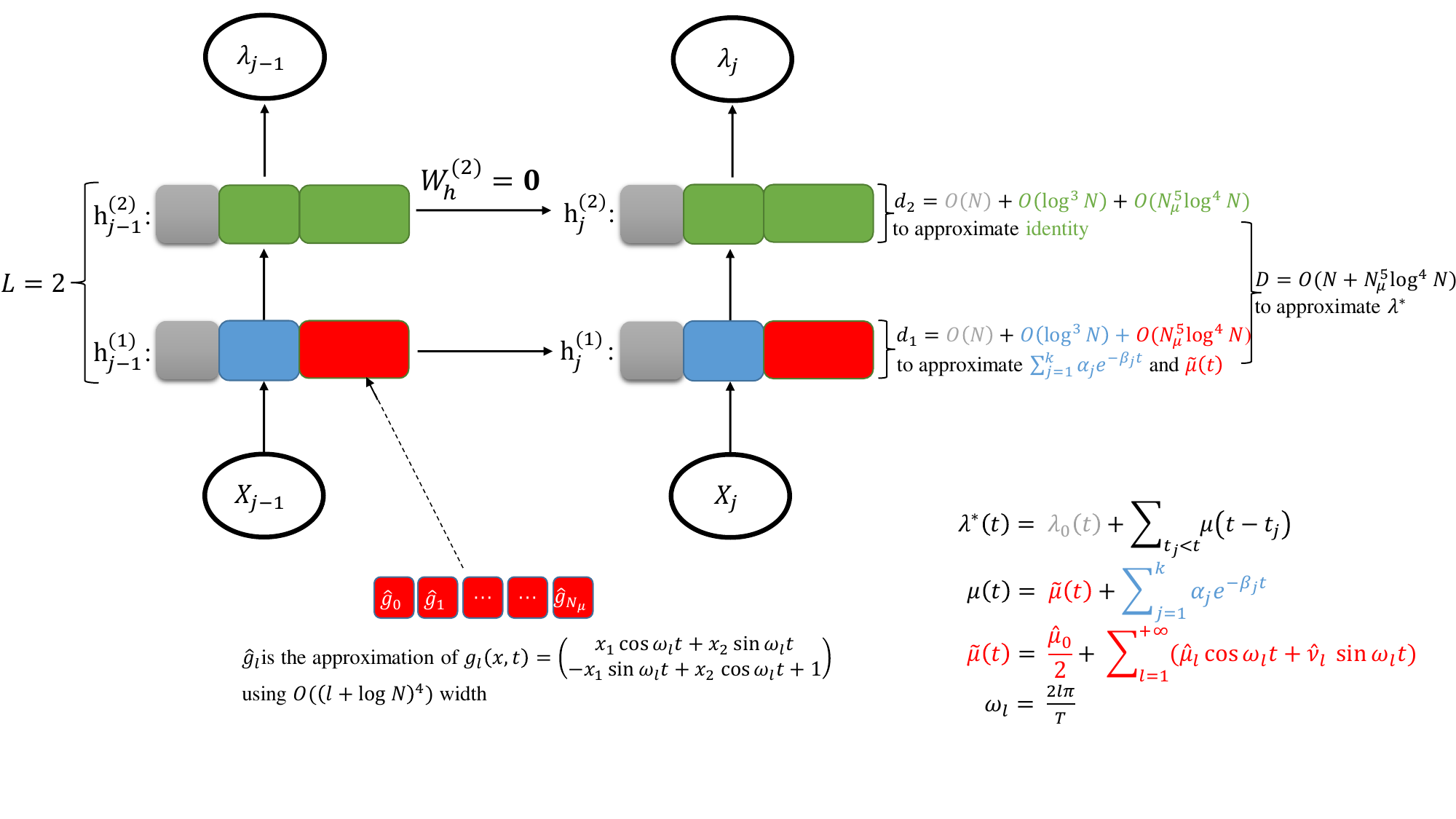}
    \caption{The construction of RNN-TPP for the case of general linear Hawkes processes.}
    \label{fig::RNN:linear Hawkes}
\end{figure}

Now we consider the general linear Hawkes process, i.e., \eqref{eq:label:hawkes} in section \ref{tpp model section}. Motivated by the approximation construction of the Vanilla Hawkes process, we want to find a decomposition for the general $\mu$ where each term has the 'Markov property' so that we can construct the corresponding RNN structure. Precisely, for $\mu \in C^{k,\infty}([0,T], C_0)$, $k \geq 2$, $k \in \mathbb N$, we can decompose $\mu$ into two parts,
\begin{align}
    \mu(t) = \underbrace{\tilde{\mu}(t)}_{\text{part}_1} + \underbrace{\sum_{j=1}^{k} \alpha_j \exp(-\beta_j t)}_{\text{part}_2}, ~~ t \in [0,T], \nonumber
\end{align}
where $\tilde{\mu}$ satisfies the \textit{boundary condition}, $\tilde{\mu}^{j}(0+) = \tilde{\mu}^{j}(T-)$, $0\leq j \leq k-1$, $j \in \mathbb N$, and $\beta_j = j/k$, $j \in [k]$. The term $\sum_{j=1}^{k} \alpha_j \exp(-\beta_j t)$ can be handled similarly to the vanilla Hawkes process. For $\tilde{\mu}$, we consider its Fourier expansion, 
\[ \tilde{\mu}(t) = \frac{\hat{\mu}_0}{2} + \sum_{l = 1}^{\infty} \left(\hat{\mu}_l \cos\Big(\frac{2l\pi}{T}t\Big) + \hat{\nu}_l \sin\Big(\frac{2l\pi}{T}t\Big) \right).\] 
Thanks to the boundary condition, $\tilde{\mu}(t)$ can be well approximated by the \textbf{finite} sum of Fourier series. 
Then we can use the ``Markov property" of the trigonometric function pairs $\cos(2l\pi t/T)$ and $\sin(2l\pi t/T)$ to construct the RNN-TPP. 
The construction is similar to that of the exponential function case but needs more thorny calculations. Combining all the approximation parts, we can get the approximation theorem for \eqref{eq:label:hawkes}.
The above ideas are visualized in Figure \ref{fig::RNN:linear Hawkes}.

\begin{theorem}
\label{linear hawkes thm}
    (Approximation for linear Hawkes process) Under model \eqref{eq:label:hawkes}, assumptions (A1)-(A3), and $\mu \in C^{k,\infty}([0,T], C_0)$, $k \geq 2$, $k \in \mathbb N$, for $N \geq 5$, $N \in \mathbb N$, there exists an RNN-TPP $\hat{\lambda}^{N,N_\mu}$ as stated in section \ref{RNN function section} with $L = 2$, $l_f = B_1$, $u_f = B_0 + O(\log N)$, and input function $x(t;S) = (t, t-F_S(t))^\top$ such that
    \begin{align}
       |\mathbb E[\text{loss}(\hat{\lambda}^{N, N_\mu}, S_{test})] - \mathbb E[\text{loss}(\lambda^{\ast}, S_{test})]| \lesssim  \frac{ (\log N)^2}{N^s} + \frac{ \log N}{N_\mu^{k-1}}  .
    \end{align}
    Moreover, the width of $\hat{\lambda}^{N,N_\mu}$  satisfies $D = O(N  +N_\mu^5 (\log N)^4)$ and the weights of $\hat{\lambda}^{N,N_\mu}$ are less than 
    \begin{align}
    \mathcal{C}_1 (\log(N N_\mu))^{12s^2(\log(N N_\mu))^2}~, \nonumber
    \end{align}
    where $\mathcal{C}_1$ is a constant related to $s, k, B_0, C_0, c_\mu$, and $T$.
\end{theorem}
We make a few explanations on Theorem \ref{linear hawkes thm}. There are two tuning parameters in $\hat{\lambda}^{N,N_\mu}$, where $N$ is the tuning parameter to control the approximation error of $\lambda_0$, $\sum_{j=1}^{k} \alpha_j \exp(-\beta_j t)$, and the finite sum of the Fourier series, and $N_\mu$ is the tuning parameter to control the number of terms in the Fourier series entering the RNN-TPP. 
The term $(\log N)^2/N^s$ is obtained similarly to that in the vanilla Hawkes process case, and the term $\log N/N_\mu^{k-1}$ is the error caused by the finite sum approximation for the Fourier series. Moreover, the $O(N_\mu^5 (\log N)^4)$ term in the width of RNN-TPP is caused by the approximation construction of the first $N_\mu$ terms of the Fourier series. Finally, combining with Theorem \ref{main variance theorem}, we can obtain the part (iii) of Theorem \ref{main theorem}.

\subsection{Nonlinear Hawkes Case}
\label{approximation for nonlinear hawkes case}
\begin{figure}
    \centering
    \includegraphics[width =  \textwidth]{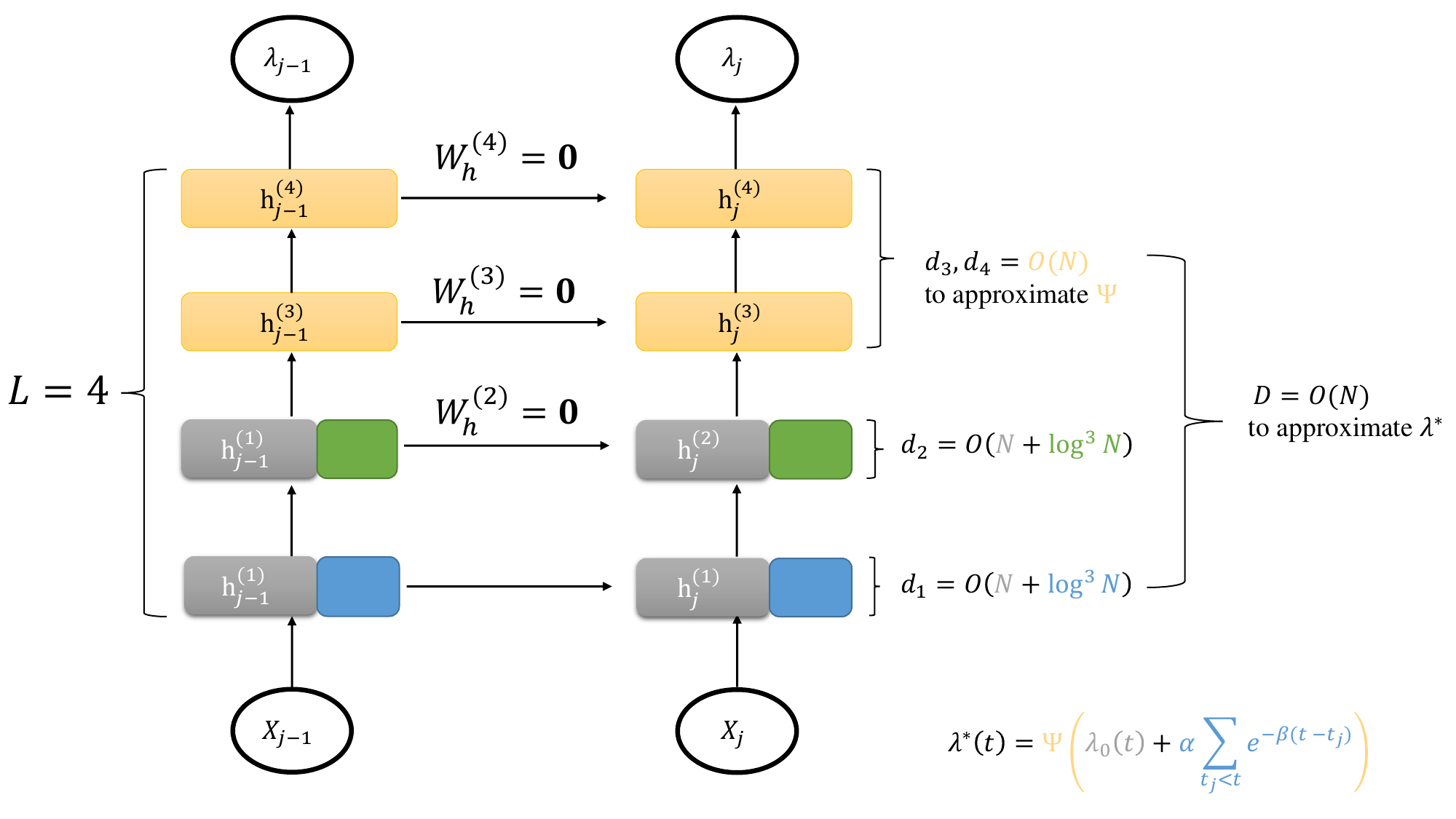}
    \caption{The construction of RNN-TPP for the case of nonlinear Hawkes process.}
    \label{fig::RNN:nonlinear Hawkes}
\end{figure}

Finally, we consider the nonlinear Hawkes process, which is defined in \eqref{eq:label:non:linear} in section \ref{tpp model section}. 
To make the statement simpler, we only consider the simple case, i.e., $\mu(t) = \alpha \exp(-\beta t)$. The results for the general $\mu$ can be obtained similarly. Compared to the vanilla Hawkes case, the additional challenge here is the existence of a nonlinear function $\Phi$. With two additional layers, we can approximate $\Phi$ well. Together with the results for the case of the vanilla Hawkes process, we can obtain the desired RNN-TPP architecture. 
To be clearer, we also provide the graphical illustration in Figure \ref{fig::RNN:nonlinear Hawkes}.

\begin{theorem}
\label{nonlinear (vanilla)hawkes thm}
    (Approximation for nonlinear Hawkes process) Under model \eqref{non linear intensity(main)}, assumptions (A1) and (A4), for $N \geq \max\{5, (2\mathcal{C} B_0 T^s + 1)^{\frac{1}{s}}\}$ with $\mathcal{C} = \sqrt{2s}5^s/(s-1)!$, there exists an RNN-TPP $\hat{\lambda}^N$ as stated in section \ref{RNN function section} with $L = 4$, $l_f = \tilde{B}_1$, $u_f = \tilde{B}_0$, and input function $x(t;S) = (t, t-F_S(t))^\top$ such that
    \begin{align}
         |\mathbb E[\text{loss}(\hat{\lambda}^{N}, S_{test})] - \mathbb E[\text{loss}(\lambda^{\ast}, S_{test})]| \lesssim \frac{\log N}{N}  .
    \end{align}
    Moreover, the width of $\hat{\lambda}^N$  satisfies $D =O(N) $ and the weights of $\hat{\lambda}^N$ are less than 
    \begin{align}
    \mathcal{C}_1 (\log N)^{12s^2(\log N)^2}~, \nonumber
    \end{align}
    where $\mathcal{C}_1$ is a constant related to $s, \tilde{B}_0, \alpha , \beta, T$, and $L$.
\end{theorem}
Since we assume $\Phi$ to be Lipschitz continuous, we can only get $\tilde{O}(N^{-1})$ approximation error. 
The rate can be improved if $\Phi$ is allowed to have better smoothness properties. 
Again, combining with Theorem \ref{main variance theorem}, we arrive at Theorem \ref{non linear case thm}.

\begin{remark}
    The universal approximation properties of one-layer RNNs are studied in \cite{schafer2007recurrent}.
    Our current results are different from theirs in the following sense. 
    (i) RNN-TPP is defined over the continuous time domain $[0,T]$, while the standard RNN only considers the discrete points. In other words, our approximation results hold uniformly over all $t \in [0,T]$.
    (ii) In \cite{schafer2007recurrent}, they do not give the explicit formula of the widths of hidden layers or parameter weights in the construction of RNN approximator. Therefore, their results cannot be directly used in computing the approximation error. 
\end{remark}

\section{Usefulness of Interpolation of Hidden States}
\label{counterexample section}

As mentioned in Remark 3, the RNN-TPP needs to take into account any continuous time point $t$ between observed time grids $t_j$ and $t_{j+1}$. 
The interpolation of hidden state $h^{(l)}(t; S)$ between $h_j^{(l)}$ and $h_{j+1}^{(l)}$ is essential and important during the construction of RNN-TPPs. 

In this section, we give a counter-example to illustrate that an RNN-TPP model with a linear interpolation of hidden states is unable to precisely capture the true intensity in terms of excess risk $\eqref{eq:def:gen:err}$. 
For simplicity, we only consider the single-layer RNN-TPP and the argument is the same for multi-layer RNN-TPPs.

We consider a (single-layer) RNN-TPP which admits the following model structure,
\begin{align}
    h_j &= \sigma(W_x x(t_j; S) + W_h h_{j-1}), \nonumber \\
    \hat{\lambda}_{ne}(t) &= f(\alpha (t - t_j) + W_y h_j + b) \in \mathbb R, ~~ t \in (t_j, t_{j+1}], \label{RNN_naive(main)}
\end{align}
where $x(t_j; S)$ is the embedding for the $j$-th event, $h_0 = \mathbf 0$, $\sigma(x) = \tanh(x)$,  $f(x) = (x \vee l_f) \wedge u_f$, and $l_f$ and $u_f$ will be determined from the true intensity. If we take $\alpha = 0$, it would be the same as the model using a constant hidden state interpolation mechanism, such as \cite{du2016recurrent}.
In other words, $h^{(1)}(t;S) \equiv h_j$ for all $t$ satisfying $t_j \leq t < t_{j+1}$ when $\alpha = 0$.

\begin{theorem}
    Suppose the true model intensity on $[0,T]$ has the following form,
    \begin{eqnarray}
        \lambda^{\ast}(t) &=& \left\{
    \begin{aligned}
        &T &, ~ &t \in [0, T/3] \\
        &\frac{9}{T}t^2 &,  ~ &t \in (T/3, 2T/3)\\
        &4T &, ~ &t \in [2T/3, T]
        \nonumber
    \end{aligned}
    \right. \quad .
    \end{eqnarray}
    Hence we can take $l_f = T$ and  $u_f = 4T$, and then there exists a constant $C>0$ such that
    \begin{align}
           \min_{\hat{\lambda}_{ne} \text{ as } \eqref{RNN_naive(main)}} \mathbb E [\text{loss}(\hat{\lambda}_{ne}, S_{test})] - \mathbb E[\text{loss}(\lambda^{\ast}, S_{test})] \geq C > 0.
            \label{eq:nonzero-bias}
    \end{align}
    \label{thm:counterexample}
\end{theorem}

Theorem \ref{thm:counterexample} tells us that the RNN-TPPs with an improper hidden state interpolation may fail to offer a good approximation, even under a very simple non-homogeneous Poisson model.    
Therefore, the user-determined input embedding vector function $x(t; S)$ plays an important role in interpolating the hidden states. 
It should be carefully chosen so that $x(t; S)$ can summarize the information of past event history to some extent. 
\begin{remark}
One can substitute the linear interpolation mechanism \eqref{RNN_naive(main)} with the exponential decaying mechanism given in \cite{mei2017neural}. 
Theorem \ref{thm:counterexample} still holds.
\end{remark}
\begin{remark}
For other different types of $f$ (e.g. Softplus) in the output layer, the failure of the linear interpolation mechanism can be obtained similarly.
\end{remark}

\section{Discussion} \label{sec:discuss}
In this paper, we give a positive answer to the question "whether the RNN-TPPs can provably have small excess risks in the estimation of the well-known TPPs". 
We establish the excess risk bounds under homogeneous Poisson process, non-homogeneous Poisson process, self-exciting process, and self-correcting process framework.
Our analysis focuses on two parts, the stochastic error and the
approximation error. 
For the stochastic error, we use a novel truncation technique to decouple the randomness and make the classical empirical process theory applicable.
We carefully compute the Lipschitz constant of multi-layer RNNs, which is a useful intermediate result for future RNN-related work.
For approximation error, we construct a series of RNNs to approximate the intensities of different TPPs by providing the explicit network depth, width, and parameter weights. 
To the best of our knowledge, our work is the first one to study the approximation ability of the multi-layer RNNs over the continuous time domain.
We believe the results in the current work add values to both learning theory and neural network fields.

There are several possible extensions along the research line of neural network-based TPPs. 
First, it is not clear whether the approximation rate can be improved by a more refined RNN structure construction (with possible fewer layers and smaller width) or other possible approaches. 
Second, we here only consider the ``large $n$" setting where the event sequences are observed in a bounded time domain $[0,T]$ with $n$ repeated samples. It is interesting to extend our results to ``large $T$" setting where the end time $T$ goes to infinity but the number of event sequences, $n$, remains fixed. 
Third, in the current work, we do not take into account the different event types. It may be useful to extend our results to the marked TPP settings. 
Moreover, it is also worth investigating the theoretical performances of other neural network architectures (e.g. Transformer-TPPs) that have performed well in recent empirical applications.

\newpage

\begin{center}
\textbf{Supplementary Material for "On Non-asymptotic Theory of Recurrent Neural Networks in Temporal Point Processes"}
\end{center}

\textbf{Additional Notations in the Supplementary:} For two random variables $X$ and $Y$, we write $X \leq_{s.t.} Y$ if $\mathbb P (X > t) \leq \mathbb P (Y > t)$ for any $t\in \mathbb R$. 
Use $\mathbb N_{+}$ to denote the set of positive integers.

\section{Proofs in section 3 and 4}

\subsection{Proof of Lemma 1}
By the definition of $\check \lambda^{\ast}$ and $\hat{\lambda}$, we have
\begin{align}
    &\quad\mathbb E[\text{loss}(\hat \lambda, S_{test})] - \mathbb E[\text{loss}(\lambda^{\ast}, S_{test})] ]\nonumber \\
    &=
    \mathbb E[\text{loss}(\hat \lambda, S_{test})] - \mathbb E[\text{loss}(\check \lambda^{\ast}, S_{test})] + \mathbb E[\text{loss}(\check \lambda^{\ast}, S_{test})] - \mathbb E[\text{loss}(\lambda^{\ast}, S_{test})]  \nonumber\\
    &\leq \mathbb E[\text{loss}(\hat \lambda, S_{test})] \underbrace{- \frac{1}{n} \sum_{i\in [n]} \text{loss}(\hat\lambda,S_i) + \frac{1}{n} \sum_{i\in [n]} \text{loss}(\check \lambda^{\ast},S_i)}_{\geq 0} - \mathbb E[\text{loss}(\check \lambda^{\ast}, S_{test})] \nonumber\\
    & \quad+ \mathbb E[\text{loss}(\check \lambda^{\ast}, S_{test})] - \mathbb E[\text{loss}(\lambda^{\ast}, S_{test})]   \nonumber \\
    &\leq 2\sup_{\lambda \in \mathcal{F}} \Big|\mathbb E[\text{loss}(\lambda, S_{test})] - \frac{1}{n} \sum_{i\in [n]} \text{loss}(\lambda,S_i)\Big| \nonumber\\
    &\quad+ \mathbb E[\text{loss}(\check \lambda^{\ast}, S_{test})] - \mathbb E[\text{loss}(\lambda^{\ast}, S_{test})]. \nonumber
\end{align}

\subsection{Proof of Lemma 2}
   From 
   model assumptions (A1) and (A2), we have $\lambda^{*}(t) = \lambda_0(t) + \sum_{j: t_j < t} \mu(t - t_j)$, ~ $\int_{0}^{T} \mu(t) \mathrm{d}t \leq c_\mu < 1$, $\lambda_0(t) \leq B_0$. Following the notations in the paper, we denote $N_e$ as the number of event time of $\lambda^{*}$ in $[0,T]$. Consider another density $\overline{\lambda}(t) = B_0 + \sum_{j: t_j < t}  \mu(t - t_j)$ and similarly denote $\overline{N}_e$ as the number of event time of $\overline{\lambda}$ in $[0,T]$. Then for any fixed event sequence $S = \{t_j\}$, $\lambda^{*}(t; S) \leq \overline{\lambda}(t; S)$, and thus $N_e \leq_{s.t.} \overline{N}_e$. By a similar formulation in \cite{daley2003introduction}, the point process with intensity $\overline{\lambda}$ is equivalent to a birth-immigration process with immigration intensity $c$ and birth intensity $\mu(t)$. Hence
    \begin{align}
        \overline{N}_e = \overline{N}_0 + \sum_{i=1}^{\infty} \overline{N}_i, \nonumber
    \end{align}
    where $\overline{N}_0 \sim \operatorname{Poisson}(B_0 T)$ and $\overline{N}_k$ is the number of event time in generation $k$, which are children of generation $k-1$.

    For $t_1 < t_2$, let $\mu_{t_1}^{t_2} = \int_{t_1}^{t_2} \mu(t-t_1) \mathrm{d}t$. We have
    \begin{align}
        \mathbb E \left[\exp\left(s\overline{N}_0\right)\right] = \exp\left(B_0 T\left(\exp(s)-1\right)\right), \nonumber
    \end{align}
    and
    \begin{align}
        \mathbb E \left[ \exp\left(s\overline{N}_{k+1}\right)\right] &= \mathbb E \left[ \mathbb E \left[\exp\left(s\overline{N}_{k+1}\right) \Big| \left\{t_j^{(k)}\right\}_{j=1}^{\overline{N}_{k}}  \right] \right] \nonumber \\
        & = \mathbb E \left[ \prod_{j=1}^{\overline{N}_{k}} \exp\left(\mu_{t_j^{(k)}}^T \left(\exp(s) - 1\right)\right) \right] \nonumber \\
        & \leq \mathbb E\left[ \exp\left(c_\mu \overline{N}_{k} \left(\exp(s) - 1 \right)\right)\right], \nonumber
    \end{align}
    for any $s > 0$.
    Since $c_\mu < 1$, for any fixed $c_1\in (c_\mu, 1]$ and any $s \in (0, \log(c_1/c_\mu)]$, we have
    \begin{align}
        \mathbb E \left[\exp\left(s\overline{N}_{k}\right)\right] \leq \mathbb E\left[ \exp\left(c_\mu \overline{N}_{k-1} \left(\exp(s) - 1 \right)\right)\right] \leq {\mathbb E}\left[\exp\left(c_1 s \overline{N}_{k-1}\right)\right] \leq \cdots \leq {\mathbb E}\left[\exp\left(c_1^k s \overline{N}_{0}\right)\right], \nonumber
    \end{align}
    \emph{i.e.}
    \begin{align}
        \mathbb E \left[ \exp\left(s\overline{N}_{k}\right)\right] \leq \mathbb E\left[\exp\left(c_1^k s \overline{N}_{0}\right)\right] = \exp\left( B_0 T \left(\exp\left(c_1^k s\right) -1\right)\right) \leq \exp\left(\frac{c_1^{k+1}}{c_\mu}(B_0 T)s\right) \nonumber
    \end{align}
    for any $k \in \mathbb N$.
    
    Since $\overline{N}_{k}(T)$ can only take integer values, we can get $\mathbb P(\overline{N}_{k} = 0) + e^s \mathbb P(\overline{N}_{k} \neq 0) \leq \mathbb E \left[ \exp(s\overline{N}_{k})\right]$. Thus
    \begin{align}
        \mathbb P\left(\overline{N}_{k} \neq 0\right) \leq \frac{\mathbb E \left[ \exp\left(s\overline{N}_{k}\right)\right] - 1}{\exp(s) - 1}  \leq \frac{c_1^{k+2}}{c_\mu^2}(B_0 T), ~ \forall s \in \left(0, \min\left\{\frac{c_\mu}{c_1^{k+1}(B_0 T)},1\right\}\log\left(\frac{c_1}{c_\mu}\right)\right]. \nonumber
    \end{align}
    Setting $c_1 \searrow c_\mu$, we get
    \begin{align}
         \mathbb P\left(\overline{N}_{k} \neq 0\right) \leq c_\mu^{k} (B_0 T). \nonumber
    \end{align}
    Now take $c_1 \in (c_\mu, 1)$, and then $c_1^{-1}(1-c_1)\sum_{k=1}^{\infty}c_1^{k} = 1$. By Bool's inequality, we have
    \begin{align}
        \mathbb P \left(\sum_{k=0}^{\infty}\overline{N}_{k} \geq N\right) &\leq \sum_{k=0}^{\infty} \mathbb P\left(\overline{N}_{k} \geq \frac{1-c_1}{c_1} c_1^{k+1} N \right) \nonumber \\
        & \leq \sum_{k=0}^{K_0 - 1} \mathbb P\left(\overline{N}_{k} \geq \frac{1-c_1}{c_1} c_1^{k+1} N \right) + \sum_{k=K_0}^{\infty} \mathbb P\left(\overline{N}_{k} \neq 0\right) \nonumber .
    \end{align}
    For the second term, $\sum_{k=K_0}^{\infty} \mathbb P\left(\overline{N}_{k} \neq 0\right) \leq \sum_{k=K_0}^{\infty} c_\mu^{k} (B_0 T) = c_\mu^{K_0}(B_0 T)/(1-c_\mu)$. Let $c_\mu^{K_0}(B_0 T)/(1-c_\mu) \leq \delta/2n$. It can be showed 
    \begin{align}
        K_0 \geq \frac{\log\left( 2nB_0 T/[\delta(1-c_\mu)]\right)}{\log\left(1/c_\mu\right)} . \nonumber
    \end{align}
    For the first term, we have
    \begin{align}
        \sum_{k=0}^{K_0 - 1} \mathbb P\left(\overline{N}_{k} \geq \frac{1-c_1}{c_1} c_1^{k+1} N \right) &\leq \sum_{k=0}^{K_0 - 1} \exp\left(-s \left(\frac{1-c_1}{c_1} c_1^{k+1} N\right)\right) \mathbb E \left[ \exp(s\overline{N}_{k})\right] \nonumber \\
        & \leq \sum_{k=0}^{K_0 - 1} \exp\left(c_1^{k+1} s\left(\frac{B_0 T}{c_\mu} - \frac{1-c_1}{c_1}  N \right) \right), \nonumber
    \end{align}
    where $s \in (0, \log(c_1/c_\mu)]$. We can take $c_1 s \left(B_0 T/c_\mu - (1-c_1)N/c_1\right) \leq \log(\delta/(2nK_0))$ so that 
    $\sum_{k=0}^{K_0 - 1} \exp\left(c_1^{k+1} s\left(B_0 T/c_\mu - (1-c_1)N/c_1 \right) \right) \leq \delta/(2n)$. Then  
    \begin{align}
        N \geq \frac{1}{1-c_1}\left( \frac{1}{s} \log\left(\frac{2nK_0}{\delta}\right) + \frac{c_1}{c_\mu}(B_0 T) \right). \nonumber
    \end{align}
    Now let $\eta = c_1/c_\mu \in (1, 1/c_\mu)$, $s = \log(c_1/c_\mu) = \log(\eta)$, and $N \geq \left[\log\left(2nK_0/\delta\right)/\log(\eta) + \eta(B_0 T) \right]/(1-c_\mu \eta)$. Taking $K_0 = \lceil \log\left(2nB_0 T/[\delta(1-c_\mu)]\right)/\log\left(1/c_\mu\right) \rceil$ and \\
    $N = \left[ \log\left(2nK_0/\delta\right)/\log(\eta) + \eta(B_0 T) \right]/(1-c_\mu \eta)$, we have $\mathbb P \left( N_e \geq N\right)\leq \mathbb P \left( \overline{N}_e \geq N\right) \leq \delta/n$.
    Since
    \begin{align}
        \mathbb P(N_{e(n)} \geq N) = 1 - \mathbb P(N_{e(n)} < N) = 1 - \prod_{i = 1}^{n} \mathbb P\left(N_{e} < N \right) \leq 1 - \left(1-\frac{\delta}{n}\right)^{n} \leq \delta , \nonumber
    \end{align}
    we get that with probability at least $1-\delta$, 
    \begin{align}
        N_{e(n)} < N \leq \frac{1}{1-c_\mu \eta} \left[ \frac{1}{\log(\eta)} \log\left(\frac{2nK_{n,\delta}}{\delta}\right) + \eta(B_0 T) \right], \nonumber
    \end{align}
    where $\eta \in (1, 1/c_\mu)$, and $K_{n,\delta} = \log\left(2nB_0 T/\delta(1-c_\mu)\right)/\log\left(1/c_\mu\right) + 1$.
    Since $1 - 1/x \leq \log(x) \leq x - 1$, we have $K_{n,\delta} \leq 2nB_0 T/[\delta(1-c_\mu)^2]$. Thus with probability at least $1-\delta$,
    \begin{align}
        N_{e(n)} <  \frac{1}{1-c_\mu \eta} \left[ \frac{2}{\log(\eta)} \log\left(\frac{2n\sqrt{B_0 T}}{\delta(1-c_\mu)} \right) + \eta(B_0 T) \right]. \nonumber
    \end{align}
    Taking $n = 1$ and $2\log\left(2\sqrt{B_0 T}/[\delta(1-c_\mu)] \right)/\log(\eta) + \eta(B_0 T) /(1-c_\mu \eta) = s$, we have 
    $\delta = 2\sqrt{B_0 T} \exp\left(\log(\eta) \left[\eta(B_0 T) - (1-c_\mu\eta)s\right]/2 \right)/(1-c_\mu)$. Then
    \begin{align}
        \mathbb P\left(N_e = s\right) \leq \mathbb P\left(N_e \geq s\right) \leq \frac{2\sqrt{B_0 T}}{1-c_\mu} \exp\left( \frac{\log(\eta)}{2} \left[\eta(B_0 T) - (1-c_\mu\eta)s\right] \right). \nonumber
    \end{align}

\subsection{Proof of Lemma 4}

    The proof is based on induction. Using the same notation, we give two claims.
    \begin{claim} \label{claim1}
    For $\forall 1 \leq l \leq L$, $1 \leq i \leq N$, $\|h_{i,1}^{(l)} - h_{i,2}^{(l)}\|_2$ is bounded by 
    {\small
    \begin{align}
        \left\|h_{i,1}^{(l)} - h_{i,2}^{(l)}\right\|_2 \leq \rho_\sigma \left(\sum_{r = 0}^{l-1} \gamma^r S_{i-1}^r \Delta_b^{l-r} + B_{\sigma} \sqrt{D}  \sum_{r = 0}^{l-2} \gamma^r S_{i-1}^r \Delta_x^{l-r} + B_{in}(T) \gamma^{l-1} S_{i-1}^{l-1} \Delta_x^{1} + B_{\sigma} \sqrt{D} \sum_{r = 0}^{l-1} \gamma^r S_{i-2}^r \Delta_h^{l-r} \right).  
        \label{delta_h_i(main)}
    \end{align}
    }
    \end{claim}
    \begin{proof}[Proof of Claim \ref{claim1}]
        When $i = 1$, we have
        \begin{align}
            \left\|h_{1,1}^{(l)} - h_{1,2}^{(l)}\right\|_2  &= \left\|\sigma\left( W_{x,1}^{(l)} h_1^{(l-1)} + b_1^{(l)} \right) - \sigma\left( W_{x,2}^{(l)} h_2^{(l-1)} + b_2^{(l)} \right)\right\|_2 \nonumber \\
            &\leq \rho_\sigma \left( \left\|W_{x,1}^{(l)} h_1^{(l-1)} -  W_{x,2}^{(l)} h_2^{(l-1)}\right\|_2 + \left\|b_1^{(l)} - b_2^{(l)}\right\|_2  \right) \nonumber \\
            &\leq \rho_\sigma \left(B_\sigma \sqrt{D} \Delta_x^l + B_x \left\|h_{1,1}^{(l-1)} - h_{1,2}^{(l-1)}\right\|_2 + \Delta_b^l \right). \nonumber
        \end{align}
        Repeat this derivation recursively, we get
        \begin{align}
            \left\|h_{1,1}^{(l)} - h_{1,2}^{(l)}\right\|_2  &\leq \rho_\sigma \left(B_\sigma \sqrt{D} \Delta_x^l + B_x \left\|h_{1,1}^{(l-1)} - h_{1,2}^{(l-1)}\right\|_2 + \Delta_b^l \right) \nonumber \\
            &\leq \rho_\sigma \Delta_b^l + \rho_\sigma B_\sigma \sqrt{D} \Delta_x^l + \gamma\left(\rho_\sigma \Delta_b^l + \rho_\sigma B_\sigma \sqrt{D} \Delta_x^l + \gamma \left\|h_{1,1}^{(l-2)} - h_{1,2}^{(l-2)}\right\|_2 \right) \nonumber \\
            &\leq \cdots \cdots \nonumber \\
            &\leq \rho_\sigma \left(\sum_{r = 0}^{l-1} \gamma^r  \Delta_b^{l-r} + B_{\sigma} \sqrt{D}  \sum_{r = 0}^{l-2} \gamma^r \Delta_x^{l-r} + B_{in}(T) \gamma^{l-1} \Delta_x^{1} \right). \nonumber
        \end{align}
        When $l = 1$, we have
        \begin{align}
            \left\|h_{i,1}^{(1)} - h_{i,2}^{(1)}\right\|_2 &= \left\|\sigma\left( W_{x,1}^{(1)} x(t_i; t_{i-1}) + W_{h,1}^{(1)} h_{i-1,1}^{(1)} + b_1^{(1)} \right) - \sigma\left( W_{x,2}^{(1)} x(t_i; t_{i-1}) + W_{h,2}^{(1)} h_{i-1,2}^{(1)} + b_2^{(1)} \right)\right\|_2 \nonumber \\
            &\leq \rho_\sigma \left( B_{in}(T) \left\|W_{x,1}^{(1)} - W_{x,2}^{(1)} \right\|_2 + \left\| W_{h,1}^{(1)}h_{i-1,1}^{(1)} - W_{h,2}^{(1)}h_{i-1,2}^{(1)} \right\|_2 + \left\|b_1^{(1)} - b_2^{(1)}\right\|_2  \right) \nonumber \\
            &\leq \rho_\sigma \left( B_{in}(T) \Delta_x^1 + B_\sigma \sqrt{D} \Delta_h^1 +  B_h \left\| h_{i-1,1}^{(1)} - h_{i-1,2}^{(1)} \right\|_2 + \Delta_b^1  \right). \nonumber
        \end{align}
        Again repeat it recursively, we can get
        \begin{align}
             &\quad\left\|h_{i,1}^{(1)} - h_{i,2}^{(1)}\right\|_2 \nonumber\\
             &\leq \rho_\sigma \left( B_{in}(T) \Delta_x^1 + B_\sigma \sqrt{D} \Delta_h^1 +  B_h \left\| h_{i-1,1}^{(1)} - h_{i-1,2}^{(1)} \right\|_2 + \Delta_b^1  \right) \nonumber \\
             &\leq \rho_\sigma \left( B_{in}(T) \Delta_x^1 +  B_\sigma \sqrt{D} \Delta_h^1 + \Delta_b^1 \right)   + \beta \left( \rho_\sigma \left( B_{in}(T) \Delta_x^1 +  B_\sigma \sqrt{D} \Delta_h^1 + \Delta_b^1 \right)+ \beta \left\| h_{i-2,1}^{(1)} - h_{i-2,2}^{(1)} \right\|_2  \right) \nonumber \\
             &\leq \cdots \cdots \nonumber \\
             &\leq \rho_\sigma \left( S_{i-1}^0 \Delta_b^{1} + B_{in}(T) S_{i-1}^0 \Delta_x^{1} + B_{\sigma} \sqrt{D}   S_{i-2}^0 \Delta_h^{1} \right). \nonumber
        \end{align}
        Now suppose for all $i < i_0$, $l < l_0$, \eqref{delta_h_i(main)} is true. Consider the case $i = i_0$, $l = l_0$, we have
        {\allowdisplaybreaks[4]
        \begin{align}
            &\quad\left\|h_{i_0,1}^{(l_0)} - h_{i_0,2}^{(l_0)}\right\|_2 \nonumber\\
            &= \left\|\sigma\left( W_{x,1}^{(l_0)} h_{i_0,1}^{(l_0-1)} + W_{h,1}^{(l_0)} h_{i_0-1,1}^{(l_0)} + b_1^{(l_0)} \right) - \sigma\left( W_{x,2}^{(l_0)}h_{i_0,2}^{(l_0-1)} + W_{h,2}^{(l_0)} h_{i_0-1,2}^{(l_0)} + b_2^{(l_0)} \right)\right\|_2 \nonumber \\
            & \leq \rho_\sigma \left( \left\|W_{x,1}^{(l_0)} h_{i_0,1}^{(l_0-1)} - W_{x,2}^{(l_0)}h_{i_0,2}^{(l_0-1)} \right\|_2 + \left\|W_{h,1}^{(l_0)} h_{i_0-1,1}^{(l_0)} - W_{h,2}^{(l_0)} h_{i_0-1,2}^{(l_0)} \right\|_2 + \left\| b_1^{(l_0)} - b_2^{(l_0)} \right\|_2  \right) \nonumber \\
            &\leq \rho_\sigma \left( B_x \left\|h_{i_0,1}^{(l_0-1)} - h_{i_0,2}^{(l_0-1)} \right\|_2 + B_\sigma \sqrt{D} \Delta_x^{l_0} + B_h \left\|h_{i_0-1,1}^{(l_0)} - h_{i_0-1,2}^{(l_0)} \right\|_2 + B_\sigma \sqrt{D} \Delta_x^{l_0} + \Delta_b^{l_0} \right) \nonumber \\
            &\leq \rho_\sigma \beta \left(\sum_{r = 0}^{l_0-1} \gamma^r S_{i_0-2}^r \Delta_b^{l_0-r} + B_{\sigma} \sqrt{D} \sum_{r = 0}^{l_0-2} \gamma^r S_{i_0-2}^r \Delta_x^{l_0-r} + B_{in}(T) \gamma^{l_0-1} S_{i_0-2}^{l_0-1} \Delta_x^{1} \right.\nonumber \\
            & \quad \quad \quad \left. + B_{\sigma} \sqrt{D} \sum_{r = 0}^{l_0-1} \gamma^r S_{i_0-3}^r \Delta_h^{l_0-r} \right)  + \rho_\sigma \gamma \left(\sum_{r = 0}^{l_0-2} \gamma^r S_{i_0-1}^r \Delta_b^{l_0-1-r} + B_{\sigma} \sqrt{D} \sum_{r = 0}^{l_0-3} \gamma^r S_{i_0-1}^r \Delta_x^{l_0-1-r} \right. \nonumber \\
            & \quad \quad \quad \left. B_{in}(T) \gamma^{l_0-2} S_{i_0-1}^{l_0-2} \Delta_x^{1}  + B_{\sigma} \sqrt{D} \sum_{r = 0}^{l_0-2} \gamma^r S_{i_0-2}^r \Delta_h^{l_0-1-r} \right) + \rho_\sigma \left(B_\sigma \sqrt{D} \Delta_x^{l_0} + B_\sigma \sqrt{D} \Delta_x^{l_0} + \Delta_b^{l_0} \right) \nonumber \\
            & \leq \rho_\sigma\left( \sum_{r = 1}^{l_0-1} \gamma^r (\beta S_{i_0-2}^{r} + S_{i_0-1}^{r-1}) \Delta_b^{l_0-r} + B_{\sigma} \sqrt{D} \sum_{r = 1}^{l_0-2} \gamma^r (\beta S_{i_0-2}^{r} + S_{i_0-1}^{r-1}) \Delta_x^{l_0-r} \right. \nonumber \\
            &\quad \quad \quad \left. + B_{in}(T) \gamma^{l_0-1} (\beta S_{i_0-2}^{l_0-1} + S_{i_0-1}^{l_0-2}) \Delta_x^{1} +  B_{\sigma} \sqrt{D} \sum_{r = 1}^{l_0-1} \gamma^r (\beta S_{i_0-3}^{r} + S_{i_0-2}^{r-1}) \Delta_h^{l_0-r}  \right) \nonumber \\
            & \quad +\rho_\sigma \left( (1 + \beta S_{i_0-2}^0)\left(\Delta_b^{l_0} + B_\sigma \sqrt{D} \Delta_x^{l_0}\right) + (1 + \beta S_{i_0-3}^0) B_\sigma \sqrt{D} \Delta_h^{l_0} \right).  \nonumber
        \end{align}
        }
        Using the fact that $1 + \beta S_{i-1}^{0} = S_{i}^{0}$ and
        \begin{align}
            \beta S_{i-1}^{r} + S_{i}^{r-1} &= \beta \sum_{j=0}^{i-1} \tbinom{j+r}{r} \beta^j + \sum_{j=0}^{i} \tbinom{j+r-1}{r-1} \beta^j  = 1 + \sum_{j=1}^{i}\left(\tbinom{j+r-1}{r} + \tbinom{j+r-1}{r-1} \right) \beta^j \nonumber \\
            &= \sum_{j=0}^{i} \tbinom{j+r}{r} \beta^j = S_i^r, \nonumber
        \end{align}
        \eqref{delta_h_i(main)} is proved.
    \end{proof}

    \begin{claim}\label{claim2}
        For $\forall 1 \leq l \leq L$, $1 \leq i \leq N$ and $t \in (t_i, t_{i+1}]$, $\|h_1^{(l)}(t; S) - h_2^{(l)}(t; S) \|_2$ is bounded by 
        \begin{align}
            \left\|h_1^{(l)}(t; S) - h_2^{(l)}(t; S) \right\|_2 & \leq \rho_\sigma \left(\sum_{r = 0}^{l-1} \gamma^r S_i^r \Delta_b^{l-r} + B_{\sigma} \sqrt{D} \sum_{r = 0}^{l-2} \gamma^r S_i^r \Delta_x^{l-r} \right.\\ 
            & \left.+ B_{in}(T) \gamma^{l-1} S_i^{l-1} \Delta_x^{1} + B_{\sigma} \sqrt{D} \sum_{r = 0}^{l-1} \gamma^r S_{i-1}^r \Delta_h^{l-r} \right).
            \label{delta_h(main)}
        \end{align}
    \end{claim}
    \begin{proof}[Proof of Claim \ref{claim2}]
        When $l = 1$, by the definition of $h^{(1)}(t; S)$ and \eqref{delta_h_i(main)}, for any $1\leq i \leq N$ and $t \in (t_i, t_{i+1}]$,  we have 
        \begin{align}
            \left\|h_1^{(1)}(t; S) - h_2^{(1)}(t; S) \right\|_2 &=  \left\| \sigma\left( W_{x,1}^{(1)} x(t; t_{i}) + W_{h,1}^{(1)} h_{i,1}^{(1)} + b_1^{(1)} \right) - \sigma\left( W_{x,2}^{(1)} x(t; t_{i}) + W_{h,2}^{(1)} h_{i,2}^{(1)} + b_2^{(1)} \right)\right\|_2 \nonumber \\
            &\leq \rho_\sigma \left( B_{in}(T) \left\|W_{x,1}^{(1)} - W_{x,2}^{(1)} \right\|_2 + \left\| W_{h,1}^{(1)}h_{i,1}^{(1)} - W_{h,2}^{(1)}h_{i,2}^{(1)} \right\|_2 + \left\|b_1^{(1)} - b_2^{(1)}\right\|_2  \right) \nonumber \\
            &\leq \rho_\sigma \left( B_{in}(T) \Delta_x^1 + B_\sigma \sqrt{D} \Delta_h^1 +  B_h \left\| h_{i,1}^{(1)} - h_{i,2}^{(1)} \right\|_2 + \Delta_b^1  \right) \nonumber \\
            & \leq \rho_\sigma \beta  \left( S_{i-1}^0 \Delta_b^{1} + B_{\sigma} \sqrt{D} S_{i-1}^0 \Delta_x^{1} + B_{\sigma} \sqrt{D}   S_{i-2}^0 \Delta_h^{1} \right) \nonumber \\
            & \quad + \rho_\sigma \left( B_{in}(T) \Delta_x^1 + B_\sigma \sqrt{D} \Delta_h^1 + \Delta_b^1  \right) \nonumber \\
            & \leq \rho_\sigma \left( S_{i}^0 \Delta_b^{1} + B_{in}(T) S_{i}^0 \Delta_x^{1} + B_{\sigma} \sqrt{D}   S_{i-1}^0 \Delta_h^{1} \right). \nonumber
        \end{align}
        Now suppose for all $l < l_0$, \eqref{delta_h_i(main)} is true for any $1 \leq i \leq N$ and $t \in (t_i, t_{i+1}]$. Considering the case $l = l_0$, for any $1 \leq i \leq N$ and $t \in (t_i, t_{i+1}]$, we have
        {\allowdisplaybreaks[4]
        \begin{align}
            &\quad\left\|h_1^{(l_0)}(t; S) - h_2^{(l_0)}(t; S) \right\|_2 \nonumber\\
            &= \left\| \sigma\left( W_{x,1}^{(l_0)} h_1^{(l_0-1)}(t; S)  + W_{h,1}^{(l_0)} h_{i,1}^{(l_0)} + b_1^{(l_0)} \right) - \sigma\left( W_{x,2}^{(l_0)} h_2^{(l_0-1)}(t; S) + W_{h,2}^{(l_0)} h_{i,2}^{(l_0)} + b_2^{(l_0)} \right)\right\|_2 \nonumber \\
            & \leq \rho_\sigma \left( \left\|W_{x,1}^{(l_0)} h_1^{(l_0-1)}(t; S) - W_{x,2}^{(l_0)} h_2^{(l_0-1)}(t; S) \right\|_2 + \left\|W_{h,1}^{(l_0)} h_{i,1}^{(l_0)} - W_{h,2}^{(l_0)} h_{i,2}^{(l_0)} \right\|_2 + \left\| b_1^{(l_0)} - b_2^{(l_0)} \right\|_2 \right) \nonumber \\
            & \leq \rho_\sigma \left( \Delta_b^{l_0} + B_\sigma \sqrt{D} \Delta_x^{l_0} + B_x  \left\|h_1^{(l_0-1)}(t; S) - h_2^{(l_0-1)}(t; S) \right\|_2 +  B_\sigma \sqrt{D} \Delta_h^{l_0} + B_h \left\| h_{i,1}^{(l_0)} - h_{i,2}^{(l_0)} \right\|_2 \right) \nonumber \\
            &\leq \rho_\sigma \gamma \left(\sum_{r = 0}^{l_0-2} \gamma^r S_i^r \Delta_b^{l_0-1-r} + B_{\sigma} \sqrt{D}  \sum_{r = 0}^{l_0-3} \gamma^r S_i^r \Delta_x^{l_0-1-r} + B_{in}(T) \gamma^{l_0-2} S_i^{l_0-2} \Delta_x^{1} \right. \nonumber \\ 
            &\quad \quad \quad \left. + B_{\sigma} \sqrt{D} \sum_{r = 0}^{l_0-2} \gamma^r S_{i-1}^r \Delta_h^{l_0-1-r} \right)  + \rho_\sigma \beta \left(\sum_{r = 0}^{l_0-1} \gamma^r S_{i-1}^r \Delta_b^{l_0-r} + B_{\sigma} \sqrt{D} \sum_{r = 0}^{l_0-2} \gamma^r S_{i-1}^r \Delta_x^{l_0-r} \right. \nonumber \\ 
            &\quad \quad \quad \left. + B_{in}(T) \gamma^{l_0-1} S_{i-1}^{l_0-1} \Delta_x^{1} + B_{\sigma} \sqrt{D} \sum_{l = 0}^{l_0-1} \gamma^r S_{i-2}^r \Delta_h^{l_0-r} \right)  + \rho_\sigma \left( \Delta_b^{l_0} + B_\sigma \sqrt{D} \Delta_x^{l_0} +  B_\sigma \sqrt{D} \Delta_h^{l_0} \right) \nonumber \\
            & \leq \rho_\sigma\left( \sum_{r = 1}^{l_0-1} \gamma^r (\beta S_{i-1}^{r} + S_{i}^{r-1}) \Delta_b^{l_0-r} + B_{\sigma} \sqrt{D} \sum_{r = 1}^{l_0-2} \gamma^r (\beta S_{i-1}^{r} + S_{i}^{r-1}) \Delta_x^{l_0-r} \right. \nonumber \\
            &\quad \quad \quad \left. + B_{in}(T) \gamma^{l_0-1} ( \beta S_{i-1}^{l_0-1} + S_{i}^{l_0-2}) \Delta_x^{1} + B_{\sigma} \sqrt{D} \sum_{r = 1}^{l_0-1} \gamma^r (\beta S_{i-2}^{r} + S_{i-1}^{r-1}) \Delta_h^{l_0-r}  \right) \nonumber \\
            & \quad +\rho_\sigma \left( (1 + \beta S_{i-2}^0)\left(\Delta_b^{l_0} + (B_\sigma \sqrt{D} \vee B_{in}(T))  \Delta_x^{l_0}\right) + (1 + \beta S_{i-3}^0) B_\sigma \sqrt{D} \Delta_h^{l_0} \right) \nonumber \\
            & \leq \rho_\sigma \left(\sum_{r = 0}^{l_0-1} \gamma^r S_i^r \Delta_b^{l_0-r} + B_{\sigma} \sqrt{D} \sum_{r = 0}^{l_0-2} \gamma^r S_i^r \Delta_x^{l_0-r}  + B_{in}(T) \gamma^{l_0-1} S_i^{l_0-1}  \Delta_x^{1}  + B_{\sigma} \sqrt{D} \sum_{r = 0}^{l_0-1} \gamma^r S_{i-1}^r \Delta_h^{l_0-r} \right). \nonumber
        \end{align}
       }
        Hence \eqref{delta_h(main)} is proved.
    \end{proof}
    Now we prove Lemma 4. For $t \in (t_i, t_{i+1}]$, we have 
    {\allowdisplaybreaks[4]
    \begin{align}
        \left| \lambda_{\theta_1}(t;S) - \lambda_{\theta_2}(t;S) \right| &= \left| f\left(W_{x,1}^{(L+1)} h^{(L)}(t;S) + b_1^{(L+1)}\right) - f\left(W_{x,2}^{(L+1)} h^{(L)}(t;S) + b_2^{(L+1)}\right) \right|  \nonumber \\
        & \leq \rho_f \left( \left\|b_1^{(L+1)} - b_2^{(L+1)} \right\|_2 + \left\| W_{x,1}^{(L+1)} h^{(L)}(t;S) - W_{x,2}^{(L+1)} h^{(L)}(t;S) \right\|_2  \right) \nonumber \\
        & \leq \rho_f \left( \Delta_b^{L+1} + B_\sigma \sqrt{D} \Delta_x^{L+1} + B_x \left\|h_1^{(L)}(t; S) - h_2^{(L)}(t; S) \right\|_2 \right) \nonumber \\
        & \leq \rho_f \gamma \left(\sum_{l = 0}^{L-1} \gamma^l S_i^l \Delta_b^{L-l} + B_{\sigma} \sqrt{D} \sum_{l = 0}^{L-2} \gamma^l S_i^l \Delta_x^{L-l} + B_{in}(T) \gamma^{L-1} S_i^{L-1} \Delta_x^{1} \right. \nonumber \\
        &\quad \quad \quad \left. + B_{\sigma} \sqrt{D} \sum_{l = 0}^{L-1} \gamma^l S_{i-1}^l \Delta_h^{L-l} \right)  + \rho_f \Delta_b^{L+1} + \rho_f B_{\sigma} \sqrt{D} \Delta_x^{L+1}. \nonumber
    \end{align}   
    }

\subsection{ Proof of Lemma 5}
    From Lemma 4, for $\forall ~ \lambda_{\theta_1}, \lambda_{\theta_2} \in \mathcal{F}$, we have
    \begin{align}
        &\quad d_N(\lambda_{\theta_1}, \lambda_{\theta_2}) \nonumber\\
        &\leq \rho_f \gamma \left(\sum_{l = 0}^{L-1} \gamma^l S_N^l \Delta_b^{L-l} +  B_{\sigma} \sqrt{D}  \sum_{l = 0}^{L-2} \gamma^l S_N^l \Delta_x^{L-l} + B_{in}(T) \gamma^{L-1} S_N^{L-1} \Delta_x^{1} + B_{\sigma} \sqrt{D} \sum_{l = 0}^{L-1} \gamma^l S_{N-1}^l \Delta_h^{L-l} \right) \nonumber \\ 
        &\quad + \rho_f \Delta_b^{L+1} + \rho_f B_{\sigma} \sqrt{D} \Delta_x^{L+1} \nonumber \\
        &\leq \rho_f \left(B_\sigma \sqrt{D} \vee B_{in}(T) \vee 1\right) \left( \gamma^L \vee 1\right) S_N^{L-1} \Delta_{\theta} \nonumber \\
        &\leq \rho_f \left(B_\sigma \sqrt{D} \vee B_{in}(T) \vee 1\right) \left( \gamma^L \vee 1\right) (N+1)^{L-1} \frac{\beta^{N+1}-1}{\beta-1} \Delta_{\theta}, \nonumber
    \end{align}
    where $\Delta_{\theta} \triangleq \sum_{l = 0}^{L+1}\left(\Delta_b^{l} + \Delta_x^{l} + \Delta_h^{l}\right)$.
    
    Define $C(N) \triangleq \rho_f \left(B_\sigma \sqrt{D} \vee B_{in}(T) \vee 1\right) \left( \gamma^L \vee 1\right) (N+1)^{L-1} (\beta^{N+1}-1)/(\beta-1) $, using Lemma \ref{matrix covering number lemma(main)} and $\|\cdot\|_2 \leq \|\cdot\|_F$, we can get
    \begin{align}
        \mathcal{N}\left(\mathcal{F}_{\Theta}^{\mathcal{B}}, \epsilon, d_N^\lambda(\cdot, \cdot) \right) &\leq \prod_{l = 1}^{L+1} \mathcal{N}\left(W_x^{(l)}, \frac{\epsilon}{C(N)(3L+2)}, \|\cdot\|_F \right) ~ \prod_{l = 1}^{L} \mathcal{N}\left(W_h^{(l)}, \frac{\epsilon}{C(N)(3L+2)}, \|\cdot\|_F \right) \nonumber \\ 
        &\quad ~ \prod_{l = 1}^{L+1} \mathcal{N}\left(b^{(l)}, \frac{\epsilon}{C(N)(3L+2)}, \|\cdot\|_2 \right)  \nonumber \\
        &\leq \left( 1 + \frac{C(N)(3L+2)B_m\sqrt{D}}{\epsilon}\right)^{D^2(3L+2)} ~, \nonumber
    \end{align}
    where $B_m = \max\{B_b, B_h, B_x\}$.

\subsection{Proof of Theorem 3}

\begin{lemma}
\label{lemma of X_theta(main)}
     Under 
     assumptions 
     (B1)-(B3), for fixed $s \in \mathbb N$, with probability at least $1-\delta$, we have 
    \begin{align}
        \sup_{\theta \in \Theta} |X_\theta(s)|  &\leq \frac{48}{\sqrt{n}}\left(T + \frac{1}{l_f}\right)(s+1)\left\{4u_f \left(\sqrt{\log\left(\frac{2}{\delta}\right)} + D \sqrt{(3L+2)\log\left(1+ M(s)\right)} ~\right) + D\sqrt{3L+2} \right\}. \nonumber
    \end{align}
    Hence 
    \begin{align}
        \sup_{\|\theta\| \leq B_m} |X_\theta(s)|  \leq \Tilde{O}\left(\sqrt{\frac{{D^2 L^2 s^3}}{n}}\right), \nonumber
    \end{align}
    where 
    $M(s) = \rho_f B_m\sqrt{D}\left(B_\sigma \sqrt{D} \vee B_{in}(T) \vee 1\right) \left( \gamma^L \vee 1\right) (s+1)^{L-1} (\beta^{s+1}-1)/(\beta-1) $, $B_m = \max\{B_b, B_h, B_x\}$, $\gamma = \rho_\sigma B_x$, and $\beta = \rho_\sigma B_h$.
\end{lemma}
\begin{proof}[Proof of Lemma \ref{lemma of X_theta(main)}]
    For $1\leq k \leq n$, denote $X_{\theta,k}(s) = \mathbb E \left[\text{loss}(\lambda_\theta, S_{test}) \mathbbm{1}_{\{N_e \leq s\}}\right] - \text{loss}(\lambda_\theta, S_k) \mathbbm{1}_{\{N_{ek} \leq s\}}$. Then $X_{\theta}(s) = n^{-1} \sum_{k=1}^{n} X_{\theta,k}(s)$. For two parameters $\theta_1$ and $\theta_2$, we have
    \begin{align}
        & ~~~~ \left|\text{loss}(\lambda_{\theta_1}, S_k) \mathbbm{1}_{\{N_{ek} \leq s\}} - \text{loss}(\lambda_{\theta_2}, S_k) \mathbbm{1}_{\{N_{ek} \leq s\}}\right| \nonumber \\
        &\leq \Big| \sum_{i=1}^{N_k}(\log \lambda_{\theta_1}(t_i) - \log \lambda_{\theta_2}(t_i)) \Big| + \Big| \int_0^T \left(\lambda_{\theta_1}(t) - \lambda_{\theta_2}(t)\right) \mathrm{d t} \Big| \nonumber \\
        & \leq \frac{1}{l_f} \sum_{i=1}^{N_k} | \lambda_{\theta_1}(t_i) - \lambda_{\theta_2}(t_i))| +  \int_0^T \left|\lambda_{\theta_1}(t) - \lambda_{\theta_2}(t)\right| \mathrm{d t} \nonumber \\
        & \leq \left(T + \frac{N_k}{l_f}\right) d_{N_k}(\lambda_{\theta_1}, \lambda_{\theta_2}) \nonumber \\
        & \leq \left(T + \frac{1}{l_f}\right) (s + 1) d_{s}(\lambda_{\theta_1}, \lambda_{\theta_2}), \nonumber
    \end{align}
    and similarly, 
    \begin{align}
        \left|\mathbb E \left[\text{loss}(\lambda_{\theta_1}, S_{test}) \mathbbm{1}_{\{N_e \leq s\}}\right] - \mathbb E \left[\text{loss}(\lambda_{\theta_2}, S_{test}) \mathbbm{1}_{\{N_e \leq s\}}\right] \right| \leq \left(T + \frac{1}{l_f}\right) (s + 1) d_{s}(\lambda_{\theta_1}, \lambda_{\theta_2}). \nonumber
    \end{align}
    Hence 
    \begin{align}
        \left|X_{\theta_1, k}(s) - X_{\theta_2, k}(s)\right| \leq 2\left(T + \frac{1}{l_f}\right) (s + 1) d_{s}(\lambda_{\theta_1}, \lambda_{\theta_2}). \nonumber
    \end{align}
     By the property of bounded variable,  $X_{\theta_1, k}(s) - X_{\theta_2, k}(s)$ is $2\left(T + 1/l_f\right) (s + 1) d_{s}(\lambda_{\theta_1}, \lambda_{\theta_2})$-sub-gaussian. Since $\{X_{\theta_1, k}(s) - X_{\theta_2, k}(s)\}_{k=1}^{n}$ is mutually independent, $X_{\theta_1}(s) - X_{\theta_2}(s)$ is $2\left(T + 1/l_f\right) (s + 1) d_{s}(\lambda_{\theta_1}, \lambda_{\theta_2})/\sqrt{n}$-sub-gaussian. 
    From assumptions 2 and 3, there exists $\|\theta_0 \| \leq B_m$ such that $\lambda_{\theta_0} \equiv 1$, implying $X_{\theta_0}(s) = 0$.
    
    The diameter of $\mathcal{F}$ under the distance $d_s(\cdot,\cdot)$ can be bounded by
    \begin{align}
        \text{diam}\left(\mathcal{F} | d_s \right)  &\leq \sup_{\theta_1, \theta_2 \in \Theta} d^{\lambda}_s(\lambda_{\theta},\lambda_{\theta_0}) 
        \leq \sup_{\theta_1, \theta_2 \in \Theta} \sup_{\# S \leq s} \|\lambda_{\theta_1}(t;S) - \lambda_{\theta_2}(t;S)\|_{L^\infty} \nonumber \\
        &\leq 2 u_f~. 
        \label{diam of F_theta(main)}
    \end{align}
    By Lemma 5,  we get
    \begin{align}
        \log \mathcal{N}\left(\mathcal{F}, \epsilon, d_s(\cdot, \cdot) \right) \leq D^2(3L+2) \log \left( 1 + \frac{C(s)(3L+2)B_m\sqrt{D}}{\epsilon}\right)~,
        \label{covering number int of F_theta(main)}
    \end{align}
    where $C(s) = \rho_f \left(B_\sigma \sqrt{D} \vee B_{in}(T) \vee 1\right) \left( \gamma^L \vee 1\right) (s+1)^{L-1} (\beta^{s+1}-1)/(\beta-1) $, $B_m = \max\{B_b, B_h, B_x\}$.
    Denote $M(s) = C(s)(3L+2)B_m\sqrt{D}$, $\mathcal{D} = \text{diam}\left(\mathcal{F}_{\Theta} | d^{\lambda}_s \right)$. We have
    \begin{align}
        \int_{0}^{2\mathcal{D}} \sqrt{\log\left( 1 + \frac{M(s)}{\epsilon} \right)} \mathrm{d} \epsilon &\leq \left(\int_{0}^{a} + \int_{a}^{2\mathcal{D}}\right) \sqrt{\log\left( 1 + \frac{M(s)}{\epsilon} \right)} \mathrm{d} \epsilon ~  ~ (\forall 0 \leq a \leq 2\mathcal{D})\nonumber\\
        & \leq \inf_{0 \leq a \leq 2\mathcal{D}} \left\{ \int_{0}^{a} \sqrt{\frac{M(s)}{\epsilon}} \mathrm{d} \epsilon + \int_{a}^{2\mathcal{D}} \sqrt{\log\left( 1 + \frac{M(s)}{\epsilon} \right)} \mathrm{d} \epsilon \right\} \nonumber \\
        & \leq \inf_{0 \leq a \leq 2\mathcal{D}} \left\{ 2\sqrt{M(s)a} + 2\mathcal{D} \sqrt{\log\left( 1 + \frac{M(s)}{a} \right)}\right\} \nonumber \\
        & \leq 2 + 2\mathcal{D} \sqrt{\log\left( 1 + {M(s)}^2\right)} ~ ~ (\text{take} ~ a = {M(s)}^{-1}) \nonumber \\
        &\leq 2 + 4\mathcal{D} \sqrt{\log\left( 1 + M(s)\right)}, 
        \label{int computation of X_theta(s)(main)}
    \end{align}
    where we need $2\mathcal{D}M(s) \geq 1$. If $2\mathcal{D}M(s) < 1$, \eqref{int computation of X_theta(s)(main)} is obvious since the integral is less than $2$.

    Combining \eqref{diam of F_theta(main)}, \eqref{covering number int of F_theta(main)}, \eqref{int computation of X_theta(s)(main)} and using Lemma \ref{concentration lemma(main)}, we have
    \begin{align}
        \sup_{\theta \in \Theta} |X_\theta(s)|  &\leq \frac{24}{\sqrt{n}}\left(T + \frac{1}{l_f}\right)(s+1)\left(\mathcal{D}\left(4\sqrt{\log\left(\frac{2}{\delta}\right)} + 4D\sqrt{(3L+2)\log(1+ M(s))} \right)+2D\sqrt{3L+2}\right) \nonumber\\
        &\leq \frac{48}{\sqrt{n}}\left(T + \frac{1}{l_f}\right)(s+1)\left\{4u_f \left(\sqrt{\log\left(\frac{2}{\delta}\right)} + D \sqrt{(3L+2)\log\left(1+ M(s)\right)} ~\right) + D\sqrt{3L+2} \right\}~. \nonumber
    \end{align}
    
\end{proof}


\begin{lemma}
    Suppose the event number $N_e$ satisfies the tail condition
     \[\mathbb P(N_e \geq s) \leq a_N \exp(-c_N s), ~ s \in \mathbb N.\]  Under 
    assumptions 
    (B1)-(B3), for fixed $s \in \mathbb N$, we have 
    \begin{align}
        \sup_{\theta \in \Theta} |E_\theta(s)| \leq \left(T+ \frac{1}{l_f}\right)(u_f+2) \frac{a_N (s+2)}{(1-\exp(-c_N))^2} \exp(-c_N(s+1)) ~. \nonumber
    \end{align}
\label{lemma of E_theta(main)}
\end{lemma}
\begin{proof} [Proof of Lemma \ref{lemma of E_theta(main)}]
    From assumptions (B2) and (B3), there exists $\theta_0 \in \Theta$ such that $\lambda_{\theta_0} \equiv 1$. Then
    \begin{align}
        |E_\theta(s)| &= \left|\mathbb E \left[\text{loss}(\lambda_\theta, S_{test}) \mathbbm{1}_{\{N_{e} > s\}}\right]\right|  \leq \mathbb E \left|\text{loss}(\lambda_\theta, S_{test}) \right| \mathbbm{1}_{\{N_e > s\}} \nonumber\\
        & \leq \mathbb E \left|\text{loss}(\lambda_\theta, S_{test}) - \text{loss}(\lambda_{\theta_0}, S_{test}) \right| \mathbbm{1}_{\{N_e > s\}} + \mathbb E \left|\text{loss}(\lambda_{\theta_0}, S_{test}) \right| \mathbbm{1}_{\{N_e > s\}} \nonumber \\
        & \leq \mathbb E \left[\left(T+ \frac{1}{l_f}\right)(N_e+1) d_{N_e} (\lambda_\theta, \lambda_{\theta_0})\right] \mathbbm{1}_{\{N_e > s\}} + T \mathbb P(N_e > s) \nonumber \\
        & \leq \left(T+ \frac{1}{l_f}\right)(u_f+1) \mathbb E [(N_e+1) \mathbbm{1}_{\{N_e > s\}}] + T \mathbb P(N_e > s) \nonumber
    \end{align}
    By the tail condition $\mathbb P(N_e \geq s) \leq a_N \exp(-c_N s), ~ s \in \mathbb N$, we have
    \begin{align}
        |E_\theta(s)| &\leq \left(T+ \frac{1}{l_f}\right)(u_f+1) \frac{a_N (s+1)}{(1-\exp(-c_N))^2} \exp(-c_N(s+1))\nonumber\\
        &\quad + \left(T+ \frac{1}{l_f}\right)(u_f+2) a_N \exp(-c_N(s+1)) \nonumber\\
        &\leq \left(T+ \frac{1}{l_f}\right)(u_f+2) \frac{a_N (s+2)}{(1-\exp(-c_N))^2} \exp(-c_N(s+1)). \nonumber
    \end{align}
\end{proof}

Now we prove Theorem 3. From Lemma 2, we have
    \begin{align}
        \mathbb P\left(\sup_{\theta \in \Theta}|X_\theta| > t \right) \leq  \mathbb P\left(\sup_{\theta \in \Theta}|X_\theta(s)| + \sup_{\theta \in \Theta}|E_\theta(s)| > t \right) + \mathbb P(N_{e(n)} > s) . \nonumber
    \end{align}
    Since
    \begin{align}
        \mathbb P (N_{e(n)} > s) \leq n \mathbb P (N_e > s) \leq n a_N \exp(- c_N s), \nonumber
    \end{align}
    we can take $s_0 = \lceil\left(\log\left(2a_{N}n/\delta\right) - 1 \right)/c_N\rceil $ such that $ n a_N \exp(- c_N s_0) \leq \delta/2$, so we only need solve $t>0$ such that
    \begin{align}
        \mathbb P\left(\sup_{\theta \in \Theta}|X_\theta(s_0)| + \sup_{\theta \in \Theta}|E_\theta(s_0)| > t \right) \leq \frac{\delta}{2} ~. \nonumber
    \end{align}
    From Lemma \ref{lemma of E_theta(main)}, we have
    \begin{align}
        \sup_{\theta \in \Theta} |E_\theta(s_0)| &\leq \left(T+ \frac{1}{l_f}\right)(u_f+2) \frac{a_N (s_0+2)}{(1-\exp(-c_N))^2} \exp(-c_N(s_0+1)) := B(s_0). \nonumber
    \end{align}
    By the definition of $s_0$, $B(s_0) \leq \left(T+ 1/l_f\right)(u_f+2) (s_0+2)\delta/[2n(1-\exp(-c_N))^2]$. Thus we only need to solve $t>0$ such that
    \begin{align}
        \mathbb P\left(\sup_{\theta \in \Theta}|X_\theta(s_0)| + \sup_{\theta \in \Theta}|E_\theta(s_0)| > t \right) \leq \mathbb P\left(\sup_{\theta \in \Theta}|X_\theta(s_0)|  > t - B(s_0)\right) \leq \frac{\delta}{2} ~. \nonumber
    \end{align}
    From Lemma \ref{lemma of X_theta(main)}, we can choose
    \begin{align}
        t_0 &= \frac{48}{\sqrt{n}}\left(T + \frac{1}{l_f}\right)(s_0+1)\left\{4u_f \left(\sqrt{\log\left(\frac{4}{\delta}\right)} + D \sqrt{(3L+2)\log\left(1+ M(s_0)\right)} \right) + D\sqrt{3L+2} \right\} + B(s_0) \nonumber \\
        &\leq \frac{48}{\sqrt{n}}\left(T + \frac{1}{l_f}\right)(s_0+1)\left\{4u_f \left(\sqrt{\log\left(\frac{4}{\delta}\right)} + D \sqrt{(3L+2)\log\left(1+ M(s_0)\right)} ~\right) + D\sqrt{3L+2} \right\} \nonumber \\
        & \quad + \left(T+ \frac{1}{l_f}\right)(u_f+2) \frac{ s_0+2}{(1-\exp(-c_N))^2} \frac{\delta}{2n} \nonumber \\
        & \leq \frac{192}{\sqrt{n}}\left(T + \frac{1}{l_f}\right)(s_0+1)u_f \left(\sqrt{\log\left(\frac{4}{\delta}\right)} + D \sqrt{(3L+2)}(\sqrt{\log\left(1+ M(s_0)\right)}+1) + \frac{1}{(1-\exp(-c_N))^2}~\right) ~. \nonumber
    \end{align}
    such that $\mathbb P\left(\sup_{\theta \in \Theta}|X_\theta| > t \right) \leq \delta$. Hence the theorem is proved.

\hspace{10mm}

\section{Proofs in section 5 and 6}

\subsection{Proof of Theorem 4}
    For $\lambda^{\ast}(t) = \lambda_0(t) \in W^{s, \infty}([0,T], B_0)$, $\delta = 1/2$, and $N \geq 5$, by Lemma \ref{theorem of tanh NN 1(main)},  there exists a two-layer NN $\hat{f}^N$ such that 
    \begin{align}
        \left|\hat{f}^N(x) - \lambda^{\ast}(Tx) \right| \leq  \frac{3\mathcal{C} B_0 T^s}{2N^s}, ~ 0 \leq x \leq 1, \nonumber
    \end{align}
    where $\mathcal{C} = \sqrt{2s}5^s/(s-1)! $.

    Then we have
    \begin{align}
        \left|\hat{f}^N(\frac{t}{T}) - \lambda^{\ast}(t) \right| \leq \frac{3\mathcal{C} B_0 T^s}{2N^s}, ~ 0 \leq t \leq T. \nonumber
    \end{align}
    Since $B_1 \leq \lambda^{*}(t) \leq B_0$, taking $l_f = B_1$, $u_f = B_0$ and  $\hat{\lambda}^N(t) =  f(\hat{f}^N(t/T))$, we have
    \begin{align}
        \left|\hat{\lambda}^N(t) - \lambda^{\ast}(t) \right| \leq \left|\hat{f}^N(\frac{t}{T}) - \lambda^{\ast}(t) \right| \leq  \frac{3\mathcal{C} T^s}{2N^s}, ~\forall 0 \leq t \leq T. \nonumber
    \end{align}
    Then
    \begin{align}
        |\mathbb E[\text{loss}(\hat{\lambda}^{N}, S_{test})] - \mathbb E[\text{loss}(\lambda^{\ast}, S_{test})]| &\leq \mathbb E \left|\text{loss}(\hat{\lambda}^{N}, S_{test}) - \text{loss}(\lambda^{\ast}, S_{test})\right| \nonumber \\
        &\leq \mathbb E \left(\Big| \sum_{i=1}^{N_e}(\log \tilde{\lambda}^{N}(t_i) - \log \lambda^{\ast}(t_i)) \Big| + \Big| \int_0^T \left(\tilde{\lambda}^{N}(t) - \lambda^{\ast}(t)\right) \mathrm{d t} \Big|\right) \nonumber \\
        &\leq \mathbb  E \left(T + \frac{N_e}{B_1}\right) \left\|\tilde{\lambda}^{N} - \lambda^{\ast}\right\|_{L^{\infty}[0,T]} \nonumber \\
        & \leq \left(T + \frac{1}{B_1}\right) \frac{3\mathcal{C} B_0 T^s}{2N^s} \mathbb  E(N_e+1).
        \label{case1_thm_eq1(main)}
    \end{align}
    Since $\lambda_0 \leq B_0$, $\mu \equiv 0$, taking $c = B_0$, $c_0 = 0$, and $\eta = e$ in Lemma 2
    , we have
    \begin{align}
        \mathbb P (N_e \geq s) \leq 2\sqrt{B_0 T}\exp \left(\frac{e B_0 T - s}{2} \right). \nonumber
    \end{align}
    Thus 
    \begin{align}
        \mathbb  E(N_e+1) \leq 1 + \sum_{s=1}^{\infty}  \mathbb P (N_e \geq s) \leq 1 + \frac{2\sqrt{B_0 T}}{1-\exp(-1/2)} \exp \left(\frac{e B_0 T - 1}{2} \right) \leq {5\sqrt{B_0 T + 1}} \exp \left(\frac{3 B_0 T}{2} \right).
        \label{case1_thm_expection(main)}
    \end{align}
    Combining \eqref{case1_thm_eq1(main)} and \eqref{case1_thm_expection(main)}, we get
    \begin{align}
        |\mathbb E[\text{loss}(\hat{\lambda}^{N}, S_{test})] - \mathbb E[\text{loss}(\lambda^{\ast}, S_{test})]| &\leq {5\sqrt{B_0 T + 1}} \exp \left(\frac{3 B_0 T}{2} \right) (T + \frac{1}{B_1}) \frac{3\mathcal{C} B_0 T^s}{2N^s}  \nonumber \\
        & \leq 15 \exp \left({2 B_0 T} \right) (T + \frac{1}{B_1}) \frac{\mathcal{C} B_0 T^{s}}{N^s}, \nonumber
    \end{align}
    where $\mathcal{C} = \sqrt{2s}5^s/(s-1)!$ .

    $\tilde{\lambda}^{N}$ can be naturally seen as an RNN by taking $W_h^{l} = 0$, $l = 1, 2$. The width and weights bound can be directly obtained by Lemma \ref{theorem of tanh NN 1(main)} and Remark \ref{the weights of approximation 1(main)}.

\subsection{Proof of Theorem 5}

    The proof is divided into several steps. Let $S = \{t_i\}_{i=1}^{N_e}$. Here we agree on $t_0 = 0$, $t_{N_e + 1} = T$. To be concise, we denote $S(t) = \sum_{t_i < t} \exp(-\beta(t-t_i))  +1 $,  $S_{i} = \sum_{0 < j < i} \exp(-\beta(t_i-t_j)) + 1$, $i \in \mathbb N_{+}$, hence $\lambda^{\ast}(t) = \lambda_0(t) + \alpha (S(t)-1)$, $S_{i+1} = S_i  \exp(-\beta(t_{i+1}-t_i)) + 1$, $S(t) = S_i \exp(-\beta(t-t_i)) + 1$, where we take $S_0 = 0$ by default.      

    We first fix $s_0 \in \mathbb N_{+}$.

\textbf{Step 1.} Construct the approximation of $g(x,y) = x \exp(-\beta y) + 1$, where \\
$ g \in C^{\infty}\left([-(s_0+1), 2(s_0+1)] \times [0,T]\right)$ . 

    Let $\tilde{g}(x,y) = g((3x-1)(s_0+1), Ty)$, then $\tilde{g} \in C^{\infty}\left([0,1]^2 \right)$. By simple computation, we have
    \begin{align}
        \|\tilde{g}\|_{W^{k,\infty}\left( [0,1]^2 \right)} \leq 3(s_0+1)(\beta T \vee 1)^k. \nonumber
    \end{align}
    Applying Lemma \ref{theorem of tanh NN 2(main)} to $\tilde{g}/[3(s_0+1)]$, for any $\mathcal{N} \in \mathbb N_{+}$, there exists a tanh neural network $\tilde{g}^{\mathcal{N}}$ with only one hidden layer and width $3\lceil \frac{\mathcal{N} + 10(\beta T \vee 1) }{2}\rceil\binom{\mathcal{N} + 10(\beta T \vee 1) + 2}{2}$ such that
    \begin{align}
        \left|\tilde{g}(x,y) - \tilde{g}^{\mathcal{N}}(x,y)\right| \leq 3(s_0+1)\exp(-\mathcal{N}),~ (x,y) \in [0,1]^2. \nonumber
    \end{align}
    By coordinate transformation, we get
    \begin{align}
        \left|g(x,y) - \tilde{g}^{\mathcal{N}}(\frac{1}{3(s_0+1)}x+\frac{1}{3},\frac{1}{T}y)\right| \leq 3(s_0+1)\exp(-\mathcal{N}),~ (x,y) \in [-(s_0+1), 2(s_0+1)] \times [0,T]. \nonumber
    \end{align}
    Define $\hat{g}^{\mathcal{N}}(x,y) = \tilde{g}^{\mathcal{N}}(x/[3(s_0+1)]+1/3, y/T)$. Then 
    \begin{align}
        \left|g(x,y) - \hat{g}^{\mathcal{N}}(x,y)\right| \leq 3(s_0+1)\exp(-\mathcal{N}),~ (x,y) \in [-(s_0+1), 2(s_0+1)] \times [0,T]. \nonumber
    \end{align}
    From  Lemma \ref{theorem of tanh NN 2(main)} and Remark \ref{the weights of approximation 2(main)}, the weights of $\hat{g}^{\mathcal{N}}$ are bounded by 
    \begin{align}
        O\left((s_0+1)\exp(\frac{{\mathcal{N}^{\prime}}^2 + \mathcal{N}^{\prime}  -3Cd\mathcal{N}^{\prime}}{2}) (\mathcal{N}^{\prime}(\mathcal{N}^{\prime}+2))^{3\mathcal{N}^{\prime}(\mathcal{N}^{\prime}+2)}\right), 
    \end{align}
    where $\mathcal{N}^{\prime} = \mathcal{N} + 10(\beta T \vee 1)$.
    Taking $\mathcal{N} \leftarrow \mathcal{N} + \lceil \log(3(s_0+1)) \rceil$, we have
     \begin{align}
        \left|g(x,y) - \hat{g}^{\mathcal{N}}(x,y)\right| \leq \exp(-\mathcal{N}),~ (x,y) \in [-(s_0+1), 2(s_0+1)] \times [0,T].
        \label{est err of g(main)}
    \end{align}
    Especially, $\left|g(x,y) - \hat{g}^{\mathcal{N}}(x,y)\right| \leq 1$.
    Since $\hat{g}^{\mathcal{N}} \in \mathbb R$, by a small tuning (precisely, width plus 1),  we can assume $\hat{g}^{\mathcal{N}}$ has the following structure:
    \begin{align}
        \hat{g}^{\mathcal{N}}(x,y) = V_1 \sigma\left( 
        \begin{pmatrix}
            W& B    
        \end{pmatrix}
        \begin{pmatrix}
            x\\y    
        \end{pmatrix}
        + b_0\right) . \nonumber
    \end{align}

\textbf{Step 2.} Construct the approximation of $S_i$ and $S(t)$ under the event $\{N_e \leq s_0 \}$.

    Let $h_0 = 0$, $\overline{S}_0 = 0$, for $1\leq i \leq s_0$. We construct $h_i^{\mathcal{N}}$ and $\overline{S}_i^{\mathcal{N}}$ recursively by
\begin{equation}
    \left\{
    \begin{aligned}
        h_{i}^{\mathcal{N}} &= \sigma\left( 
        \begin{pmatrix}
            W& B    
        \end{pmatrix}
        \begin{pmatrix}
            V_1 h_{i-1}^{\mathcal{N}}\\t_{i}-t_{i-1}    
        \end{pmatrix}
        + b_0\right), \\
        \overline{S}_i^{\mathcal{N}} &= V_1 h_i .
    \end{aligned}
    \right.  
    \nonumber
\end{equation}
    Hence $\overline{S}_i^{\mathcal{N}} = \hat{g}^{\mathcal{N}}(\overline{S}_{i-1}^{\mathcal{N}},t_i-t_{i-1}), 1 \leq i \leq s_0$, here $t_0 = 0$.

    Similarly, we can define $\overline{S}^{\mathcal{N}}(t), t \in (t_{i-1}, t_i]$ by
    \begin{equation}
        \left\{
            \begin{aligned}
                h^{\mathcal{N}}(t) &= \sigma\left( 
            \begin{pmatrix}
                W& B    
            \end{pmatrix}
            \begin{pmatrix}
                V_1 h_{i-1}^{\mathcal{N}}\\t-t_{i-1}    
            \end{pmatrix}
            + b_0\right), \\
            \overline{S}^{\mathcal{N}}(t) &= V_1 h(t) .
        \end{aligned}
        \right. 
        \nonumber
    \end{equation} 
    Hence $\overline{S}^{\mathcal{N}}(t) = \hat{g}^{\mathcal{N}}(\overline{S}_{i-1}^{\mathcal{N}},t-t_{i-1}), t \in (t_{i-1}, t_i]$. The approximation error can be bounded by 
    \begin{align}
        |S(t) - \overline{S}^{\mathcal{N}}(t)| &= \left|g(S_{i-1}, t_{i} - t_{i-1}) - \hat{g}^{N}(\overline{S}_{i-1}^{\mathcal{N}}, t_{i} - t_{i-1}) \right|  \nonumber\\
        &\leq \left|g(S_{i-1}, t_{i} - t_{i-1}) - g(\overline{S}_{i-1}^{\mathcal{N}}, t_{i} - t_{i-1}) \right| + \left|g(\overline{S}_{i-1}^{\mathcal{N}}, t_{i} - t_{i-1}) - \hat{g}^{N}(\overline{S}_{i-1}^{\mathcal{N}}, t_{i} - t_{i-1}) \right| \nonumber\\
        &\leq \left|S_{i-1} - \overline{S}_{i-1}^{\mathcal{N}}\right| + \left\|g - \hat{g}^{\mathcal{N}} \right\|_{\infty} \nonumber \\
        &\leq \cdots \nonumber\\
        &\leq i \left\|g - \hat{g}^{\mathcal{N}} \right\|_{\infty}, ~ t\in (t_{i-1}, t_{i}]. \nonumber
    \end{align}

    Under the event $\{N_e \leq s_0 \}$, we have 
    \begin{align}
    \left|S(t) - \overline{S}^{\mathcal{N}}(t)\right| \leq (s_0+1) \left\|g - \hat{g}^{\mathcal{N}} \right\|_{\infty} .
    \label{est err of S(t)(main)}
    \end{align}

\textbf{Step 3.} Construct the approximation of identity.

    By Lemma 3.1 of \cite{de2021approximation}, for any $\epsilon > 0$, there exists a one-layer tanh neural network $\psi_h$ such that
    \begin{align}
        |x - \psi_h(x)| \leq (6M)^4 h^2, ~ x \in [-M, M].
        \label{est err of psi_h(main)}
    \end{align}
    Actually, $\psi_h$ can be represented as 
    \begin{align}
        \psi_h(x) = \frac{1}{\sigma^{'}(0) h} \left[\sigma\left(\frac{hy}{2}\right) -  \sigma\left(-\frac{hy}{2}\right)\right] = \frac{2}{\sigma^{'}(0) h} \sigma\left(\frac{hy}{2}\right). \nonumber
    \end{align}

\textbf{Step 4.} Construct the approximation of $\lambda^{\ast}(t)$ under the event $\{N_e \leq s_0 \}$.

    Since $\lambda_0 \in W^{s, \infty}([0,T], B_0)$,  from the proof of Theorem 4
    , there exists a two-layer tanh neural network $\overline{\lambda}_0^N$ with width less than $3\lceil s/2\rceil + 6N$ such that
    \begin{align}
        \left|\overline{\lambda}_0^N(t) - \lambda_0(t) \right| \leq \frac{3\mathcal{C} T^s}{2N^s}, ~t \in [0,T].
        \label{est err of lambda_0(main)}
    \end{align}
    Moreover, the  weights of $\overline{\lambda}_0^N$ can be bounded by 
    \begin{align}
        O\left( \left[\frac{\sqrt{2s}5^s}{(s-1)!} B_0 T ^s \right]^{-s/2} N^{(1+s^2)/2} (s(s+2))^{3s(s+2)}\right). \nonumber
    \end{align}
    Here we assume $\overline{\lambda}_0^N(t)$ have the following structure
    \begin{align}
        \overline{\lambda}_0^N(t) = V_2^{'} \sigma \left( V_1^{'} \sigma\left( B^{'} t
        + b_0^{'}\right) + b_1^{'} \right) + b_2^{'}. \nonumber
    \end{align}

    Since $\lambda(t) = \lambda_0(t) + \alpha (S(t) - 1)$, we can construct its approximation by 
    \begin{eqnarray}
        h_i^{(1)} &=& \sigma\left( 
        \begin{pmatrix}
            W V_1& 0 \\ 0& 0     
        \end{pmatrix}
        h_{i-1} + 
        \begin{pmatrix}
            B&  0\\ 0& B^{'}   
        \end{pmatrix}
        \begin{pmatrix}
            t_i - t_{i-1} \\ t_i
        \end{pmatrix}
        + 
        \begin{pmatrix}
            b_0 \\ b_0^{'}
        \end{pmatrix}
        \right), ~ 1\leq i \leq s_0,
        \nonumber
    \end{eqnarray}
    and
    \begin{eqnarray}
        h^{(1)}(t;S) &=& \sigma\left( 
        \begin{pmatrix}
            W V_1& 0 \\ 0& 0     
        \end{pmatrix}
        h_i^{(1)} + 
        \begin{pmatrix}
            B&  0\\ 0& B^{'}   
        \end{pmatrix}
        \begin{pmatrix}
            t - t_i \\ t
        \end{pmatrix}
        + 
        \begin{pmatrix}
            b_0 \\ b_0^{'}
        \end{pmatrix}
        \right), \nonumber\\
        h^{(2)}(t;S) &=& \sigma\left( 
        \begin{pmatrix}
            \frac{h}{2} V_1& 0 \\ 0& V_1^{'}     
        \end{pmatrix}
        h^{(1)}(t;S) + 
        \begin{pmatrix}
            0 \\ b_1^{'}
        \end{pmatrix}
        \right)\nonumber, \\
        \hat{\lambda}(t; S) &=& f \left( 
        \begin{pmatrix}
            \frac{2 \alpha}{\sigma^{'}(0) h}&  V_2^{'}
        \end{pmatrix}
        h^{(2)}(t;S) + \left(b_2^{'} - \alpha \right)
        \right) \in \mathbb{R}^1, ~ t\in(t_i, t_{i+1}]. \label{final estmation of lambda(main)}
    \end{eqnarray}
    Under the event $\{ N_e \leq s_0 \}$, we have $B_1 \leq \lambda(t) \leq B_0 + \alpha s_0$. Recall that $f(x) = \min\{\max\{x, l_f\}, u_f\}$.  Here we can take $l_f = B_1$, $u_f = B_0 + \alpha s_0$.    

\textbf{Step 5.} Estimate the approximation error under the event $\{N_e \leq s_0 \}$.

    We rewrite \eqref{final estmation of lambda(main)} as $\hat{\lambda}(t; S) = f(\overline{\lambda}(t; S))$. Under the event $\{N_e \leq s_0 \}$ and the construction of $f$, we have
    \begin{eqnarray}
        \| \lambda^{\ast} - \hat{\lambda} \|_{L^\infty} \leq \| \lambda^{\ast} - \overline{\lambda} \|_{L^\infty} ~. \nonumber
    \end{eqnarray}
    From the constuction of $\overline{\lambda}$, we get
    \begin{align}
        \overline{\lambda}(t) = \overline{\lambda}_0^N(t) + \alpha \psi_h(\overline{S}^{\mathcal{N}}(t)) - \alpha ~,
        \label{overline_lambda(t)(main)}
    \end{align}
    then
    \begin{align}
        \left|\lambda^{\ast} (t) - \overline{\lambda}(t)\right| \leq \left|\lambda_0(t) - \overline{\lambda}_0^N(t)\right| + \alpha \left|S(t) - \psi_h(\overline{S}^{\mathcal{N}}(t))\right| ~.
        \label{est err of lambda(main)}
    \end{align}
    From \eqref{est err of S(t)(main)} \eqref{est err of psi_h(main)} and \eqref{est err of g(main)}, under the event $\{N_e \leq s_0 \}$, we have
    \begin{align}
        \left|S(t) - \psi_h(\overline{S}^{\mathcal{N}}(t))\right| &\leq \left|S(t) - \overline{S}^{\mathcal{N}}(t)\right| + \left|\overline{S}^{\mathcal{N}}(t) - \psi_h(\overline{S}^{\mathcal{N}}(t))\right| \nonumber \\
        &\leq  (s_0 + 1) \left\|g - \hat{g}^{\mathcal{N}}\right\|_\infty + (6M)^4 h^2 \nonumber \\
        &\leq (s_0 + 1) \exp(-\mathcal{N}) + (12(s_0+1))^4 h^2, ~ t \in [0,T], \nonumber
    \end{align}
    where we take $M = 2(s_0 + 1)$ to ensure that $\overline{S}^{\mathcal{N}}(t)$ can be well approximated by $\psi_h(\overline{S}^{\mathcal{N}}(t))$. On the other hand, \eqref{est err of lambda_0(main)} shows that
    \begin{align}
        \left|\overline{\lambda}_0^N(t) - \lambda_0(t) \right| \leq \frac{3\mathcal{C} B_0 T^s}{2N^s}, ~ t \in [0,T]. \nonumber
    \end{align}
    To trade off the two error terms in \eqref{est err of lambda(main)}, let $\exp(-\mathcal{N}) \asymp N^{-s}$, and then we can take 
   $\mathcal{N} =\lceil s \log(N) \rceil$. 
    Moreover, take $\mathcal{N} \leftarrow \mathcal{N} + \lceil \log(s_0 + 1)\rceil$ and $h = (12(s_0 + 1))^{-2} N^{-s/2}$. Hence, under $\{N_e \leq s_0 \}$ , we have
    \begin{align}
        \left|\lambda^{\ast} (t) - \hat{\lambda}(t)\right| \leq \left|\lambda^{\ast} (t) - \overline{\lambda}(t)\right| \leq \frac{3\mathcal{C} B_0 T^s + 4}{2N^s}, ~ t \in [0,T].
        \label{Case2: err term in step5(main)}
    \end{align}

\textbf{Step 6.} Estimate the final approximation error.

Similar to \eqref{case1_thm_eq1(main)}, we have
\begin{align}
    &\quad|\mathbb E[\text{loss}(\hat{\lambda}, S_{test})] - \mathbb E[\text{loss}(\lambda^{\ast}, S_{test})]| \nonumber \\
    &\leq \mathbb E \left|\text{loss}(\hat{\lambda}, S_{test}) - \text{loss}(\lambda^{\ast}, S_{test})\right| \nonumber \\
        &\leq \mathbb E \left(\Big| \sum_{i=1}^{N_e}(\log \hat{\lambda}(t_i) - \log \lambda^{\ast}(t_i)) \Big| + \Big| \int_0^T \left(\hat{\lambda}(t) - \lambda^{\ast}(t)\right) \mathrm{d t} \Big|\right) \nonumber \\
        &\leq \mathbb E \left[ \left(\Big| \sum_{i=1}^{N_e}(\log \hat{\lambda}(t_i) - \log \lambda^{\ast}(t_i)) \Big| + \Big| \int_0^T \left(\hat{\lambda}(t) - \lambda^{\ast}(t)\right) \mathrm{d t}  \Big|\right) \mathbbm{1}_{\{ N_e \leq s_0 \}} \right. \nonumber\\
        & \quad \quad  + \left. \left(\Big| \sum_{i=1}^{N_e}(\log \hat{\lambda}(t_i) - \log \lambda^{\ast}(t_i)) \Big| + \Big| \int_0^T \left(\hat{\lambda}(t) - \lambda^{\ast}(t)\right) \mathrm{d t}  \Big|\right) \mathbbm{1}_{\{ N_e > s_0 \}} \right] \nonumber\\
        &\leq \mathbb  E\left[ (T + \frac{N_e}{B_1}) \left\|\hat{\lambda} - \lambda^{\ast}\right\|_{L^\infty} \mathbbm{1}_{\{ N_e \leq s_0 \}} \right] + \mathbb E \left[ (T + \frac{N_e}{B_1}) \left\|\hat{\lambda} - \lambda^{\ast}\right\|_{L^\infty} \mathbbm{1}_{\{ N_e > s_0 \}} \right] \nonumber \\
        & := \mathbb{I}_1 + \mathbb{I}_2 ~. \label{hawkes approx err decomp(main)}
\end{align}

Since $\lambda_0(t) \leq B_0$, $\mu(t) = \alpha \exp(-\beta(t))$, taking  $c_\mu = \alpha/\beta$, $\eta = (\alpha + \beta)/(2\alpha)$ in  
Lemma 2
, we have
\begin{align}
    \mathbb P\left(N_e \geq s\right) &\leq \frac{2\sqrt{B_0T}}{1-c_\mu} \exp\left( \frac{\log(\eta)}{2} \left[\eta(B_0T) - (1-c_\mu\eta)s\right] \right) \nonumber\\
    &\leq \frac{2\beta \sqrt{B_0 T}}{\beta - \alpha} \exp\left( \frac{\log \left(\frac{\alpha + \beta}{2\alpha}\right)}{2} \left[\frac{\alpha + \beta}{2\alpha}(B_0 T) - \frac{\beta - \alpha}{2\beta}s\right] \right) \nonumber \\
    &:= a_e \exp\left( - c_e s \right). \nonumber
\end{align}

By \eqref{Case2: err term in step5(main)}, 
\begin{align}
    \mathbb{I}_1 &\leq \left(T + \frac{1}{B_1}\right)\frac{3\mathcal{C} B_0 T^s + 2}{2N^s} \mathbb E \left[(N_e + 1) \mathbbm{1}_{\{ N_e \leq s_0 \}} \right] \nonumber\\
    &\leq \left(T + \frac{1}{B_1}\right)\frac{3\mathcal{C} B_0 T^s + 2}{2N^s} \mathbb E \left[(N_e + 1)  \right] \nonumber \\
    &= \left(T + \frac{1}{B_1}\right)\frac{3\mathcal{C} B_0 T^s + 2}{2N^s}\left( 1+ \sum_{s=1}^{\infty} \mathbb P(N_e \geq s)\right) \nonumber \\
    &\leq \left(T + \frac{1}{B_1}\right)\left( 1+ \frac{a_e \exp(-c_e)}{1 - \exp(-c_e)}\right) \frac{3\mathcal{C} B_0 T^s + 4}{2N^s}  \label{I_1(main)}
\end{align}

On the other hand, from $\|\hat{\lambda}\|_{L^\infty} \leq B_0 + \alpha s_0$ and $\|\lambda^{\ast}\|_{L^\infty} \leq B_0 + \alpha N_e$, we have
\begin{align}
     \mathbb{I}_2 &\leq \mathbb E \left[ \left(T + \frac{N_e}{B_1}\right) \|\hat{\lambda} \|_{L^\infty} \mathbbm{1}_{\{ N_e > s_0 \}} \right] + \mathbb E \left[ \left(T + \frac{N_e}{B_1}\right) \| \lambda^{\ast}\|_{L^\infty} \mathbbm{1}_{\{ N_e > s_0 \}} \right] \nonumber \\
     &\leq \left(T + \frac{1}{B_1}\right) (B_0 + \alpha s_0)  \mathbb E \left[(N_e+1)\mathbbm{1}_{\{ N_e > s_0 \}} \right] + \left(T + \frac{1}{B_1}\right) E \left[(N_e+1) (B_0 + \alpha N_e)\mathbbm{1}_{\{ N_e > s_0 \}} \right] \nonumber \\
     &\leq \left(T + \frac{1}{B_1}\right) (B_0 + \alpha s_0)\left((s_0+1)\mathbb P (N_e \geq s_0 + 1) + \sum_{s=s_0 + 1}^{\infty} \mathbb P(N_e \geq s)\right) \nonumber \\
     &\quad +\left(T + \frac{1}{B_1}\right) \left((s_0+1)(B_0+\alpha s_0)  \mathbb P(N_e \geq s_0 + 1) + \sum_{s=s_0 +1}^{\infty} (2\alpha s + B_0) \mathbb P(N_e \geq s)\right) \nonumber \\
     &\leq \left(T + \frac{1}{B_1}\right) (B_0 + \alpha s_0) a_e \exp(- c_e (s_0+1))\left((s_0+1)  + \frac{1}{1 - \exp(-c_e)}\right) \nonumber \\
     &\quad + \left(T + \frac{1}{B_1}\right) a_e \exp(- c_e (s_0+1))\left( (s_0+1)(B_0+\alpha s_0) + \frac{2\alpha(s_0+1)+ B_0}{(1- \exp(-c_e))^2} \right) \nonumber \\
     &\leq \left(T + \frac{1}{B_1}\right) a_e \exp(- c_e (s_0+1)) \left( 2(s_0+1)(B_0+\alpha s_0) + \frac{3\alpha(s_0+1)+ 2 B_0}{(1- \exp(-c_e))^2} \right) . \label{I_2(main)}
\end{align}
Combing \eqref{hawkes approx err decomp(main)} \eqref{I_1(main)} \eqref{I_2(main)}, we have
\begin{align}
    &\quad|\mathbb E[\text{loss}(\hat{\lambda}, S_{test})] - \mathbb E[\text{loss}(\lambda^{\ast}, S_{test})]| \nonumber\\
    &\leq \left(T + \frac{1}{B_1}\right) a_e \exp(- c_e (s_0+1)) \left( 2(s_0+1)(B_0+\alpha s_0) + \frac{3\alpha(s_0+1)+ 2 B_0}{(1- \exp(-c_e))^2} \right) \nonumber \\
    &\quad + \left(T + \frac{1}{B_1}\right) \left( 1+ \frac{a_e \exp(-c_e)}{1 - \exp(-c_e)}\right) \frac{3\mathcal{C} B_0 T^s + 4}{2N^s} . \nonumber
\end{align}
Let $s_0 = \lceil s\log(N)/c_e \rceil$,  and denote $\hat{\lambda}^{N} = \hat{\lambda}$. We have
\begin{align}
    |\mathbb E[\text{loss}(\hat{\lambda}^{N}, S_{test})] - \mathbb E[\text{loss}(\lambda^{\ast}, S_{test})]| \lesssim  \frac{(\log N)^2}{N^s} ~. \nonumber
\end{align}

\textbf{Step 7.} Bound the sizes of the network width and weights.

From step 1-6, the width of the network is less than 
\begin{align}
    3\left\lceil \frac{\tilde{\mathcal{N}}}{2}\right\rceil\binom{\tilde{\mathcal{N}} + 2}{2} + 3\left\lceil \frac{s}{2}\right\rceil + 6N + 2 ~, \nonumber
\end{align}
where $\tilde{\mathcal{N}} = \mathcal{N} + 10(\beta T \vee 1) +  2\lceil\log(3(s_0 + 1))\rceil$. Since $s_0 = \lceil \frac{s}{c_e} \log(N)\rceil$ and $\mathcal{N} = \lceil s\log(N) \rceil$, we have $D \leq O(N) $ .

From the construction of $\hat{g}^{\mathcal{N}}$, $\psi_h$ and $\overline{\lambda}_0^{N}$, the weights of the network is less than 
\begin{align}
    O&\left(\max\left\{\frac{2}{\sigma^{'}(0)h}   \exp(\frac{{\tilde{\mathcal{N}}}^2 + \tilde{\mathcal{N}}  -3Cd\tilde{\mathcal{N}}}{2}) (\tilde{\mathcal{N}}(\tilde{\mathcal{N}}+2))^{3\tilde{\mathcal{N}}(\tilde{\mathcal{N}}+2)},\right.\right.\nonumber\\
    &\quad\quad\quad \left.\left.\left[\frac{\sqrt{2s}5^s}{(s-1)!} B_0 T ^s \right]^{-s/2} N^{(1+s^2)/2} (s(s+2))^{3s(s+2)}\right\}\right) , \nonumber
\end{align}
where $\tilde{\mathcal{N}} = \mathcal{N} + 10(\beta T \vee 1) +  2\lceil\log(3(s_0 + 1))\rceil$. Since $s_0 = \lceil \frac{s}{c_e} \log(N)\rceil$, $h = (12(s_0 + 1))^{-2} N^{-s/2}$,  $\mathcal{N} = \lceil s\log(N) \rceil$, the weights are less than
\begin{align}
    \mathcal{C}_1 (\log(N))^{12s^2(\log(N))^2}~, \nonumber
\end{align}
where $\mathcal{C}_1$ is a constant related to $s, B_0, \alpha , \beta$, and $T$.

\subsection{Proof of Theorem 6}

\begin{lemma}
\label{fourier error bound lemma}
    Suppose $\mu \in C^{k,\infty}([0,T], C_0)$, $k \geq 2$, $k \in \mathbb N$.
    The fourier series of $\mu$ is given by
    \begin{align}
        S_{\infty}(t)  = \frac{\hat{\mu}_0}{2} + \sum_{l = 1}^{\infty} \left(\hat{\mu}_l \cos\left(\frac{2l\pi}{T}t\right) + \hat{\nu}_l \sin\left(\frac{2l\pi}{T}t\right) \right),
        \label{fourier expansion of mu(main)}
    \end{align}
    where $\hat{\mu}_l =2\int_0^T \mu(t) \cos(2l\pi t/T) \mathrm{d}t/T$, $\hat{\nu}_l =2\int_0^T \mu(t) \sin(2l\pi t/T) \mathrm{d}t$/T, $l \geq 0$. If $\mu^{(j)}(0+) = \mu^{(j)}(T-)$, $0\leq j\leq k-1$, then
    \begin{align}
        |\hat{\mu}_l | \leq \frac{2C_0 T^{k}}{(2l\pi)^k},  |\hat{\nu}_l | \leq \frac{2C_0 T^{k}}{(2l\pi)^k} \nonumber
    \end{align}
    and $S_{\infty}(t) = \mu(t)$ on $t\in [0,T]$. Moveover, denote the partial sum of $S_{\infty}(t)$ as $S_{N_\mu}(t) = \hat{\mu}_0/2 + \sum_{l = 1}^{N_\mu} \left(\hat{\mu}_l \cos(2l\pi t/T) + \hat{\nu}_l \sin(2l\pi t/T) \right)$,  
    \begin{align}
        \left|\mu(t) -  S_{N_\mu}(t) \right| \leq  \frac{2C_0T^{k+1}}{(k-1) (2\pi)^k N_\mu^{k-1}}, ~ t \in [0,T]. \nonumber
    \end{align}
\end{lemma}

\begin{proof}[Proof of Lemma \ref{fourier error bound lemma}]
    The proof is a standard Fourier analysis exercise and we omit it. 
\end{proof}

\begin{theorem}
    Under model assumption 5 and $\mu^{(j)}(0+) = \mu^{(j)}(T-)$, $0\leq j\leq k - 1$ , for $N \geq 5$, there exists an RNN structure $\hat{\lambda}^{N, N_\mu}$ as stated in section 2.2
    with $L = 2$, $l_f = B_1$, $u_f = B_0 + O(\log N)$, and input function $x(t;S) = (t, t-F_S(t))^\top$ such that
    \begin{align}
        |\mathbb E[\text{loss}(\hat{\lambda}^{N, N_\mu}, S_{test})] - \mathbb E[\text{loss}(\lambda^{\ast}, S_{test})]| \lesssim  \frac{1}{1-c_\mu} \exp\left( \frac{2 B_0 T}{c_\mu^2}\right)\left(\frac{T^{s} + \log^2 N}{N^s} + \frac{T^k \log N}{N_\mu^{k-1}} \right) ~.\nonumber
    \end{align}
    Moreover, the width of $\tilde{\lambda}^N$  satisfies $D \lesssim N  +N_\mu^5 \log^4 N $ and the weights of $\hat{\lambda}^N$ are less than 
    \begin{align}
    \mathcal{C}_1 (\log(N N_\mu))^{12s^2(\log(N N_\mu))^2}~, \nonumber
    \end{align}
    where $\mathcal{C}_1$ is a constant related to $s, B_0, C_0, c_\mu$, and $T$.
    \label{thm of general case}
\end{theorem}
\begin{proof} [Proof of Theorem \ref{thm of general case}]
     Similar to the proof of Theorem 5
     , the proof is divided into several steps.  Denote $w_l = 2l\pi/T$, $g_{l,1}(x,t) = x_1\cos w_l t + x_2\sin w_l t$,  $g_{l,2}(x,t) =  - x_1\sin w_l t + x_2\cos w_l t + 1$, $g_{l}(x,t) = (g_{l,1}(x,t), g_{l,2}(x,t))^{\top} \in \mathbb R^2$, where $x \in \mathbb R
    ^2$, $l \in \mathbb N_{+}$.  For $l \in \mathbb N_{+}$, define
    \begin{align}
        S_l(t)= \sum_{t_i < t} \begin{pmatrix}
            \sin w_l (t-t_i)
            \\ \cos w_l(t-t_i)
        \end{pmatrix}
        +
        \begin{pmatrix}
            0\\1
        \end{pmatrix}
        , \nonumber
    \end{align}
    and
    \begin{align}
        S_{l,i} = \sum_{0<j<i} \begin{pmatrix}
            \sin w_l (t_i-t_j)
            \\ \cos w_l(t_i-t_j)
        \end{pmatrix}
        +
        \begin{pmatrix}
            0\\1
        \end{pmatrix}
        . \nonumber
    \end{align}
    Hence we have
    \begin{align}
        S_{l,i+1} = 
        \begin{pmatrix}
            \cos w_l (t_{i+1} - t_i) & \sin w_l (t_{i+1} - t_i)\\
            -\sin w_l (t_{i+1} - t_i) & \cos w_l (t_{i+1} - t_i)
        \end{pmatrix}
        S_{l,i}+
        \begin{pmatrix}
            0\\1
        \end{pmatrix}
        = g_{l}(S_{l,i},t_{i+1}-t_{i}) \nonumber
    \end{align}
    and
    \begin{align}
        S_l(t) =  g_{l}(S_{l,i},t-t_{i}), ~t \in (t_{i}, t_{i+1}]. \nonumber
    \end{align}
    where we agree on $S_{l,0} = \mathbf{0}$. Define $S_{0}(t) = \#\{i: t_i < t\}$. If we assume $t_1 > 0$, 
    the true intensity can be rewritten as
    \begin{align}
        \lambda^{\ast}(t) = \lambda_0 (t) +  \frac{\hat{\mu}_0}{2}(S_0(t) - 1) + \sum_{l =1}^{\infty} (\hat{\nu}_l,\hat{\mu}_l) \cdot \left(S_l(t) - 
        \begin{pmatrix}
            0\\1
        \end{pmatrix}
        \right)  , ~ t \in [0,T],
        \label{expansion of true lambda(main)}
    \end{align}
    where $a\cdot b$ refers to the standard inner product of vectors $a$ and $b$.

    We first fix $s_0 \in \mathbb N_{+}$. Since $\mathbb P(t_1 = 0) =0 $. We assume $t_1 > 0$ so that \eqref{expansion of true lambda(main)} holds.

\textbf{Step 1.} Construct the approximation of $g_{l}(x,t) = (g_{l,1}(x,t), g_{l,2}(x,t))^{\top} \in \mathbb R^2$, where $ g_{l,1}, g_{l,2} \in C^{\infty}\left([-3({s_0+1}), 3({s_0+1})]^2 \times [0,T]\right)$ .  Here $x \in \mathbb R^2$.

    Let $\tilde{g}_{l, i}(x,t) = g_{l, i}(3(s_0+1)(2x-1), Tt)$, $i =1,2$.   Then $\tilde{g}_{l, i} \in C^{\infty}\left([0,1]^3 \right)$. By simple computation, we have
    \begin{align}
        \|\tilde{g}_{l, i}\|_{W^{k,\infty}\left( [0,1]^3 \right)} \leq 6({s_0+1})(w_l T)^k. \nonumber
    \end{align}
    Applying Lemma \ref{theorem of tanh NN 2(main)} to $\tilde{g}_{l, i}/[6({s_0+1})]$, for any $\mathcal{N} \in \mathbb N_{+}$, there exists a tanh neural network $\tilde{g}_{l, i}^{\mathcal{N}}$ with only one hidden layer and width $3\lceil (\mathcal{N} + 15w_n T)/2\rceil \binom{\mathcal{N} + 15w_n T + 3}{3}$ such that
    \begin{align}
        \left|\tilde{g}_{l, i}(x,t) - \tilde{g}_{l, i}^{\mathcal{N}}(x,t)\right| \leq 6({s_0+1})\exp(-\mathcal{N}),~ (x,t) \in [0,1]^3. \nonumber
    \end{align}
    
    By coordinate transformation, we get
    \begin{align}
        \left|g_{l,i}(x,t) - \tilde{g}_{l,i}^{\mathcal{N}}\left(\frac{x}{6({s_0+1})}+\frac{1}{2},\frac{t}{T}\right)\right| \leq 6({s_0+1})\exp(-\mathcal{N}),~ (x,y) \in[-3({s_0+1}), 3({s_0+1})]^2 \times [0,T]. \nonumber
    \end{align}
    Define $\hat{g}_{l,i}^{\mathcal{N}}(x,t) = \tilde{g}_{l,i}^{\mathcal{N}}(x/[6({s_0+1})]+1/2,t/T)$, then 
    \begin{align}
        \left|g_{l,i}(x,t) - \hat{g}_{l,i}^{\mathcal{N}}(x,t)\right| \leq  6({s_0+1})\exp(-\mathcal{N}),~ (x,y) \in[-3({s_0+1}), 3({s_0+1})]^2 \times [0,T]. \nonumber
    \end{align}
    Taking $\mathcal{N} \leftarrow \mathcal{N} + \lceil \log (6({s_0+1})) \rceil$, we have
     \begin{align}
        \left|g_{l,i}(x,t) - \hat{g}_{l,i}^{\mathcal{N}}(x,t)\right| \leq \exp(-\mathcal{N}),~ (x,y) \in [-3({s_0+1}), 3({s_0+1})]^2 \times [0,T]. \nonumber
    \end{align}
    Especially, $\left|g_{l,i}(x,t) - \hat{g}_{l,i}^{\mathcal{N}}(x,t)\right| \leq 1$. The width of this NN is bounded by $3\lceil u/2\rceil \binom{u + 3}{3} $.
    From  Lemma \ref{theorem of tanh NN 2(main)} and Remark \ref{the weights of approximation 2(main)}, the weights of $\hat{g}_{l,i}^{\mathcal{N}}$ are bounded by 
    \begin{align}
        O\left( ({s_0+1}) \exp(\frac{{\mathcal{N}^{\prime}}^2 + \mathcal{N}^{\prime}  -3Cd\mathcal{N}^{\prime}}{2}) (\mathcal{N}^\prime(\mathcal{N}^\prime+2))^{3\mathcal{N}^\prime(\mathcal{N}^\prime+2)} \right), \nonumber
    \end{align}
    where $\mathcal{N}^\prime = \mathcal{N} +  \lceil \log (6({s_0+1})) \rceil + 15 w_l T$.
     Since $\hat{g}_{l,i}^{\mathcal{N}} \in \mathbb R$, by a small tuning(precisely, let width plus 1),  we can assume $\hat{g}_{l,i}^{\mathcal{N}}$ has the following structure:
    \begin{align}
        \hat{g}_{l,i}^{\mathcal{N}}(x,y) = V_{l,i}\sigma\left( 
        \begin{pmatrix}
            W_{l,i}& B_{l,i}   
        \end{pmatrix}
        \begin{pmatrix}
            x_{l,i}\\t_{l,i}    
        \end{pmatrix}
        + b_{l,i}\right) . \nonumber
    \end{align}
    Denote $\hat{g}_{l}^{\mathcal{N}}(x,t) = \left(\hat{g}_{l,1}^{\mathcal{N}}(x,t), \hat{g}_{l,2}^{\mathcal{N}}(x,t)\right)^{\top} $.

\textbf{Step 1$^{\prime}$.} Construct the approximation of identity and $g_0(x) = x + 1$, $x \in [-(s_0+1), 2(s_0+1)]$. Here $x \in \mathbb R$.

Similarly to step 3 in the proof of Theorem 5
, taking $\psi_h(x) = 2\sigma\left(hy/2\right)/[\sigma^{'}(0)h]$, we have
\begin{align}
    |x - \psi_h(x)| \leq (6M)^4 h^2, ~ x \in [-M, M]. \nonumber
\end{align}

For $g_0(x) = x + 1$, $x \in [-(s_0+1), 2(s_0+1)]$, we can construct a similar approximation as the proof of of Theorem 5.
There exists a tanh neural network $\hat{g}_0^{\mathcal{N}}$ with only one hidden layer and width $3\lceil (\mathcal{N}^{\prime \prime} + 5)/2\rceil$ such that
\begin{align}
    \left|g_{0}(x) - \hat{g}_{0}^{\mathcal{N}}(x)\right| \leq \exp(-\mathcal{N}),~ x \in[-(s_0+1), 2(s_0+1)], \nonumber
\end{align}
where $\mathcal{N}^{\prime \prime} = \mathcal{N} + \lceil (s_0+3)\log 2\rceil$. The weight of $\hat{g}_0^{\mathcal{N}}$ is bounded by
\begin{align}
     O\left( ({s_0+1}) \exp\left(\frac{{\mathcal{N}^{\prime\prime}}^2 + \mathcal{N}^{\prime\prime}  -3Cd\mathcal{N}^{\prime\prime}}{2}\right) \left[\mathcal{N}^{\prime\prime}(\mathcal{N}^{\prime\prime}+2)\right]^{3\mathcal{N}^{\prime\prime}(\mathcal{N}^{\prime\prime}+2)}  \right). \nonumber
\end{align}

\textbf{Step 2.} Construct the approximation of $S_{n,i}$ and $S_n(t)$ under the event $\{N_e \leq s_0 \}$.

 Let $h_{l,0}^{\mathcal{N}} = (h_{l,0,1}^{\mathcal{N}}, h_{l,0,2}^{\mathcal{N}})^{\top} = \mathbf 0$, $\overline{S}_{l,0}^{\mathcal{N}} = (\overline{S}_{l,0,1}^{\mathcal{N}}, \overline{S}_{l,0,2}^{\mathcal{N}})^{\top} = \mathbf 0$, for $1\leq i \leq s_0$. We construct $h_{l,i}^{\mathcal{N}}$ and $\overline{S}_{l,i}^{\mathcal{N}}$ recursively by
\begin{equation}
    \left\{
    \begin{aligned}
        h_{l,i}^{\mathcal{N}} &= \sigma\left( 
        \begin{pmatrix}
            W_{l,1} 
            \begin{pmatrix}
             V_{l,1}& 0 \\
            0& V_{l,2}
        \end{pmatrix}  \\
            W_{l,2} 
            \begin{pmatrix}
             V_{l,1}& 0 \\
            0& V_{l,2}
        \end{pmatrix}
        \end{pmatrix}
        h_{l,i-1}^{\mathcal{N}}    
        +
        \begin{pmatrix}
             B_{l,1} \\
             B_{l,2}
        \end{pmatrix}
        \begin{pmatrix}
            t_{i}-t_{i-1}    
        \end{pmatrix} 
        +
        \begin{pmatrix}
            b_{l,1}\\b_{l,2}    
        \end{pmatrix}
        \right), \\
        \overline{S}_{l,i}^{\mathcal{N}} &= 
        \begin{pmatrix}
            V_{l,1}& 0 \\
            0& V_{l,2}
        \end{pmatrix}
        h_{l,i}^{\mathcal{N}} . \nonumber
    \end{aligned}
    \right.   
\end{equation}
    Hence $\overline{S}_{l,i}^{\mathcal{N}} = \hat{g}_{l}^{\mathcal{N}}(\overline{S}_{l,i-1}^{\mathcal{N}},t_i-t_{i-1}), 1 \leq i \leq s_0$. Here we agree on $t_0 = 0$.
    
    Similarly, we can define $\overline{S}_{l}^{\mathcal{N}}(t), t \in (t_{i-1}, t_i]$ by
    \begin{equation}
    \left\{
    \begin{aligned}
        h_{l}^{\mathcal{N}} (t) &= \sigma\left( 
        \begin{pmatrix}
            W_{l,1} 
            \begin{pmatrix}
             V_{l,1}& 0 \\
            0& V_{l,2}
        \end{pmatrix}  \\
            W_{l,2} 
            \begin{pmatrix}
             V_{l,1}& 0 \\
            0& V_{l,2}
        \end{pmatrix}
        \end{pmatrix}
        h_{l,i-1}^{\mathcal{N}}    
        +
        \begin{pmatrix}
             B_{l,1} \\
             B_{l,2}
        \end{pmatrix}
        \begin{pmatrix}
            t_{i}-t_{i-1}    
        \end{pmatrix} 
        +
        \begin{pmatrix}
            b_{l,1}\\b_{l,2}    
        \end{pmatrix}
        \right), \\
        \overline{S}_{l}^{\mathcal{N}}(t) &= 
        \begin{pmatrix}
            V_{l,1}& 0 \\
            0& V_{l,2}
        \end{pmatrix}
        h_{l}^{\mathcal{N}} (t). \nonumber
    \end{aligned}
    \right. 
    \end{equation} 
    Hence $\overline{S}_{l}^{\mathcal{N}}(t) = \hat{g}_{l}^{\mathcal{N}}(\overline{S}_{l,i-1}^{\mathcal{N}},t-t_{i-1}), t \in (t_{i-1}, t_i]$.  The approximation error can be bounded by 
    \begin{align}
        &\quad\left\|S_{l}(t) - \overline{S}_{l}^{\mathcal{N}}(t)\right\|_{2}\nonumber\\
        &= \left\| g_{l}(S_{l,i-1},t-t_{i-1}) - \hat{g}_{l}^{\mathcal{N}}(\overline{S}_{l,i-1}^{\mathcal{N}},t-t_{i-1})\right\|_{2} \nonumber \\
        &\leq \left\| g_{l}(S_{l,i-1},t-t_{i-1}) - g_{l}(\overline{S}_{l,i-1}^{\mathcal{N}},t-t_{i-1}) \right\|_{2} + \left\| g_{l}(\overline{S}_{l,i-1}^{\mathcal{N}},t-t_{i-1}) - \hat{g}_{l}^{\mathcal{N}}(\overline{S}_{l,i-1}^{\mathcal{N}},t-t_{i-1})\right\|_{2} \nonumber\\
        &\leq  \left\|S_{l,i-1} - \overline{S}_{l,i-1}^{\mathcal{N}} \right\|_{2} + \sqrt{2} \max \left\{\left\|g_{l,1} - \hat{g}_{l,1}^{\mathcal{N}} \right\|_{\infty} \vee \left\|g_{l,2} - \hat{g}_{l,2}^{\mathcal{N}} \right\|_{\infty} \right\}\nonumber \\
        &\leq \left\|S_{l,i-1} - \overline{S}_{l,i-1}^{\mathcal{N}} \right\|_{2} + \sqrt{2} \exp(-\mathcal{N}) \label{induction eq(main)} \\
        &\leq \cdots \nonumber \\
        &\leq \sqrt{2} i \exp(-\mathcal{N}), ~ t\in (t_{i-1}, t_{i}]. \nonumber
    \end{align}

    Under the event $\{N_e \leq s_0\}$, we have
    \begin{align}
         \left\|S_{l}(t) - \overline{S}_{l}^{\mathcal{N}}(t)\right\|_{2} \leq \sqrt{2}(s_0+1) \exp(-\mathcal{N}). \nonumber
    \end{align}
    Moreover, $\left\|\overline{S}_{l,i}^{\mathcal{N}}\right\|_{2} \leq \left\|S_{l,i} - \overline{S}_{l,i}^{\mathcal{N}}\right\|_{2} + \left\|S_{l,i}\right\|_2 \leq \sqrt{2}(s_0+1) + (s_0 + 1) \leq 3(s_0+1)$, $i \leq s_0 $, then $\overline{S}_{l,i}^{\mathcal{N}} \in [-3(s_0+1), 3(s_0 + 1)]^2$ and \eqref{induction eq(main)} can be verified by induction under the event $\{N_e \leq s_0\}$.

    For the approximation of $S_0(t)$, we can similarly construct a simple RNN such that $\overline{S}_{0,i}^{\mathcal{N}} = \hat{g}_0^{\mathcal{N}} (\overline{S}_{0,i-1}^{\mathcal{N}})$ and $\left|S_{0}(t) - \overline{S}_{0}^{\mathcal{N}}(t)\right| \leq (s_0+1) \exp(-\mathcal{N})$.

\textbf{Step 3.} Construct the approximation of $\lambda^{\ast}(t)$ under the event $\{N_e \leq s_0 \}$.


Since $\lambda_0 \in W^{s, \infty}([0,T], B_0)$,  from the proof of Theorem 4
, there exists a two layer tanh neural network $\overline{\lambda}_0^N$ with width less than $3\lceil s/2\rceil + 6N$ such that
    \begin{align}
        \left|\overline{\lambda}_0^N(t) - \lambda_0(t) \right| \leq \frac{3\mathcal{C} B_0 T^s}{2N^s}, ~\forall 0 \leq t \leq T.
        \label{est err of lambda_0 (in general case)(main)}
    \end{align}
    Moreover, the  weights of $\overline{\lambda}_0^N$ can be bounded by 
    \begin{align}
        O\left( \left[\frac{\sqrt{2s}5^s}{(s-1)!} B_0 T ^s \right]^{-\frac{s}{2}} N^{(1+s^2)/2} (s(s+2))^{3s(s+2)}\right)~. \nonumber
    \end{align}
    Here we assume $\overline{\lambda}_0^N(t)$ have the following structure
    \begin{align}
        \overline{\lambda}_0^N(t) = V_2^{'} \sigma \left( V_1^{'} \sigma\left( B^{'} t
        + b_0^{'}\right) + b_1^{'} \right) + b_2^{'} ~. \nonumber
    \end{align}
    Since $\lambda^{\ast}(t) = \lambda_0 (t) +  \hat{\mu}_0(S_0(t) - 1)/2 + \sum_{l =1}^{\infty} (\hat{\nu}_l,\hat{\mu}_l) \cdot \left(S_l(t) - 
        \begin{pmatrix}
            0\\1
        \end{pmatrix}
        \right)$, we can construct its (finite sum) approximation by
    \begin{align}
        \overline{\lambda}(t) = \overline{\lambda}_0^N(t) + \frac{\hat{\mu}_0}{2}(\psi_h(\overline{S_0}^{\mathcal{N}}(t)) - 1) + \sum_{l =1}^{N_\mu} (\hat{\nu}_l,\hat{\mu}_l) \cdot \left(\psi_h(\overline{S_l}^{\mathcal{N}}(t)) - 
        \begin{pmatrix}
            0\\1
        \end{pmatrix}
        \right)   ~.\nonumber
    \end{align}
    It can be seen as a parallelism of $( N_\mu + 2)$ RNNs defined before.

    Under the event $\{ N_e \leq s_0 \}$, we have $B_1 \leq \lambda(t) \leq B_0 + C_0 s_0$. Recall that $f(x) = \min\{\max\{x, l_f\}, u_f\}$.  Here we can take $l_f = B_1$, $u_f = B_0 + C_0 s_0$. The final output is $\hat{\lambda}(t; S) = f(\overline{\lambda}(t; S))$.

\textbf{Step 4.} Compute the approximation error under the event $\{N_e \leq s_0 \}$.

    Under the event $\{N_e \leq s_0 \}$ and the construction of $f$, we have
    \begin{eqnarray}
        \| \lambda^{\ast} - \hat{\lambda} \|_{L^\infty} \leq \| \lambda^{\ast} - \overline{\lambda} \|_{L^\infty} ~.
        \label{general case first ineq of lambda(main)}
    \end{eqnarray}
    By the construction of $\overline{\lambda}$, 
    \begin{align}
        \left|\lambda^{\ast} (t) - \overline{\lambda}(t)\right| \leq &\left|\lambda_0(t) - \overline{\lambda}_0^N(t)\right| + \frac{\hat{\mu}_0}{2} \left|S_0(t) - \psi_h(\overline{S}_0^{\mathcal{N}}(t))\right| + \left|\sum_{l =1}^{N_\mu} (\hat{\nu}_l,\hat{\mu}_l) \cdot \left(S_l(t) - \psi_h(\overline{S_l}^{\mathcal{N}}(t))\right)\right| \nonumber \\
        &+ \left| \sum_{l> N_\mu} (\hat{\nu}_l,\hat{\mu}_l) \cdot \left(S_l(t) - 
        \begin{pmatrix}
            0\\1
        \end{pmatrix}
        \right) \right| ~.
        \label{err decomp. of lambda*ast (general case)(main)}
    \end{align}
    Under the event $\{N_e \leq s_0\}$, for the second term, we have
    \begin{align}
        \left|S_0(t) - \psi_h(\overline{S}_0^{\mathcal{N}}(t))\right| &\leq \left|S_0(t) - \overline{S}_0^{\mathcal{N}}(t)\right| + \left|\overline{S}_0^{\mathcal{N}}(t) - \psi_h(\overline{S}_0^{\mathcal{N}}(t))\right| \nonumber\\
        &\leq  (s_0 + 1) \left\|g_0 - \hat{g}_0^{\mathcal{N}}\right\|_\infty + (6M)^4 h^2 \nonumber \\
        &\leq (s_0 + 1) \exp(-\mathcal{N}) + (18(s_0+1))^4 h^2, ~ 0 \leq t \leq T.
        \label{err of second term(main)}
    \end{align}
    For the third term,  similarly,
    \begin{align}
        \left|\sum_{l =1}^{N_\mu} (\hat{\nu}_l,\hat{\mu}_l) \cdot \left(S_l(t) - \psi_h(\overline{S_l}^{\mathcal{N}}(t))\right)\right| &\leq \sum_{l=1}^{N_\mu} \left\|(\hat{\nu}_l,\hat{\mu}_l)^{\top}\right\|_2 \left\|S_l(t) - \psi_h(\overline{S_l}^{\mathcal{N}}(t))\right\|_2 \nonumber\\
        &\leq \sum_{l=1}^{N_\mu}  \sqrt{2} C_0 \left(\left\|S_l(t) - \overline{S_l}^{\mathcal{N}}(t)\right\|_2 + \left\|\overline{S_l}^{\mathcal{N}}(t) - \psi_h(\overline{S_l}^{\mathcal{N}}(t))\right\|_2\right) \nonumber\\
        &\leq 2C_0 N_\mu\left((s_0+1) \exp(-\mathcal{N}) + (18(s_0+1))^4 h^2 \right),  ~ 0 \leq t \leq T,
        \label{err of third term(main)}
    \end{align}
     where we take $M = 3(s_0 + 1)$ in \eqref{err of second term(main)} and \eqref{err of third term(main)} to ensure that $\overline{S}_0^{\mathcal{N}}(t)$ and $\overline{S}_l^{\mathcal{N}}(t)$ can be well approximated by $\psi_h(\overline{S}_0^{\mathcal{N}}(t))$ and $\psi_h(\overline{S}_l^{\mathcal{N}}(t))$.
     
     For the fourth term, using Lemma \ref{fourier error bound lemma},
     \begin{align}
         \left| \sum_{l> N_\mu} (\hat{\nu}_l,\hat{\mu}_l) \cdot \left(S_l(t) - 
        \begin{pmatrix}
            0\\1
        \end{pmatrix}
        \right) \right| & = \sum_{t_i < t} \left| \mu(t-t_i) - S_{N_\mu}(t-t_i)\right| \nonumber \\
        &\leq s_0 \frac{4C_0T^{k}}{(k-1) (2\pi)^k N_\mu^{k-1}}. \nonumber
     \end{align}
    Here $S_{N_\mu}(t)$ is the finite sum of the fourier series defined in Lemma \ref{fourier error bound lemma} .

    Finally, by \eqref{est err of lambda_0 (in general case)(main)}, we have $|\overline{\lambda}_0^N(t) - \lambda_0(t)| \leq 3\mathcal{C} B_0 T^s/(2N^s), ~ 0 \leq t \leq T$. To trade off the error terms in \eqref{err decomp. of lambda*ast (general case)(main)}, take $\mathcal{N} = \lceil \log ((s_0+1) N^s N_\mu ) \rceil$ and $h = (18(s_0+1))^{-2} N^{-s/2}N_\mu^{-1/2}$. Then under the event $\{N_e \leq s_0 \}$, we have 
    \begin{align}
        \left|\lambda^{\ast} (t) - \overline{\lambda}(t)\right| &\leq \frac{2C_0 N_\mu+1}{N^s N_\mu} + s_0 \frac{4C_0T^{k}}{(k-1) (2\pi)^k N_\mu^{k-1}} + \frac{3\mathcal{C} B_0 T^s}{2N^s} \nonumber\\
        &\leq \frac{3\mathcal{C} B_0 T^s + 4 C_0+2}{2N^s} + s_0 \frac{4C_0T^{k}}{(k-1) (2\pi)^k N_\mu^{k-1}} , ~ t \in [0,T].
        \label{general case: error term in step 4(main)}
    \end{align}

\textbf{Step 5.} Compute the final approximation error.

By \eqref{hawkes approx err decomp(main)},
\begin{align}
    &\quad|\mathbb E[\text{loss}(\hat{\lambda}, S_{test})] - \mathbb E[\text{loss}(\lambda^{\ast}, S_{test})]| \nonumber\\
    &\leq \mathbb  E\left[ (T + \frac{N_e}{B_1}) \left\|\hat{\lambda} - \lambda^{\ast}\right\|_{L^\infty} \mathbbm{1}_{\{ N_e \leq s_0 \}} \right] + \mathbb E \left[ (T + \frac{N_e}{B_1}) \left\|\hat{\lambda} - \lambda^{\ast}\right\|_{L^\infty} \mathbbm{1}_{\{ N_e > s_0 \}} \right] \nonumber \\
    & := \mathbb{I}_1 + \mathbb{I}_2 ~.
    \label{general case approx err decomp(main)}
\end{align}
Taking $\eta = (c_\mu + 1)/(2 c_\mu)$ in  
Lemma 2
, we have
\begin{align}
    \mathbb P\left(N_e \geq s\right) &\leq \frac{2\sqrt{B_0 T}}{1-c_\mu} \exp\left( \frac{\log(\eta)}{2} \left[\eta(B_0 T) - (1-c_\mu\eta)s\right] \right) \nonumber\\
    &\leq \frac{2\sqrt{B_0 T}}{1- c_\mu} \exp\left( \frac{\log \left(\frac{c_\mu + 1}{2 c_\mu}\right)}{2} \left[\frac{c_\mu + 1}{2 c_\mu}(B_0 T) - \frac{1-c_\mu}{2}s\right] \right) \nonumber \\
    &:= a_e \exp\left( - c_e s \right) ~.\nonumber
\end{align}
By \eqref{general case first ineq of lambda(main)} and \eqref{general case: error term in step 4(main)}, 
\begin{align}
    \mathbb{I}_1 &\leq \left(T + \frac{1}{B_1}\right) \| \lambda^{\ast} - \hat{\lambda} \|_{L^\infty} \mathbb E \left[(N_e + 1) \mathbbm{1}_{\{ N_e \leq s_0 \}} \right] \nonumber\\
    &\leq\left(T + \frac{1}{B_1}\right) \| \lambda^{\ast} - \overline{\lambda} \|_{L^\infty} \mathbb E \left[(N_e + 1)  \right] \nonumber \\
    &= \left(T + \frac{1}{B_1}\right) \| \lambda^{\ast} - \overline{\lambda} \|_{L^\infty} \left( 1+ \sum_{s=1}^{\infty} \mathbb P(N_e \geq s)\right) \nonumber \\
    &\leq\left(T + \frac{1}{B_1}\right) \left( 1+ \frac{a_e \exp(-c_e)}{1 - \exp(-c_e)}\right) \left(\frac{3\mathcal{C} B_0 T^s + 4 C_0 +2}{2N^s} + s_0 \frac{4C_0T^{k}}{(k-1) (2\pi)^k N_\mu^{k-1}}\right) . \label{general case I_1(main)}
\end{align}

Since $\|\hat{\lambda}\|_{L^\infty} \leq B_0 + C_0 s_0$ and $\|\lambda^{\ast}\|_{L^\infty} \leq B_0 + C_0 N_e$, similar to
\eqref{I_2(main)}, we have
\begin{align}
     \mathbb{I}_2 &\leq \mathbb E \left[ \left(T + \frac{N_e}{B_1}\right) \|\hat{\lambda} \|_{L^\infty} \mathbbm{1}_{\{ N_e > s_0 \}} \right] + \mathbb E \left[ \left(T + \frac{N_e}{B_1}\right) \left\| \lambda^{\ast}\right\|_{L^\infty} \mathbbm{1}_{\{ N_e > s_0 \}} \right] \nonumber \\
     &\leq \left(T + \frac{1}{B_1}\right) (B_0 + C_0 s_0)  \mathbb E \left[(N_e+1)\mathbbm{1}_{\{ N_e > s_0 \}} \right] + \left(T + \frac{1}{B_1}\right) E \left[(N_e+1) (B_0 + C_0 N_e)\mathbbm{1}_{\{ N_e > s_0 \}} \right] \nonumber \\
     &\leq \left(T + \frac{1}{B_1}\right) a_e \exp(- c_e (s_0+1)) \left( 2(s_0+1)(B_0+C_0 s_0) + \frac{3C_0(s_0+1)+ 2 B_0}{(1- \exp(-c_e))^2} \right) . \label{general case I_2(main)}
\end{align}
Combining \eqref{general case approx err decomp(main)}, \eqref{general case I_1(main)}, and \eqref{general case I_2(main)}, we have
\begin{align}
    &\quad|\mathbb E[\text{loss}(\hat{\lambda}, S_{test})] - \mathbb E[\text{loss}(\lambda^{\ast}, S_{test})]| \nonumber\\
    &\leq \left(T + \frac{1}{B_1}\right) a_e \exp(- c_e (s_0+1)) \left( 2(s_0+1)(B_0+C_0 s_0) + \frac{3C_0(s_0+1)+ 2 B_0}{(1- \exp(-c_e))^2} \right) \nonumber \\
    &\quad + \left(T + \frac{1}{B_1}\right) \left( 1+ \frac{a_e \exp(-c_e)}{1 - \exp(-c_e)}\right) \left(\frac{3\mathcal{C} B_0 T^s + 4 C_0 + 2}{2N^s} + s_0 \frac{4C_0T^{k}}{(k-1) (2\pi)^k N_\mu^{k-1}}\right)  . \nonumber
\end{align}
Let $s_0 = \lceil s\log(N)/c_e \rceil$ and denote $\hat{\lambda}^{N,N_\mu} = \hat{\lambda}$. We have
\begin{align}
    |\mathbb E[\text{loss}(\hat{\lambda}^{N, N_\mu}, S_{test})] - \mathbb E[\text{loss}(\lambda^{\ast}, S_{test})]| \lesssim  \frac{1}{1-c_\mu} \exp\left( \frac{2 B_0 T}{c_\mu^2}\right)\left(\frac{T^{s} + \log^2 N}{N^s} + \frac{T^k \log N}{N_\mu^{k-1}} \right)  . \nonumber
\end{align}

\textbf{Step 6.} Bound the sizes of the network width and weights.

From step 1-5, we have the width of the network being less than
\begin{align}
    \left(3\lceil \frac{\mathcal{N}^\prime}{2}\rceil \binom{\mathcal{N}^\prime + 3}{3} \right) 2 N_\mu + 3\Big\lceil\frac{s}{2}\Big\rceil + 6N + 3 \lceil\frac{\mathcal{N}^{\prime\prime}+5}{2}\rceil \nonumber
\end{align}
where $\mathcal{N}^\prime = \mathcal{N} +  \lceil \log (6({s_0+1})) \rceil + 15 w_{N_\mu} T$,  $\mathcal{N}^{\prime\prime} = \mathcal{N} + \lceil (s_0+3)\log 2\rceil$, $\mathcal{N} = \lceil \log ((s_0+1) N^s N_\mu ) \rceil$,  $s_0 = \lceil s\log(N)/c_e\rceil$. Hence
\begin{align}
    D \lesssim N  +N_\mu^5 \log^4 N  ~. \nonumber
\end{align}
From the construction of $\hat{g}_{l,i}^{\mathcal{N}}$, $\hat{g}_0^{\mathcal{N}}$, $\psi_h$, $\overline{\lambda}_0^N$, the weights of the network is less than
\begin{align}
    \mathcal{C}_1^{\prime}\max&\left\{\left( ({s_0+1}) \exp(\frac{{\mathcal{N}^{\prime}}^2 + \mathcal{N}^{\prime}  -3Cd\mathcal{N}^{\prime}}{2}) (\mathcal{N}^\prime(\mathcal{N}^\prime+2))^{3\mathcal{N}^\prime(\mathcal{N}^\prime+2)} \right),\right.\nonumber\\
    &\quad \left( ({s_0+1}) \exp(\frac{{\mathcal{N}^{\prime\prime}}^2 + \mathcal{N}^{\prime\prime}  -3Cd\mathcal{N}^{\prime\prime}}{2}) (\mathcal{N}^{\prime\prime}(\mathcal{N}^{\prime\prime}+2))^{3\mathcal{N}^{\prime\prime}(\mathcal{N}^{\prime\prime}+2)} \right), \frac{2}{\sigma^{'}(0) h}, \nonumber\\
    &\quad \left. \left( \left[\frac{\sqrt{2s}5^s}{(s-1)!} B_0 T ^s \right]^{-s/2} N^{(1+s^2)/2} (s(s+2))^{3s(s+2)}\right) \right\}, \nonumber
\end{align}
where $h = (18(s_0+1))^{-2} N^{-s/2}N_\mu^{-1/2}$. Hence the weights of the network is less than 
\begin{align}
    \mathcal{C}_1 (\log(N N_\mu))^{12s^2(\log(N N_\mu))^2}~, \nonumber
\end{align}
where $\mathcal{C}_1$ is a constant related to $s, B_0, C_0, c_\mu$, and $T$.
\end{proof}

\begin{lemma}
    Let $\delta_j = \frac{j}{k}$, $1 \leq j \leq k$, $\delta = (\delta_1, \delta_2, \cdots, \delta_k)^{\top}$ and 
    \begin{align}
        V_\delta = \begin{pmatrix} 1 & 1 &\cdots&1 \\ \delta_1 & \delta_2  &\cdots&\delta_k \\ \delta_1^2 &\delta_2^2 &\cdots&\delta_k^2 \\\vdots&\vdots&\ddots&\vdots\\\delta_1^{k-1}&\delta_2^{k-1}&\cdots &\delta_k^{k-1}
        \end{pmatrix} \nonumber
    \end{align}
    then $V_\delta$ is invertible and $\|V_\delta^{-1}\|_{\infty} \leq C *8^{k}$, where $C$ is a universal constant.
    \label{vander monde lemma(main)}
\end{lemma}
\begin{proof}[Proof of Lemma \ref{vander monde lemma(main)}]
    See \cite{Gautschi1990HowA}.
\end{proof}

\begin{lemma}
     For $\mu \in C^{k,\infty}([0,T], C_0)$ and $\delta = (\delta_1, \delta_2, \cdots, \delta_k)^{\top}$ defined in Lemma \ref{vander monde lemma(main)}, there exists $\alpha = (\alpha_1, \alpha_2, \cdots, \alpha_k)^{\top}$ such that
     \begin{align}
         \tilde{\mu}(t) := \mu(t) +\sum_{j = 1}^{k} \alpha_j \exp(-\delta_j t) 
         \label{tilde mu eq(main)}
     \end{align}
     satisfing $\|\alpha\|_\infty \leq 2C_0 C 8^k/(1-\exp(-T))$,  $\tilde{\mu} \in C^{k,\infty}([0,T], C_0 + k \|\alpha\|_\infty)$ and $\tilde{\mu}^{(j)}(0+) = \tilde{\mu}^{(j)}(T-)$, $0\leq j\leq k-1$, where the constant $C$ is defined in Lemma \ref{vander monde lemma(main)}.
     \label{interpolation lemma(main)}
\end{lemma}
\begin{proof}[Proof of Lemma \ref{interpolation lemma(main)}]
    We only need to solve the following equations:
    \begin{align}
        \tilde{\mu}^{(j)}(0+) = \tilde{\mu}^{(j)}(T-), ~0\leq j\leq k-1.
        \label{interpolation eq(main)}
    \end{align}
    In matrix form,
    \
    \begin{align}
         & ~~~~ \begin{pmatrix} 1 - e^{-\delta_1 T} & 1 - e^{-\delta_2 T} &\cdots&1 -e^{-\delta_k T} \\ (-\delta_1)(1 - e^{-\delta_1 T}) & (-\delta_2)(1 - e^{-\delta_2 T})  &\cdots&(-\delta_k)(1 -e^{-\delta_k T}) \\ 
         \vdots&\vdots&\ddots&\vdots\\(-\delta_1)^{k-1}(1 - e^{-\delta_1 T}) & (-\delta_2)^{k-1}(1 - e^{-\delta_2 T})  &\cdots&(-\delta_k)^{k-1}(1 - e^{-\delta_k T} ) 
        \end{pmatrix}
        \begin{pmatrix}
            \alpha_1\\\alpha_2\\\vdots\\\alpha_k
        \end{pmatrix}
        \nonumber \\
        & =
        \begin{pmatrix}
            {\mu}(T-) - {\mu}(0+)\\{\mu}^{(1)}(T-) - {\mu}^{(1)}(0+)\\\vdots\\{\mu}^{(k-1)}(T-) - {\mu}^{(k-1)}(0+)
        \end{pmatrix} .
        \label{matrix form of interpolation : long version(main)}
    \end{align}
    Rewrite \eqref{matrix form of interpolation : long version(main)} as
    \begin{align}
        D V_\delta \Lambda_\delta \alpha = \Delta_\mu , \nonumber
    \end{align}
    where $D = \mathrm{diag}\{1, -1, \cdots, (-1)^{k-1}$, $\Lambda_\delta = \mathrm{diag}\{1-e^{-\delta_1 T}, 1-e^{-\delta_2 T}, \cdots, 1-e^{-\delta_k T}\}$, $\Lambda_\mu = ({\mu}(T-) - {\mu}(0+), {\mu}^{(1)}(T-) - {\mu}^{(1)}(0+), \cdots, {\mu}^{(k-1)}(T-) - {\mu}^{(k-1)}(0+))^{\top}$, and $V_\delta$ is defined in Lemma \ref{vander monde lemma(main)}. By Lemma \ref{vander monde lemma(main)} and $\delta_j = j/k$, $1\leq j \leq k$, we have  $D V_\delta \Lambda_\delta$ is invertible and 
    \begin{align}
        \|\alpha\|_\infty &\leq \|D^{-1}\|_\infty \|V_\delta^{-1}\|_\infty  \|\Lambda_\delta^{-1}\|_\infty \|\Delta_\mu\|_\infty \nonumber\\
        &\leq(C * 8^k) \frac{1}{(1-\exp(-T))}  (2C_0) = \frac{2C_0 C 8^k}{1-\exp(-T)}, \nonumber
    \end{align}
    where the constant $C$ is defined in Lemma \ref{vander monde lemma(main)}. By \eqref{tilde mu eq(main)}, we have $\tilde{\mu} \in C^{k,\infty}([0,T], C_0 + k \|\alpha\|_\infty)$.
\end{proof}


Now we prove Theorem 6.
    The proof is based on Theorem 5, Theorem \ref{thm of general case}, and Lemma \ref{interpolation lemma(main)}.
    From Lemma \ref{interpolation lemma(main)}, for $\mu \in C^{k,\infty}([0,T], C_0)$, there exists $\alpha = (\alpha_1, \alpha_2, \cdots, \alpha_k)^{\top} \in \mathbb R^{k}$ such that $\tilde{\mu}(t) := \mu(t) +\sum_{j = 1}^{k} \alpha_j \exp(-\delta_j t)$ satisfying the boundary condition $\tilde{\mu}^{(j)}(0+) = \tilde{\mu}^{(j)}(T-)$, $0\leq j\leq k-1$, and we have $\tilde{\mu} \in C^{k,\infty}([0,T], C_0 + k \frac{2C_0 C 8^k}{1-\exp(-T)})$. Define $\tilde{\nu}(t) := \mu(t) - \tilde{\mu}(t) = - \sum_{j=1}^{k} \alpha_j \exp(-\delta_j t)$. Denote 
    \begin{align}
        \lambda^{\ast}_1 (t) &:= \lambda_0(t) + \sum_{t_i < t} \tilde{\mu}(t - t_i), \nonumber\\
        \lambda^{\ast}_2 (t) &:= \sum_{t_i < t}  \tilde{\nu}(t - t_i) = \sum_{j=1}^{k} \sum_{t_i < t} (-\alpha_j) \exp(-\delta_j(t-t_i)) := \sum_{j=1}^{k}\lambda_{2j}^{\ast}(t),\nonumber
    \end{align}
    and then $\lambda^{\ast} (t) = \lambda^{\ast}_1 (t) + \lambda^{\ast}_2(t)$.

    Fix $s_0 \in \mathbb N_{+}$. By the proof of Theorem \ref{thm of general case}, under the event $\{N_e \leq s_0 \}$, there exists an RNN (without the output layer) $\overline{\lambda}_1(t)$ such that
    \begin{align}
        \left|\lambda_1^{\ast} (t) - \overline{\lambda}_1(t)\right| \leq \frac{3\mathcal{C} B_0 T^s + 4 \tilde{C_0}+2}{2N^s} + s_0 \frac{4\tilde{C_0}T^{k}}{(k-1) (2\pi)^k N_\mu^{k-1}} , ~ t \in [0,T], \nonumber
    \end{align}
    where $\tilde{C_0} =  C_0 + 2kC_0 C 8^k/(1-\exp(-T))$.

    By the proof of Theorem 5
    , under the event $\{N_e \leq s_0 \}$, for$1\leq j \leq k$, there exists an RNN (without the output layer) $\overline{\lambda}_{2j}(t)$ such that
    \begin{align}
        \left|\lambda_{2j}^{\ast} (t) - \overline{\lambda}_{2j}(t)\right| \leq \frac{2\alpha_j}{N^s} \leq \frac{4C_0 C 8^k}{(1-\exp(-T)) N^s} , t \in [0,T]. \nonumber
    \end{align}
    Let $\overline{\lambda}_{2}(t) = \sum_{j=1}^{k} \overline{\lambda}_{2j}(t)$. We have
    \begin{align}
        \left|\lambda_{2}^{\ast} (t) - \overline{\lambda}_{2}(t)\right| \leq  \frac{2(\tilde{C}_0 - C_0)}{N^s} , t \in [0,T]. \nonumber
    \end{align}
    Let $\overline{\lambda}(t) = \overline{\lambda}_1(t) + \overline{\lambda}_2(t)$, 
    \begin{align}
        \left|\lambda^{\ast} (t) - \overline{\lambda}(t)\right| \leq \left|\lambda_1^{\ast} (t) - \overline{\lambda}_1(t)\right| +  \left|\lambda_{2}^{\ast} (t) - \overline{\lambda}_{2}(t)\right| \leq \frac{3\mathcal{C} B_0 T^s + 8 \tilde{C_0}+2}{2N^s} + s_0 \frac{4\tilde{C_0}T^{k}}{(k-1) (2\pi)^k N_\mu^{k-1}}. \nonumber
    \end{align}
    Under the event $\{N_e \leq s_0\}$,  $B_1 \leq \lambda^{\ast} \leq B_0 + C_0 s_0$. Hence we can take $l_f = B_1$ and $u_f = B_0 +  C_0 s_0$ and denote $\hat{\lambda}(t) = f(\overline{\lambda}(t))$. Then $\| \lambda^{\ast} - \hat{\lambda} \|_\infty \leq \| \lambda^{\ast} - \overline{\lambda} \|_\infty$.
    By similar arguments in Theorem \ref{thm of general case}, we have
    \begin{align}
        &\quad|\mathbb E[{\text{loss}}(\hat{\lambda}, S_{test})] - \mathbb E[{\text{loss}}(\lambda^{\ast}, S_{test})]| \nonumber\\
        &\leq \left(T + \frac{1}{B_1}\right) a_e \exp(- c_e (s_0+1)) \left( 2(s_0+1)(B_0+C_0 s_0) + \frac{3C_0(s_0+1)+ 2 B_0}{(1- \exp(-c_e))^2} \right) \nonumber \\
    &\quad + \left(T + \frac{1}{B_1}\right) \left( 1+ \frac{a_e \exp(-c_e)}{1 - \exp(-c_e)}\right) \left(\frac{3\mathcal{C} B_0 T^s + 8 \tilde{C}_0}{2N^s} + s_0 \frac{4\tilde{C}_0T^{k}}{(k-1) (2\pi)^k N_\mu^{k-1}}\right)  ~. \nonumber
    \end{align}
    Let $s_0 = \lceil s \log(N)/c_e\rceil$ and denote $\hat{\lambda}^{N,N_\mu} = \hat{\lambda}$. We have
\begin{align}
    |\mathbb E[\tilde{\text{loss}}(\hat{\lambda}^{N,N_\mu})] - \mathbb E[\tilde{\text{loss}}(\lambda^{\ast})]| \lesssim  \frac{\log^2 N}{N^s} + \frac{\log N}{N_\mu^{k-1}} \nonumber   
\end{align}
The width and elements weights bound can also be obtained similarly to the proof of Theorem \ref{thm of general case}.

\subsection{Proof of Theorem 7}
    Denote $\lambda_1^{*} (t) = \lambda_0(t) + \sum_{t_i < t} \alpha \exp(-\beta(t - t_i))$. Then $\lambda^{*} (t) = \Psi\left( \lambda_1^{*} (t) \right)$. Fix $s_0 \in \mathbb N_{+}$.
    From the proof of Theorem 5
    , under the event $\{N_e \leq s_0 \}$,  there exists a 2-layer recurrent neural network $\overline{\lambda}_1 (t)$ as \eqref{overline_lambda(t)(main)} such that
    \begin{align}
        \left| \overline{\lambda}_1 (t) - \lambda_1^{*} (t)\right| \leq \frac{3\mathcal{C} B_0 T^s + 2}{2N_1^s}, ~ \forall t \in  [0,T].
        \label{approximation error of lambda1(main)}
    \end{align}
    Moreover, the width of $\overline{\lambda}_1 (t)$ satisfies $D \lesssim N_1$ and the weights of $\overline{\lambda}_1 (t)$ are bounded by 
    \begin{align}
        O\left( (\log N_1)^{12s^2(\log N_1)^2} \right)~. \nonumber
    \end{align}
    Under the event $\{N_e \leq s_0\}$, the function $\lambda_1^{*}(t)$ satisfies $0 \leq \lambda_1^{*} \leq B_0 + \alpha s_0$. Using \eqref{approximation error of lambda1(main)} and taking $(3\mathcal{C} B_0 T^s + 2)/2N_1^s \leq 1$, we have $\overline{\lambda}_1 \in [-1,  B_0 + \alpha s_0 + 1]$. Hence we need to construct  an approximation of $\Psi$ on $[-1, B_0 + \alpha s_0 +1]$. Let $\tilde{\Psi}(x) = \Psi(\rho x - 1)$, where $\rho = { B_0 + \alpha s_0 +2}$. Then $\Psi(x) = \tilde{\Psi}((x+1)/\rho)$.
    
    Since $\Psi$ is L-lipschitz and $\tilde{\Psi}$ is defined on $[0,1]$ and $\rho L$-Lipschitz, by the Corollary 5.4 of \cite{de2021approximation}, there exists a tanh neural network $\tilde{\Psi}^{N_2}$ with 2 hidden layers such that
    \begin{align}
        \left\|\tilde{\Psi} - \tilde{\Psi}^{N_2}\right\|_{L^{\infty}[0,1]} \leq \frac{7(\rho L \vee \tilde{B}_0)}{N_2}. \nonumber
    \end{align}
    Let $\Psi^{N_2}(x) = \tilde{\Psi}^{N_2}((x+1)/\rho)$. Then 
    \begin{align}
         \left|\Psi(x) - \Psi^{N_2}(x)\right| \leq \frac{7(\rho L \vee \tilde{B}_0)}{N_2} , ~ x \in [-1, B_0 + \alpha s_0 + 1] . \nonumber
    \end{align}
    Then under the event $\{N_e \leq s_0\}$, we have 
    \begin{align}
        \left|\Psi(\lambda_1^{*}(t)) - \Psi^{N_2}(\overline{\lambda}_1 (t))\right| &\leq \left|\Psi(\lambda_1^{*}(t)) - \Psi(\overline{\lambda}_1 (t))\right| + \left|\Psi(\overline{\lambda}_1 (t)) - \Psi^{N_2}(\overline{\lambda}_1 (t))\right| \nonumber \\
        & \leq L \left| \overline{\lambda}_1 (t) - \lambda_1^{*} (t)\right| + \left\|\Psi - \Psi^{N_2}\right\|_{L^\infty} \nonumber \\
        & \leq L \frac{3\mathcal{C} B_0 T^s + 2}{2N_1^s} + \frac{7(\rho L \vee \tilde{B}_0)}{N_2} .
        \label{nonlinear_case_finite_approximation_without_f(main)}
    \end{align}
    Recall that $f(x) = \min\{\max\{x, l_f\}, u_f\}$. Since $\tilde{B}_1 \leq \Psi \leq \tilde{B}_0$, we can take $l_f = \tilde{B}_1$ and $u_f = \tilde{B}_0$. Define $\hat{\lambda}(t) =  f\left(\Psi^{N_2}(\overline{\lambda}_1 (t))\right)$. We have
    \begin{align}
           \left|\lambda^{*}(t) - \hat{\lambda}(t)\right| \leq \left|\Psi(\lambda_1^{*}(t)) - \Psi^{N_2}(\overline{\lambda}_1 (t))\right| , ~ \forall t \in [0,T].
           \label{nonlinear_case_finite_approximation_with_f(main)}
    \end{align}
    Similar to \eqref{hawkes approx err decomp(main)}, we have
    \begin{align}
        &\quad|\mathbb E[\text{loss}(\hat{\lambda}, S_{test})] - \mathbb E[\text{loss}(\lambda^{\ast}, S_{test})]| \nonumber\\
        &\leq \mathbb E \left|\text{loss}(\hat{\lambda}, S_{test}) - \text{loss}(\lambda^{\ast}, S_{test})\right| \nonumber \\
        &\leq \mathbb E \left(\Big| \sum_{i=1}^{N_e}(\log \hat{\lambda}(t_i) - \log \lambda^{\ast}(t_i)) \Big| + \Big| \int_0^T \left(\hat{\lambda}(t) - \lambda^{\ast}(t)\right) \mathrm{d t} \Big|\right) \nonumber \\
        &\leq \mathbb E \left[ \left(\Big| \sum_{i=1}^{N_e}(\log \hat{\lambda}(t_i) - \log \lambda^{\ast}(t_i)) \Big| + \Big| \int_0^T \left(\hat{\lambda}(t) - \lambda^{\ast}(t)\right) \mathrm{d t}  \Big|\right) \mathbbm{1}_{\{ N_e \leq s_0 \}} \right. \nonumber\\
        & \quad \quad  + \left. \left(\Big| \sum_{i=1}^{N_e}(\log \hat{\lambda}(t_i) - \log \lambda^{\ast}(t_i)) \Big| + \Big| \int_0^T \left(\hat{\lambda}(t) - \lambda^{\ast}(t)\right) \mathrm{d t}  \Big|\right) \mathbbm{1}_{\{ N_e > s_0 \}} \right] \nonumber\\
        &\leq \mathbb  E\left[ (T + \frac{N_e}{\tilde{B_1}}) \left\|\hat{\lambda} - \lambda^{\ast}\right\|_{L^\infty} \mathbbm{1}_{\{ N_e \leq s_0 \}} \right] + \mathbb E \left[ (T + \frac{N_e}{\tilde{B_1}}) \left\|\hat{\lambda} - \lambda^{\ast}\right\|_{L^\infty} \mathbbm{1}_{\{ N_e > s_0 \}} \right] \nonumber \\
        & := \mathbb{I}_1 + \mathbb{I}_2 ~. \label{nonlinear hawkes approx err decomp(main)}
    \end{align}
Since $\Psi \leq \tilde{B_0}$, similar to \eqref{hawkes approx err decomp(main)}, taking $\eta = e$ in 
Lemma 2
, we have
    \begin{align}
        \mathbb P (N_e \geq s) \leq 2\sqrt{\tilde{B_0} T}\exp \left(\frac{e \tilde{B_0} T - s}{2} \right), \nonumber
    \end{align}
and similar to \eqref{case1_thm_expection(main)}, we have
\begin{align}
     \mathbb  E(N_e+1) \leq 1 + \sum_{s=1}^{\infty}  \mathbb P (N_e \geq s) \leq 1 + \frac{2\sqrt{\tilde{B_0} T}}{1-\exp(-1/2)} \exp \left(\frac{e \tilde{B_0} T - 1}{2} \right) \leq {5\sqrt{\tilde{B_0} T + 1}} \exp \left(\frac{3 \tilde{B_0} T}{2} \right).
    \label{case3_thm_expection(main)}
\end{align}
By \eqref{nonlinear_case_finite_approximation_without_f(main)},  \eqref{nonlinear_case_finite_approximation_with_f(main)}, and \eqref{case3_thm_expection(main)}, 
\begin{align}
     \mathbb{I}_1 &\leq \left(T + \frac{1}{\tilde{B_1}}\right)\frac{3\mathcal{C} B_0 T^s + 2}{2N^s} \mathbb E \left[(N_e + 1) \mathbbm{1}_{\{ N_e \leq s_0 \}} \right] \nonumber\\
    &\leq \left(T + \frac{1}{\tilde{B_1}}\right)\frac{3\mathcal{C} B_0 T^s + 2}{2N^s} \mathbb E \left[(N_e + 1)  \right] \nonumber \\
    &\leq \left(T + \frac{1}{\tilde{B_1}}\right) {5\sqrt{\tilde{B_0} T + 1}} \exp \left(\frac{3 \tilde{B_0} T}{2} \right) \left( L \frac{3\mathcal{C} B_0 T^s + 2}{2N_1^s} + \frac{7(\rho L \vee \tilde{B}_0)}{N_2} \right) \nonumber \\
    &\leq 5 L \exp \left({2 \tilde{B_0} T} \right) \left(T + \frac{1}{\tilde{B_1}}\right) \left(  \frac{3\mathcal{C} B_0 T^s + 2}{2N_1^s} + \frac{7(\rho \vee (\tilde{B}_0/L)) }{N_2} \right) .
    \label{ nonlinear I_1(main)}
\end{align}

On the other hand, since $\|\hat{\lambda}\|_{L^\infty} \leq \tilde{B_0} $ and $\|\lambda^{\ast}\|_{L^\infty} \leq \tilde{B_0}$, we have
\begin{align}
    \mathbb{I}_2 &\leq \mathbb E \left[ \left(T + \frac{N_e}{\tilde{B_1}}\right) \|\hat{\lambda} \|_{L^\infty} \mathbbm{1}_{\{ N_e > s_0 \}} \right] + \mathbb E \left[ \left(T + \frac{N_e}{\tilde{B_1}}\right) \left\| \lambda^{\ast}\right\|_{L^\infty} \mathbbm{1}_{\{ N_e > s_0 \}} \right] \nonumber \\
     &\leq 2 \left(T + \frac{1}{\tilde{B_1}}\right) \tilde{B_0}  \mathbb E \left[(N_e+1)\mathbbm{1}_{\{ N_e > s_0 \}} \right] \nonumber \\
     &\leq 2 \left(T + \frac{1}{\tilde{B_1}}\right) \tilde{B_0} \left((s_0+1)\mathbb P (N_e \geq s_0 + 1) + \sum_{s=s_0 + 1}^{\infty} \mathbb P(N_e \geq s)\right) \nonumber \\
     &\leq 4 \left(T + \frac{1}{\tilde{B_1}}\right) \tilde{B_0}  \sqrt{\tilde{B_0} T}\exp \left(\frac{e \tilde{B_0} T - (s_0+1)}{2}\right)\left((s_0+1) + \frac{1}{1-e^{-\frac{1}{2}}}  \right) \nonumber\\
     &\leq 4 \left(T + \frac{1}{\tilde{B_1}}\right) \tilde{B_0}  \sqrt{\tilde{B_0} T}\exp \left(\frac{3 \tilde{B_0} T - (s_0+1)}{2}\right)\left(s_0+4  \right) \nonumber \\
     &\leq 4 \left(T + \frac{1}{\tilde{B_1}}\right) \tilde{B_0} \exp \left({2 \tilde{B_0} T} \right) \left(s_0+4  \right) \exp \left(-\frac{s_0+1}{2}\right) .
     \label{nonlinear I_2(main)}
\end{align}
Combining \eqref{nonlinear hawkes approx err decomp(main)}, \eqref{ nonlinear I_1(main)}, and \eqref{nonlinear I_2(main)}, we have
\begin{align}
    |\mathbb E[\text{loss}(\hat{\lambda}, S_{test})] - \mathbb E[\text{loss}(\lambda^{\ast}, S_{test})]| &\leq 5 L \exp \left({2 \tilde{B_0} T} \right) \left(T + \frac{1}{\tilde{B_1}}\right) \left(  \frac{3\mathcal{C} B_0 T^s + 2}{2N_1^s} + \frac{7(\rho \vee (\tilde{B}_0/L)) }{N_2} \right) \nonumber \\
    &\quad + 4 \left(T + \frac{1}{\tilde{B_1}}\right) \tilde{B_0} \exp \left({2 \tilde{B_0} T} \right) \left(s_0+4  \right) \exp \left(-\frac{s_0+1}{2}\right) . \nonumber
\end{align}
Let $s_0 = \lceil 2 \log N \rceil$, $N_1 = N_2 = N$ and denote $\hat{\lambda}^N = \hat{\lambda}$. We have 
\begin{align}
    |\mathbb E[\text{loss}(\hat{\lambda}^{N}, S_{test})] - \mathbb E[\text{loss}(\lambda^{\ast}, S_{test})]| \lesssim \frac{\log N}{N} . \nonumber
\end{align}
Similar to the proof of Theorem 5, we can bound the width of the network by 
\begin{align}
    \max \left\{  3\left\lceil \frac{\tilde{\mathcal{N}}}{2}\right\rceil\binom{\tilde{\mathcal{N}} + 2}{2} + 3\left\lceil \frac{s}{2}\right\rceil + 6N + 2 , 6N \right\}, \nonumber
\end{align}
where $\tilde{\mathcal{N}} = \lceil s \log (N) \rceil + 10(\delta T \vee 1) +  2\lceil\log(3(s_0 + 1))\rceil$. Hence we have $D \lesssim N$.

Moreover, from the construction of $\hat{\lambda}$, the weights of the network is less than 
\begin{align}
    \mathcal{C}_1^\prime \max \left\{ (\log(N))^{12s^2(\log(N))^2},  \frac{N}{\rho \sqrt{\rho L}}  \right\}, \nonumber
\end{align}
where $\rho = B_0 + \alpha s_0 + 2 = B_0 + \alpha \lceil 2 \log N \rceil + 2$,  $\mathcal{C}_1^\prime$ is a constant related to $s, B_0, \alpha , \delta, T, \tilde{B}_0$, and $L$. Then the weights of the network can be bounded by 
\begin{align}
    \mathcal{C}_1 (\log(N))^{12s^2(\log(N))^2}, \nonumber
\end{align}
where $C_1$ are constants related to $s, B_0, \alpha , \delta, T, \tilde{B}_0$, and $L$.

\subsection{Proof of Theorem 8}
    Without loss of generality, we denote $t_1 = T/3$, $t_2 = 2T/3$ for simplicity. Since the compensator of $N(t)$ is $\Lambda(t) = \int_{0}^{t} \lambda^{\ast}(s) \mathrm{d}s$, for a predictable stochastic process $\lambda(t), t \in [0,T]$,  we have
    \begin{align}
        \mathbb E [\text{loss}(\lambda, S_{test})] &= \mathbb E \left[-\sum_{t_i < T} \log \lambda(t_i) + \int_0^T \lambda(t)\mathrm{d}t \right] \nonumber\\
        &= \mathbb E \left[-\int_0^T \log \lambda(t) \mathrm{d} N(t) + \int_0^T \lambda(t)\mathrm{d}t \right] \nonumber\\
        &=  \mathbb E \left[\int_0^T \left(\lambda(t) - \log \lambda(t) * \lambda^{\ast}(t) \right) \mathrm{d} t  \right] . \nonumber
    \end{align}
    Since both $\lambda^{\ast}$ and
     $\hat{\lambda}_{ne}$ are predictable, we have
     \begin{align}
          &\mathbb E [\text{loss}(\hat{\lambda}_{ne}, S_{test})] - \mathbb E [\text{loss}(\lambda^{\ast}, S_{test})] \nonumber\\
          = & \mathbb E \left[\int_0^T \left(\hat{\lambda}_{ne}(t) - \log \hat{\lambda}_{ne}(t) * \lambda^{\ast}(t) \right) \mathrm{d} t  \right] - E \left[\int_0^T \left(\lambda^{\ast}(t) - \log \lambda^{\ast}(t) * \lambda^{\ast}(t) \right) \mathrm{d} t  \right] \nonumber \\
          := & \mathbb E \left[\int_0^T \left(g (\hat{\lambda}_{ne}(t), \lambda^{\ast}(t)) - g (\lambda^{\ast}(t),\lambda^{\ast}(t))  \right) \mathrm{d} t \right] , \nonumber
     \end{align}
     where $g(x, y) = x - \log x * y \geq y - \log y * y =g(y,y)$, $\forall x, y > 0$, and the equality holds if and only if $x = y$. Thus 
     \begin{align}
         \mathbb E [\text{loss}(\hat{\lambda}_{ne}, S_{test})] - \mathbb E [\text{loss}(\lambda^{\ast}, S_{test})] = \mathbb E \left[\int_0^T \left(g (\hat{\lambda}_{ne}(t), \lambda^{\ast}(t)) - g (\lambda^{\ast}(t),\lambda^{\ast}(t))  \right) \mathrm{d} t \right] \geq 0. \nonumber
     \end{align}

Denote $\mathcal{E} =\{ \text{there is no event in } [0, 2T/3] \}$,  and $ \mathbb P (\mathcal{E}) > 0$. Denote $I_0 = \left[T/3, 2T/3 \right]$. By a similar argument, we have 
\begin{align}
      \mathbb E [\text{loss}(\hat{\lambda}_{ne}, S_{test})] - \mathbb E [\text{loss}(\lambda^{\ast}, S_{test})] \geq \mathbb E \left[\left(\int_{I_0} \left(g (\hat{\lambda}_{ne}(t), \lambda^{\ast}(t)) - g (\lambda^{\ast}(t),\lambda^{\ast}(t))  \right) \mathrm{d} t\right)\mathbbm{1}_{\mathcal{E}}\right].
      \label{induce in I0 in counterex(main)}
\end{align}
Under the event $\mathcal{E}$, 
\begin{align}
    \hat{\lambda}_{ne}(t) = f(\alpha t + b) = \left\{
    \begin{aligned}
        &T &, ~ & \alpha t + b < T \\
        &\alpha t + b &,  ~ & T \leq \alpha t + b \leq 4T\\
        &4T &, ~ &\alpha t + b > 4T  
    \end{aligned}
    \right. ~, ~ t \in I_0, \nonumber
\end{align}
and 
\begin{align}
    &\mathbb E \left[\left(\int_{I_0} \left(g \left(\hat{\lambda}_{ne}(t), \lambda^{\ast}(t)\right) - g \left(\lambda^{\ast}(t),\lambda^{\ast}(t)\right)  \right) \mathrm{d} t\right)\mathbbm{1}_{\mathcal{E}}\right] \nonumber \\
    = & \left[\int_{I_0} \left(g \left(f(\alpha t + b), \frac{9}{T}t^2\right) - g \left(\frac{9}{T}t^2, \frac{9}{T}t^2\right)  \right) \mathrm{d} t\right]\mathbb P(\mathcal{E}) \nonumber \\
    := & F(\alpha, b) P(\mathcal{E}).
    \label{def of F in counterex(main)}
\end{align}
Then we only need to show 
\begin{align}
    \inf_{\alpha \in \mathbb R, b \in \mathbb R} F(\alpha, b) > 0. \nonumber
\end{align}

\textbf{Case 1.} $|\alpha| > 18$, $b \in \mathbb R$.

Since $\left|3T/\alpha\right| \leq T/6$, $\hat{\lambda}_{ne}(t) \in \{T, 4T\}$ on $I_1 := \left[T/3, 5T/12\right]$ or $I_2 :=[7T/12, 2T/3]$. From $g(x,y) \geq g(y,y), x,y >0$, 
\begin{align}
    \inf_{|\alpha| > 18, b \in \mathbb R}F(\alpha, b) \geq \min & \left\{ \int_{I_1} \left(g \left(T, \frac{9}{T}t^2\right) - g \left(\frac{9}{T}t^2, \frac{9}{T}t^2\right)  \right) \mathrm{d} t,\right. \nonumber\\
    &\int_{I_2} \left(g \left(T, \frac{9}{T}t^2\right) - g \left(\frac{9}{T}t^2, \frac{9}{T}t^2\right)  \right) \mathrm{d} t, \nonumber\\
    &\int_{I_1} \left(g \left(4T, \frac{9}{T}t^2\right) - g \left(\frac{9}{T}t^2, \frac{9}{T}t^2\right)  \right) \mathrm{d} t, \nonumber\\
    &\left.\int_{I_2} \left(g \left( 4T, \frac{9}{T}t^2\right) - g \left(\frac{9}{T}t^2, \frac{9}{T}t^2\right)  \right) \mathrm{d} t \right\}\nonumber\\
    := C_{1} & > 0.
    \label{C1 in counterex(main)}
\end{align}

\textbf{Case 2.} $|\alpha| \leq 18$, $|b| > 16T$. 

In this case , we can check that $\{t: T \leq \alpha t + b \leq 4T\} \cap I_0 = \emptyset$. Hence 
\begin{align}
    \inf_{|\alpha| \leq 18, |b| > 16T}F(\alpha, b) \geq \min & \left\{ \int_{I_0} \left(g \left(T, \frac{9}{T}t^2\right) - g \left(\frac{9}{T}t^2, \frac{9}{T}t^2\right)  \right) \mathrm{d} t,\right. \nonumber\\
    &\left.\int_{I_0} \left(g \left( 4T, \frac{9}{T}t^2\right) - g \left(\frac{9}{T}t^2, \frac{9}{T}t^2\right)  \right) \mathrm{d} t \right\}\nonumber\\
    := C_{2} & > 0.
     \label{C2 in counterex(main)}
\end{align}

\textbf{Case 3.} $|\alpha| \leq 18$, $|b| \leq  16T$. 

By \eqref{def of F in counterex(main)} , $F$ is continuous with respect to $(\alpha, b)$. For  fixed $(\alpha, b)$, since $f(\alpha t + b)\not\equiv \frac{9}{T^2}t^2$, $F(\alpha, b) > 0$.
Since $\{|\alpha| \leq 18, |b| \leq  16T\}$ is a compact set in $\mathbb R^2$, there exists $C_3 > 0$ such that
\begin{align}
     \inf_{|\alpha| \leq 18, |b| \leq 16T}F(\alpha, b) \geq C_3 > 0.
      \label{C3 in counterex(main)}
\end{align}
By \eqref{induce in I0 in counterex(main)}, \eqref{def of F in counterex(main)}, \eqref{C1 in counterex(main)}, \eqref{C2 in counterex(main)}, and \eqref{C3 in counterex(main)}, 
\begin{align}
    \mathbb E [\tilde{\text{loss}}(\hat{\lambda}_{ne})] - \mathbb E[\tilde{\text{loss}}(\lambda^{\ast})] \geq \min\{C_1, C_2, C_3\}\mathbb P(\mathcal{E}) := C > 0. \nonumber
\end{align}
Hence Theorem 8
is proved.

\begin{remark}
    Note that we have proved the excess risk
    \begin{align}\label{ES}
        \mathbb E[\text{loss}(\hat \lambda, S_{test})] - 
    \mathbb E[\text{loss}(\lambda^{\ast}, S_{test})]
    \end{align}
    is always positive if $\hat{\lambda} \neq \lambda^{\ast}$ in the proof of Theorem 8. Thus \eqref{ES} is a well-defined excess risk.
\end{remark}

\section{Supporting Lemmas}
\begin{lemma}
    (Lemma 8 in \cite{chen2019generalization}) \, Let $\mathcal{G} = \{A \in \mathbb R^{d_1 \times d_2}: \|A\|_2 \leq \lambda \}$ be the set of matrices with bounded spectral norm and $\epsilon > 0$ be given. The covering number $\mathcal{N}(\mathcal{G}, \epsilon, \|\cdot\|_{F})$ is bounded above by
    \begin{align}
        \mathcal{N}(\mathcal{G}, \epsilon, \|\cdot\|_{F}) \leq \left(1 + \frac{(\sqrt{d_1}\wedge \sqrt{d_2})\lambda}{\epsilon} \right)^{d_1 d_2} . \nonumber
    \end{align}
    \label{matrix covering number lemma(main)}
\end{lemma}
The following lemma is a bridge between the covering number and the upper bound of sub-gaussian process.
\begin{definition}
    A stochastic process $\{X_h\}_{h \in H}$ is called a sub-gaussian process for metric $d(\cdot, \cdot)$ on $H$ if 
    \begin{equation}
        \mathbb{E}\left[\exp \left(\lambda\left(X_{h_1}-X_{h_2}\right)\right)\right] \leq \exp \left(\frac{\lambda^2 d(h_1, h_2)^2}{2}\right) ~ \text { for } \lambda \in \mathbb{R},~ h_1, h_2 \in H. \nonumber
    \end{equation}
    A stochastic process $\{X_h\}_{h \in H}$ is called a centered sub-gaussian process for metric $d(\cdot, \cdot)$ on $H$ if $\{X_h\}_{h \in H}$ is a sub-gaussian process for metric $d(\cdot, \cdot)$ and $\mathbb{E}[X_h] = 0, ~ \forall h \in H$.
\end{definition}

\begin{lemma}
    \label{concentration lemma(main)}
    Suppose $\{X_h\}_{h \in H}$ is a centered sub-gaussian process for metric $K \cdot d(\cdot, \cdot)$ on metric space $H$, where the diameter of $H$ is finite, i.e.
    $\operatorname{diam}(H) = \sup_{h_1, h_2 \in H} d(h_1, h_2) < +\infty$.
    Then with probability at least $1 - \delta$,  for any fixed $h_0 \in H$, we have
    \begin{eqnarray}
        \sup _{h \in H} |X_h - X_{h_0}| \leq 6K \left(8 \operatorname{diam}(H) \sqrt{\log \left(\frac{2}{\delta}\right)} + \sum_{k=-\kappa}^{\infty} 2^{-k} \sqrt{\log \mathcal{N}\left(H, d, 2^{-k}\right)}\right) \nonumber
    \end{eqnarray}
    and
    \begin{eqnarray}
        \sup _{h \in H} |X_h - X_{h_0}| \leq 12K \left(4  \operatorname{diam}(H) \sqrt{\log \left(\frac{2}{\delta}\right)} + \int_0^{2\operatorname{diam}(H)} \sqrt{\log \mathcal{N}\left(H, d, \epsilon\right)}  ~ \mathrm{d} \epsilon \right), \nonumber
    \end{eqnarray}
    where $\kappa \in \mathbb{Z}_+$ satisfies $2^{\kappa - 1} < \operatorname{diam}(H) \leq 2^{\kappa}$.
\end{lemma}

\begin{proof}[Proof of Lemma \ref{concentration lemma(main)}]
    Let $\kappa \in \mathbb{Z}_+$ satisfy $2^{\kappa - 1} < \operatorname{diam}(H) \leq 2^{\kappa}$. Define $\epsilon_k = 2^{-k}, k \in \mathbb{Z}, k \geq -\kappa$. Let $H_k$ be the $\epsilon_k\text{-net}$ of $H$ with metric $d(\cdot, \cdot)$, i.e., $H_k \subset H$ covers $H$ at scale $\epsilon_k$ with respect to the metric $d(\cdot, \cdot)$. Clearly $|H_{-\kappa}| = 1$. We take $H_{-\kappa} = \{h_0\}$. Define $\pi_k(h)$ as the closest element of $h$ in $H_k$ under the metric $d(\cdot, \cdot)$. Then $\forall h \in H, \forall N \geq -\kappa, N \in \mathbb{Z}$, we have 
    \begin{eqnarray}
         X_h - X_{h_0} = \sum_{k=-\kappa + 1}^{\infty} \left(X_{\pi_{k}(h)}-X_{\pi_{k-1}(h)}\right) \quad a.s. ~. \nonumber
    \end{eqnarray}
    Thus
    \begin{eqnarray}
        \sup _{h \in H} |X_h-X_{h_0}| \leq \sum_{k=-\kappa + 1}^{\infty} \sup _{h \in H}\left|X_{\pi_{k}(h)}-X_{\pi_{k-1}(h)}\right| \quad a.s. ~. \nonumber
    \end{eqnarray}
    Consider $P_k = \{X_{\pi_{k}(h)}-X_{\pi_{k-1}(h)}| h \in H\}$, $|P_k| \leq |H_{k-1}| |H_k| \leq |H_{k}|^2$ and any element in $P_k$ is $K(\epsilon_{k} + \epsilon_{k-1})$ sub-gaussian. By Hoeffding's inequality and union bound argument, we have 
    \begin{align}
        \mathbb{P}\left(\sup_{X \in P_k} |X| \geq t \right) &= \mathbb{P}\left(\bigcup_{X \in P_k} \left\{|X| \geq t \right\}\right) \nonumber \\
        &\leq \sum_{X \in P_k} \mathbb{P}\left(|X| \geq t\right) \nonumber \\
        &\leq 2|P_k|\exp\left(-\frac{t^2}{2K^2(\epsilon_{k-1} + \epsilon_k)^2}\right) \nonumber \\
        &\leq 2|P_k|\exp\left(-\frac{t^2}{18K^2\epsilon_k^2}\right) .  \nonumber 
    \end{align}
    Let $2|P_k|\exp\left(-t^2/18K^2\epsilon_k^2\right) = \delta_{k} \leq 1/2$,  $t = \sqrt{18}K\epsilon_k\sqrt{\log(|P_k|) + \log(2/\delta_{k})} \leq 3\sqrt{2}K\epsilon_k (\sqrt{\log(|P_k|)} +\sqrt{\log(2/\delta_{k})})$ . Then with probability at least $1-\delta_{k}$, we have
    \begin{align}
        \sup_{X \in P_k} |X| &\leq 3\sqrt{2}K\epsilon_k \left(\sqrt{\log(|P_k|)} +\sqrt{\log(2/\delta_{k})}\right)  \nonumber \\
        &\leq 6K\epsilon_{k} \left(\sqrt{\log(|H_{k}|)} +\sqrt{\log(1/\delta_{k})}\right) . \nonumber
    \end{align}
    Thus, with probability at least $1-
    \sum_{k = -\kappa }^{+\infty}\delta_k$, we get
    \begin{eqnarray}
        \sup _{h \in H} |X_h - X_{h_0}| \leq 6 K \sum_{k=-\kappa}^{\infty} 2^{-k}\left(\sqrt{\log \mathcal{N}\left(H, d, 2^{-k}\right)}+\sqrt{\log \left(1 / \delta_k\right)}\right), \nonumber
    \end{eqnarray}
    Let $\delta_k = \delta/2^{k + \kappa + 1}$. Then $\sum_{k = -\kappa}^{\infty}\delta_k = \delta$. We have
    \begin{align}
    \sum_{k=-\kappa}^{\infty} 2^{-k} \sqrt{\log \left(1 / \delta_k\right)} &= \sum_{k=-\kappa}^{\infty} 2^{-k} \sqrt{\log \left(2^{k + \kappa + 1} / \delta \right)}  \nonumber\\
    &\leq  \sum_{k=-\kappa}^{\infty} 2^{-k}\sqrt{k+\kappa+1}\sqrt{\log \left(2 / \delta \right)} \nonumber \\
    &\leq 8\operatorname{diam}(H)\sqrt{\log \left(2 / \delta \right)} ~.  \nonumber
    \end{align}
    Thus,
    \begin{eqnarray}
        \sup _{h \in H} |X_h - X_{h_0}| \leq 6K \left(8 \operatorname{diam}(H) \sqrt{\log \left(\frac{2}{\delta}\right)} + \sum_{k=-\kappa}^{\infty} 2^{-k} \sqrt{\log \mathcal{N}\left(H, d, 2^{-k}\right)}\right). \nonumber
    \end{eqnarray}
    Since 
    \begin{align}
    \sum_{k=-\kappa}^{\infty} 2^{-k} \sqrt{\log \mathcal{N}\left(H, d, 2^{-k}\right)} \leq 2\int_0^{2^\kappa} \sqrt{\log \mathcal{N}\left(H, d, \epsilon\right)} ~ \mathrm{d} \epsilon \leq 2\int_0^{2\operatorname{diam}(H)} \sqrt{\log \mathcal{N}\left(H, d, \epsilon\right)} ~ \mathrm{d} \epsilon  ~,  \nonumber
    \end{align}
    the lemma is proved.
\end{proof}

\begin{lemma}
\label{theorem of tanh NN 1(main)}
    ~(Theorem 5.1 in \cite{de2021approximation})  ~ Let $d, s \in \mathbb N_{+}$, $\delta > 0$ and $f \in W^{s, \infty}([0,1]^d)$. There exist constants $\mathcal{C}(d,s,f)$ and $N_0(d) > 0$ such that for every integer $N > N_0(d)$, there exists a tanh neural network $\hat{f}^N$ with two hidden layers, with one width at most $3\lceil s/2\rceil\tbinom{s+d-1}{d}+d(N-1)$ and  the other width at most $3\lceil(d+2)/2\rceil\tbinom{2d+1}{d}N^d$ (or $3\lceil s/2\rceil + N - 1$ and $6N$ for $d=1$),  such that
    \begin{align}
        \left\|f - \hat{f}^N\right\|_{L^\infty([0,1]^d)} \leq (1+\delta) \frac{\mathcal{C}(d,s,f)}{N^s} ~. \nonumber
    \end{align}
    If $f \in C^s([0,1]^d)$, then it holds that
    \begin{align}
        \mathcal{C}(d,s,f) = \frac{(3d)^s}{s!2^s}\|f\|_{W^{s, \infty}([0,1]^d)}, \quad N_0(d) = \frac{3d}{2}, \nonumber
    \end{align}
    and else, it holds that
    \begin{align}
        \mathcal{C}(d,s,f) = \frac{\pi^{1/4}\sqrt{s}(5d)^s}{(s-1)!}\|f\|_{W^{s, \infty}([0,1]^d)}, \quad N_0(d) = 5d^2. \nonumber
    \end{align}
    Moreover, the weights of $\hat{f}^N$ scale as $O(\mathcal{C}(d,s,f)^{-s/2} N^{d(d+s^2)/2} (s(s+2))^{3s(s+2)})$.   
\end{lemma}
\begin{remark}
\label{the weights of approximation 1(main)}
    By Lemma \ref{theorem of tanh NN 1(main)}, there exists a constant $C(\delta)$ which is only dependent with $\delta$, such that
    \begin{align}
        |\text{the weights of $\hat{f}^N$}| \leq C(\delta) \mathcal{C}(d,s,f)^{-s/2} N^{d(d+s^2)/2} (s(s+2))^{3s(s+2)}. \nonumber
    \end{align}
\end{remark}

\begin{lemma}
\label{theorem of tanh NN 2(main)}
~(Corollary 5.8 in \cite{de2021approximation}) Let $d \in \mathbb N_{+}$, $\Omega \subset \mathbb R^d$ open with $[0,1]^d \subset \Omega$ and let $f$ be analytic on $\Omega$. If, for some $C > 0$, $f$ satisfies that $\|f\|_{W^{s, \infty}([0,1]^d)} \leq C^s$ for all $s \in\mathbb N$, then for any $\mathcal{N} \in \mathbb N_{+}$, there exists a one-layer $\tanh$ neural network $\hat{f}^{\mathcal{N}}$ of width $3\lceil (\mathcal{N} + 5Cd)/2\rceil\tbinom{\mathcal{N} + (5C+1)d}{d}$ (or $3\lceil\mathcal N/2\rceil$ for $d = 1$) such that
\begin{align}
    \left\|f - \hat{f}^N\right\|_{L^\infty([0,1]^d)} \leq \exp(-\mathcal{N}) ~. \nonumber
\end{align}
\end{lemma}
\begin{remark}
\label{the weights of approximation 2(main)}
    In \cite{de2021approximation}, the construction of $\hat{f}^{\mathcal{N}}$ in Lemma \ref{theorem of tanh NN 2(main)} uses Lemma \ref{theorem of tanh NN 1(main)} directly. Hence the weights of $\hat{f}^{\mathcal{N}}$ can be derived from Lemma \ref{theorem of tanh NN 1(main)}. Then there exists a constant $\tilde{C}$ such that
    \begin{align}
        |\text{the weights of $\hat{f}^{\mathcal{N}}$}| \leq \tilde{C} \exp(\frac{{\mathcal{N}^{\prime}}^2 + \mathcal{N}^{\prime}  -3Cd\mathcal{N}^{\prime}}{2}) (\mathcal{N}^{\prime}(\mathcal{N}^{\prime}+2))^{3\mathcal{N}^{\prime}(\mathcal{N}^{\prime}+2)}, \nonumber
    \end{align}
    where $\mathcal{N}^{\prime} = \mathcal{N} + 5Cd$.
    We emphasize that the original literature \citep{de2021approximation} does not give this result, but it can be obtained by simple calculations.
\end{remark}

\bibliographystyle{plainnat}
\bibliography{tpp}

\end{document}